\documentclass{article}

\usepackage[T1]{fontenc}

\usepackage{microtype}
\usepackage{graphicx}
\usepackage{subcaption}
\usepackage{booktabs} 
\usepackage{csvsimple}
\usepackage{longtable}
\usepackage{seqsplit}
\usepackage{float}
\usepackage[section,above,below]{placeins}

\usepackage{hyperref}
\usepackage{csvsimple-l3}
\usepackage[table]{xcolor}
\usepackage{rotating}
\usepackage{enumitem}
\usepackage{makecell}

\usepackage[accepted]{icml2026}

\usepackage{amsmath}
\usepackage{amssymb}
\usepackage{mathtools}
\usepackage{amsthm}
\usepackage{xurl}

\usepackage[capitalize,noabbrev]{cleveref}

\theoremstyle{plain}

\theoremstyle{definition}

\theoremstyle{remark}

\newcommand*{\ourmodel}{C3LM}
\newcommand*{\traindataset}{CREED}
\newcommand*{\benchmark}{ChemCensor}
\newcommand*{\expertset}{URSA-expert-2026}

\usepackage{multirow}

\usepackage[most]{tcolorbox}
\usepackage{xcolor}
\usepackage{caption}

\tcbset{before skip=0.8em, after skip=0.8em} 

\newtcolorbox{windowbox}[2][]{%
  enhanced,
  breakable,
  colback=white,
  colframe=black,
  boxrule=0.8pt,
  arc=2mm,
  left=3mm,right=3mm,top=2mm,bottom=2mm,
  fonttitle=\bfseries,
  coltitle=white,
  colbacktitle=black,
  title={#2},
  attach boxed title to top left={xshift=0mm,yshift=-1mm},
  boxed title style={sharp corners, boxrule=0pt, interior style={fill=black}},
  #1
}

\newtcolorbox{fieldbox}[2][]{%
  enhanced,
  breakable,
  colback=gray!3,
  colframe=black!25,
  boxrule=0.5pt,
  arc=1.5mm,
  left=2mm,right=2mm,top=1mm,bottom=1mm,
  fonttitle=\bfseries,
  title={#2},
  listing only,
  listing options={
    basicstyle=\ttfamily\footnotesize,
    columns=fullflexible,
    breaklines=true,
    showstringspaces=false,
  },
  #1
}

\usepackage[textsize=tiny]{todonotes}
\usepackage{pdfpages}

\icmltitlerunning{When Single Answer Is Not Enough: Rethinking Single-Step Retrosynthesis Benchmarks for LLMs}

\usepackage{listings}
\begin{document}

\twocolumn[
  \icmltitle{When Single Answer Is Not Enough: \\
  Rethinking Single-Step Retrosynthesis Benchmarks for LLMs}
    


  \icmlsetsymbol{equal}{*}

  \begin{icmlauthorlist}
    \icmlauthor{Bogdan Zagribelnyy}{uae}
    \icmlauthor{Ivan Ilin}{uae}
    \icmlauthor{Maksim Kuznetsov}{can}
    \icmlauthor{Nikita Bondarev}{uae}
    \icmlauthor{Mathieu Reymond}{can}
    \icmlauthor{Roman Schutski}{chi}
    \icmlauthor{Thomas MacDougall}{can}
    \icmlauthor{Rim Shayakhmetov}{uae}
    \icmlauthor{Zulfat Miftahutdinov}{can}
    \icmlauthor{Mikolaj Mizera}{chi}
    \icmlauthor{Vladimir Aladinskiy}{uae}
    \icmlauthor{Alex Aliper}{uae}
    \icmlauthor{Alex Zhavoronkov}{uae,can,chi}
  \end{icmlauthorlist}

  \icmlaffiliation{uae}{Insilico Medicine AI Limited, Level 6, Unit 08, Block A, IRENA HQ Building, Masdar City, Abu Dhabi, UAE}
  \icmlaffiliation{chi}{Insilico Medicine Hong Kong Ltd., Unit 310, 3/F, Building 8W, Phase 2, Hong Kong Science Park, Pak Shek Kok, New Territories, Hong Kong, Hong Kong SAR, China}
  \icmlaffiliation{can}{Insilico Medicine Canada Inc., 3710-1250 René-Lévesque Blvd W, Montreal, Quebec, H3B 4W8, Canada}

  \icmlcorrespondingauthor{Bogdan Zagribelnyy}{bogdan@insilicomedicine.com}

  \icmlkeywords{Machine Learning, ICML}

  \vskip 0.3in
]



\printAffiliationsAndNotice{}  


\begin{abstract}
Recent progress has expanded the use of large language models (LLMs) in drug discovery, including synthesis planning. However, objective evaluation of retrosynthesis performance remains limited. Existing benchmarks and metrics typically rely on published synthetic procedures and Top-K accuracy based on single ground-truth, which does not capture the open-ended nature of real-world synthesis planning. We propose a new benchmarking framework for single-step retrosynthesis that evaluates both general-purpose and chemistry-specialized LLMs using ChemCensor, a novel metric for chemical plausibility. By emphasizing plausibility over exact match, this approach better aligns with human synthesis planning practices. We also introduce CREED, a novel dataset comprising millions of ChemCensor-validated reaction records for LLM training, and use it to train a model that improves over the LLM baselines under this benchmark.
\end{abstract}


\section{Introduction}
\label{sec:intro}

\begin{figure}[t]
  \centering

  \includegraphics[width=\columnwidth]{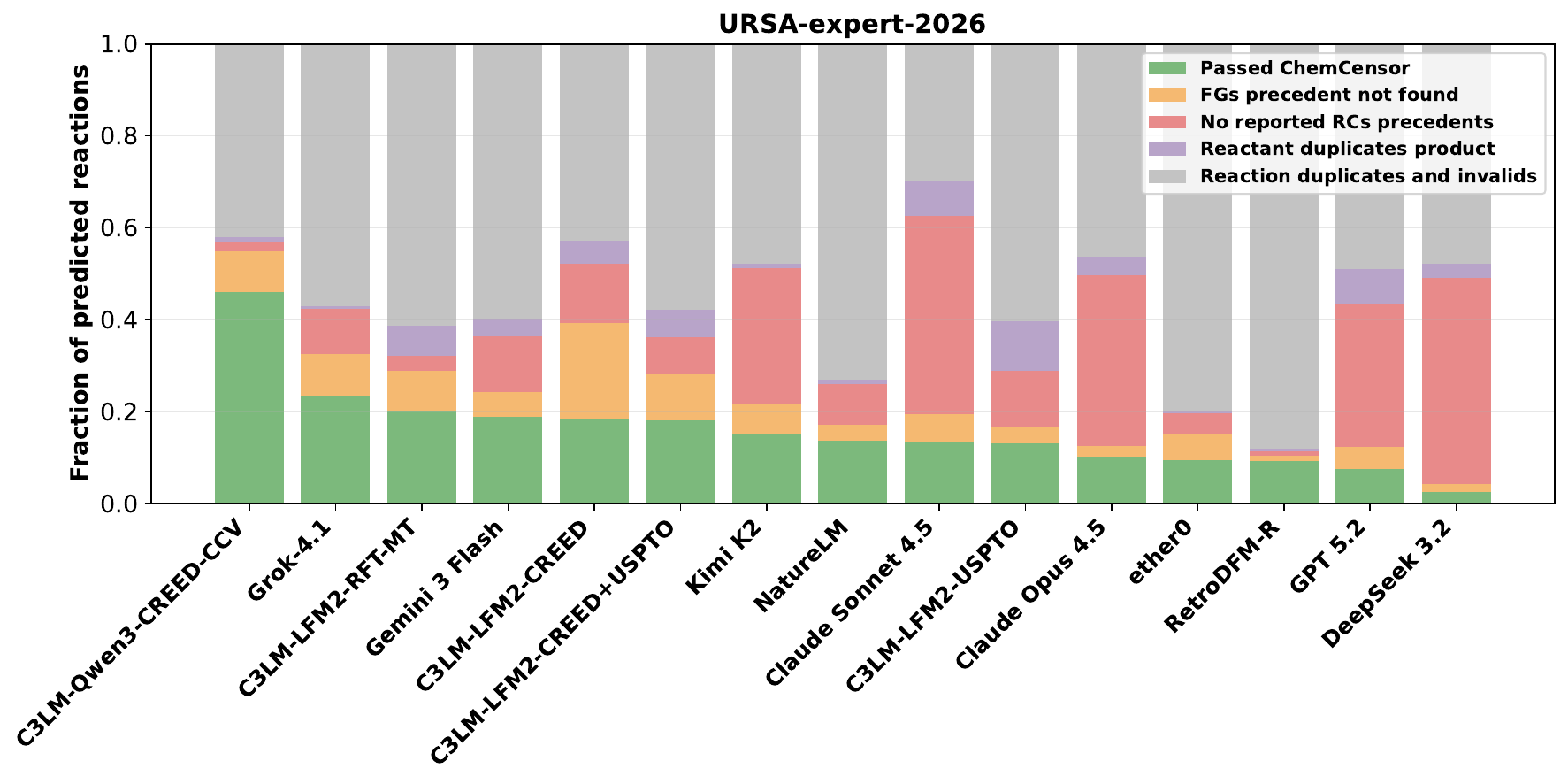}
  \vspace{0.00cm}
  \includegraphics[width=\columnwidth]{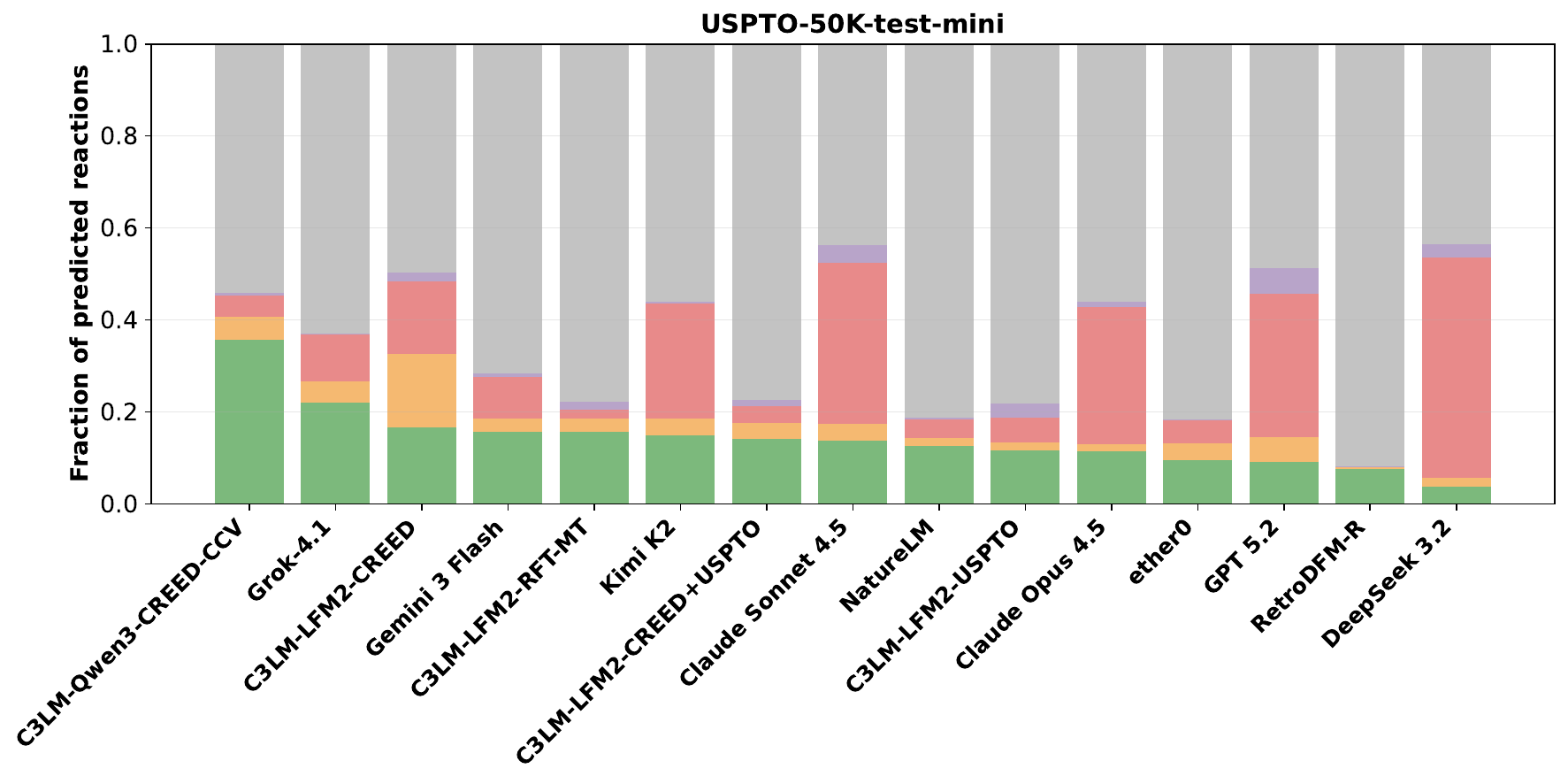}
  \caption{Assessment of generated reactants from benchmarked models by ChemCensor.}
  \label{fig:chemcensor_categ_by_model}
\end{figure}

The modern small molecule drug development process requires thoughtful control over the synthetic accessibility of the designed molecules~\cite{gao2020synthesizability}.
In practice, synthetic feasibility is assessed through synthesis planning, most prominently retrosynthetic analysis~\cite{VLEDUTS1963117, corey1967general}. 
Computer-aided synthesis planning (CASP) aims to automate this process by proposing routes to available building blocks~\cite{tu2023computer}. 
While direct multistep models are emerging~\cite{shee2025directmultistep}, many CASP tools still decompose the problem into (i) \textit{single-step retrosynthesis} (SSRS), which proposes candidate disconnections, and (ii) \textit{multistep retrosynthesis} (MSRS), which assembles these steps into complete routes~\cite{zhong2024recent}.

Despite substantial progress of CASP systems~\cite{grzybowski2018chematica}, their evaluation remains a bottleneck. 
Reliable automated assessment is challenging due to incomplete and heterogeneous databases, strong dependence of reactivity on molecular context, and the limited scalability of expert review ~\cite{maziarz2025syntheseus}. 
As a result, benchmarking practices are fragmented and often hard to reproduce across models and datasets~\cite{jiang2022-aiforrs}. Commonly used metrics, e.g., Top-K exact-match accuracy for SSRS~\cite{segler2017ml-in-rs}, solvability for MSRS~\cite{segler2018-solvability,choe2025-crosstalk}, and related variants~\cite{schwaller2020roundtrip}, provide only indirect and limited signals of chemical plausibility \cite{morgunov2025procrusteanbedaidrivenretrosynthesis}. 
In particular, exact-match metrics implicitly assume a single “ground-truth” disconnection, even though many targets admit multiple valid retrosynthetic choices, and they do not directly test key chemical criteria such as reaction-center validity, functional-group compatibility, and selectivity~\cite{westerlund2025-azfg2,maziarz2025syntheseus}. 
Expert databases (e.g., Reaxys~\cite{reaxys}, SciFinder~\cite{scifinder}) can support detailed reaction-by-reaction analysis, but they are not designed for automated benchmarking of model outputs at a large scale.

The rise of large language models (LLMs) \cite{wolf2020transformers, brown2020gpt}, their flourishing variety and continuous emergence of new LLMs alongside the claims of their potential applicability in the synthetic planning domain \cite{xuan2025synthelite}  demands unprecedented benchmarking approaches to objectively assess the performance of LLMs in real-world out-of-domain (OOD) tasks of retrosynthetic analysis applied to the drug discovery process, since the disadvantages of the traditional USPTO-50K-test are recognized by the community~\cite{Peraire2024ModelsMatter}.

In our study, we utilize the new metric called \benchmark{} as an alternative to traditional Top-K accuracy to assess SSRS outputs of LLMs, emphasizing the detrimental impact of “ground truth” dogma promoted by Top-K accuracy. \benchmark{} enables benchmarking with broader chemical context augmentation and structured benchmark compound sets creation exemplified by the \textbf{\expertset{}} set of $100$ completely novel molecular structures with expert assessment of their synthetic feasibility. The unique chemical reactions dataset (\textbf{\traindataset{}}) of unprecedented size ($\sim22.7$M) and reliability was created using $\sim3$K in-house reaction templates. \traindataset{} was used to fine-tune the novel \textbf{\ourmodel{}} language model, achieving performance of the best general-purpose foundation and chemical generalist models. 

The key contributions of our work are as follows:
\begin{enumerate}[itemsep=0pt, topsep=0pt, parsep=0pt, partopsep=0pt]
\item We propose \textbf{\benchmark{}}, a metric for evaluating SSRS predictions beyond exact-match Top-K accuracy. 

\item We construct \textbf{\traindataset{}}, a 22.7M reactions dataset derived from expert-coded templates. We also propose a ChemCensor-verified \textbf{CREED-CCV} subset.

\item We release \textbf{\expertset{}}, an expert-annotated benchmark of $100$ novel targets for SSRS evaluation.

\item We train the \textbf{\ourmodel{}} family of models on the \textbf{\traindataset{}}, improving the performance over general-purpose LLMs and chemical generalist models.
\end{enumerate}

\textbf{Conflict of Interest Disclosure}
The authors are employed by Insilico Medicine, a group of commercial AI companies, leading the development of the C3LM family of models.


\section{Related Work}
\label{sec:related_work}

\subsection{Benchmarking solutions for SSRS}

\textbf{Metrics.} Top-K accuracy ~\cite{segler2017ml-in-rs} is a standard evaluation metric for SSRS. For each product molecule, a model outputs an ordered list of candidate precursor (reactant) sets ranked by confidence; Top-K accuracy reports the fraction of test cases in which the recorded ground-truth precursor set appears among the top K predictions. 

Round-trip accuracy is an evaluation metric that measures whether a retrosynthesis model’s predicted reactants can be forward-validated by a reaction prediction model ~\cite{schwaller2020roundtrip}. It quantifies the percentage of retrosynthesis predictions that, when passed through a forward reaction model, reproduce the original product. 

Beyond exact-match Top-K, several works advocate distribution- and diversity-aware evaluation to better reflect how a model behaves as a generator of chemically meaningful suggestions \cite{Schwaller2019EvalMetrics}. In this line, coverage quantifies how broadly the model explores reaction space, for example, as the fraction of reaction classes (or templates) present in the reference set that also appear among the model’s predictions \cite{Hastedt2024RetroBM}. Class diversity captures how varied the predicted chemistry is, helping detect mode collapse where a model repeatedly proposes the same few transformations. Finally, the Jensen–Shannon divergence \cite{Lin1991JSD} measures how closely the distribution of predicted reaction classes matches the reference distribution.

The MaxFrag accuracy~\cite{Tetko2020NatCommun} emphasizes the largest fragment in the predicted reactant set, offering a targeted evaluation that is less influenced by ambiguous or unclear reagent information in USPTO-based reaction sets. This metric is particularly useful when annotation noise or reagent ambiguity can obscure the true performance of retrosynthetic models. In a different approach~\cite{Liu2023FusionRetro}, SSRS models are evaluated via linkage to a search algorithm in terms of exact reaction-match accuracy across the rebuilt reaction network from USPTO. Nevertheless, both the latter approach and MaxFrag accuracy inherit the “ground-truth” dogma of the Top-K accuracy metric.

Recent LLM-style retrosynthesis papers \cite{Zhao2025ChemDFM, Lin2025ChemDual} report a broad suite of evaluation metrics, including exact match, \textsc{BLEU}~\cite{Papineni2002BLEU}, Levenshtein edit distance~\cite{Levenshtein1966}, and molecular similarity-based scores computed between the predicted precursor set and the corresponding ground-truth precursors. 

RetroTrim \cite{sadowski2025retrotrim} proposes an ensemble of
reaction scorers for hallucination filtering and plausibility assessment, but its current preprint does not release code or
weights for independent benchmarking. 

\textbf{Benchmarking infrastructure.} Syntheseus is an open, modular benchmarking platform designed to evaluate and compare conventional SSRS models under standardized and reproducible conditions \cite{maziarz2025syntheseus}. In addition to classic Top-K accuracy, Syntheseus reports a few technical metrics, which are useful, but do not capture aspects of chemistry-oriented evaluation.

\subsection{LLMs for the SSRS task}

The sequential representation of molecules via SMILES enables SSRS to be formulated as a product-to-reactant translation task with LLMs, typically trained via large-scale pretraining and then adapted via instruction tuning or reaction-specific fine-tuning on patent reaction corpora, e.g., USPTO-derived sets. Recent work spans (i) \emph{generalist or science foundation} models, including NatureLM~\cite{Xia2025NatureLM}, nach0~\cite{livne2024nacho}, and ChemDFM~\cite{Zhao2025ChemDFM}, can be adapted to retrosynthesis; (ii) \emph{data-augmentation} strategies that expand reaction supervision, including template-driven large-scale generation and pretraining in RSGPT~\cite{Deng2025RSGPT} as well as fragment/recombination-style augmentation and joint reaction--retrosynthesis training objectives in ChemDual~\cite{Lin2025ChemDual}; and (iii) \emph{tool-augmented} LLM systems that pair a general LLM with external chemistry tools for planning, prediction, and validation, notably ChemCrow~\cite{Bran2023ChemCrow}, SynAsk~\cite{Zhang2024SynAsk}). Beyond pure single-step generators, multimodal and route-context methods incorporate additional signals such as textual route context in RetroInText~\cite{Lin2025RETROINTEXT}, while other lines explore using GP LLMs inside decision/search programs for synthesis planning, as in LLM-augmented decision programs~\cite{Wang2025lmdecisionprograms, liu2025retror1}.

In parallel, larger chemistry-focused LLMs and instruction-tuned checkpoints such as BatGPT-Chem~\cite{Yang2024BatGPTChem}, ChemLLM~\cite{Zhang2024ChemLLM}, Mol-Instructions / LLaMA-MolInst adapters~\cite{Fang2023MolInstructions} aim to improve general chemical competence and controllability for downstream reaction tasks. Finally, emerging “reasoning-oriented” post-training directions target more structured chemical reasoning for retrosynthesis as seen in Chemma~\cite{Zhang2025Chemma}, atom-anchored LLM retrosynthesis~\cite{Hassen2025AtomAnchored}, and ether0~\cite{narayanan2025ether0}, attempting to improve decomposition/editing behavior without requiring additional chemistry-domain pretraining from scratch.

\textbf{Benchmarking of LLMs.}
While several recent benchmarks evaluate LLMs on broad chemistry knowledge and reasoning \cite{Guo2024Whatcanlargelanguagemodelsdo, Runcie2026Assessing-the-chemical-intelligence, li2026rxnbench, song2025evaluatinglargelanguagemodels, zhao2025superchembenchmark}, they typically include general chemistry tasks, and few or no synthesis-planning tasks and therefore only indirectly reflect retrosynthetic capability.

\subsection{Conventional SSRS models}
\label{subsec:convSSRS}

While LLMs are emerging as CASP tools, many of specialized lightweight SSRS models have long been used in synthesis-planning workflows. Syntheseus allows for user-friendly evaluation of such models, including LocalRetro \cite{chen2021localretro}, RetroKNN \cite{xie2023retrosknn}, R-SMILES \cite{Zhong2022-root-aligned-smiles} and others. 

\section{Approach}
\label{sec:approach}

\begin{figure*}[th]
  \centering
  \includegraphics[
    width=\linewidth,
    trim=40 200 20 150,
    clip
  ]{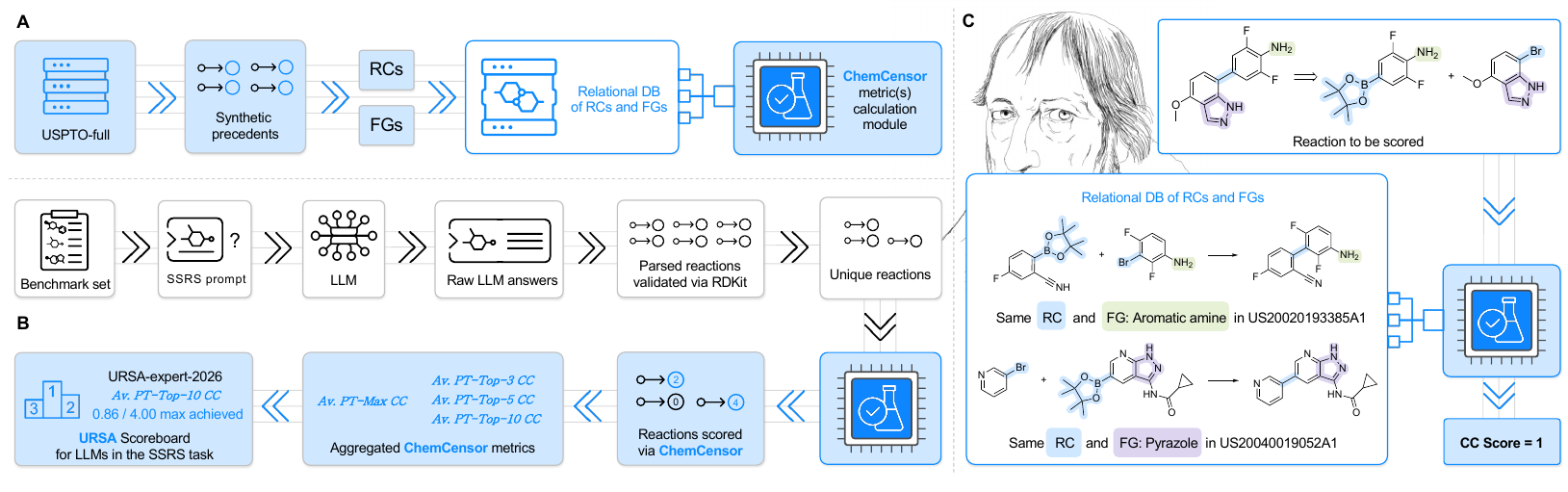}
  \caption{
ChemCensor part of the URSA benchmark. \textbf{A.} The procedure of retrieval and chemistry-aware knowledge aggregation from \textbf{USPTO-full} into the relational DB of RCs and FGs. \textbf{B.} The pipeline of LLMs benchmarking for the SSRS task. \textbf{C.} The example of RC and FGs intercompatibility assessment during single reaction scoring within ChemCensor. Portrait credits to \citeauthor{dreamstime_hegel_284053415}.
}
  \label{fig:CC_pipe}
\end{figure*}

The vast majority of SSRS benchmarks are based on the USPTO-50K test set \cite{maziarz2025syntheseus, Peraire2024ModelsMatter, segler2018-solvability, schwaller2020roundtrip}. This set consists of $\sim5$K pairs of products and corresponding reactant(s), with each product typically having only one unique set of reactants. Any predicted set of reactants that does not fully match the provided answers would be considered incorrect and the respective Top-K accuracy values will go down. This approach completely disregards the actual purpose of SSRS, in which multiple plausible options for synthesizing a compound in the next-step reaction can be considered. For the vast majority of compounds, especially complex ones, there are decisively more than one way to synthesize them (see examples at \autoref{app:when_single_answer_is_not_enough}). Undoubtedly, having a reaction exactly matching the reported example is a great fortune for a chemist. However, in most cases, because of the dispersed nature of chemical reaction space, when designing a retrosynthetic route, a chemist operates with more general concepts of reaction centers' and functional groups' intercompatibility, making a few steps of abstraction from exact matches. Here, to tackle the drawbacks of the previous "single ground-truth" paradigm, we propose \textbf{ChemCensor}, a new data-driven framework to score reactions based on the extraction of reaction centers and functional groups and their eigenprecedents and intercompatibility precedents analysis. By design, the framework mimics the way how a synthetic chemist evaluates a potential reaction to be included in the retrosynthetic route within search in chemical expert databases (e.g., Reaxys, SciFinder). 
In other words, ChemCensor was inspired by a chemistry-aware reformulation of the Hegelian principle: \textit{“what is plausible is real; and what is real is plausible”} (\citeauthor{Hegel2008_Philosophy_of_Right}). Unlike classical Top-K accuracy benchmarks, which treat a single ultra-concrete ground-truth reactant set as the main criterion of correctness, ChemCensor shifts the focus from exact historical reproduction to chemically grounded plausibility, preserving the abstract chemical logic connecting reaction centers and functional-group environments observed across real synthetic precedents. In this sense, it operationalizes a synthesis-practice intuition: chemists do not search for arbitrary formal possibilities, but for transformations whose \textit{plausibility is grounded in accumulated experimental reality using a reasonable level of abstraction}. 

\subsection{ChemCensor Metric}
\label{sec:chemcensor}
\textbf{ChemCensor} (CC) is a precedent-based, quantitative metric for measuring \textit{chemical plausibility} (\autoref{fig:CC_pipe}). For each reaction, the algorithm extracts its reaction center (\textbf{RC}, what bonds/atoms change) and the “static” functional-group (\textbf{FG}) context (which FGs remain present but are not transformed). The algorithm then checks whether the RCs and FGs have been observed in a relational database of reported reactions. The output (\textbf{ChemCensor Score}) is an integer confidence level ($0–5$) that directly reflects how well the step is supported by documented \textit{synthetic precedents}. The higher CC values correspond to the broader substructural context of RC that can be matched to the actual synthetic precedent. 
In practical terms, CC answers a simple question: \textit{“Have chemists successfully done this type of transformation before, and in the presence of these FGs?”} If yes, the reaction receives a higher score; if the synthetic precedents match only a topologically limited substructural context of RC, the score is low; if the algorithm cannot find supporting precedents for the RC and/or its FGs environment, the score is zero. 
The scoring procedure follows four steps:

\begin{enumerate}[itemsep=0pt, topsep=0pt, parsep=0pt, partopsep=0pt]
 \item \textbf{Split “what reacts” (RC) vs “what must be tolerated” (FGs)}. The reaction is decomposed into 
  (i) a hierarchical RC representation and
  (ii) a FG signature for the non-participating parts of the structures. 
\item \textbf{Look up precedents for the RC}. The extracted RC patterns are matched against a precomputed library built from a reference dataset of confirmed reactions. 
\item \textbf{Verify FGs compatibility}. The system checks whether the FGs present in the proposed step are compatible with that reaction center, based on FGs signatures aggregated from precedent examples. 
\item \textbf{Assign a confidence level (CC Score)}. 
\end{enumerate}

In this work, we distinguish five levels of RC annotation, denoted RC1, RC2, RC3, RC4 and RC5 corresponding to confidence levels and CC Scores from $1$ to $5$, respectively. The confidence level number provides the knowledge of the substructural context size of RC supported within synthetic precedents and FGs ({$N$=$516$}) compatibility precedents at this RC substructural context size. The smaller RC context size might be more popular in the reference dataset, but possesses low confidence. Oppositely, a larger RC context size provides greater confidence, but obviously is less represented in the reference set, especially considering the common context of non-reacting FGs. The details on how RC1-5 and FGs are organized are provided in \autoref{app:chemcensor_rc}, while the examples of how ChemCensor handles various aspects of chemical plausibility (including chemo-, regio- and stereoselectivity) are provided in \autoref{app:detailed_chemcensor_output}. Independently, the outputs of selected baseline LLMs and C3LMs were scored by CC and expert chemists to prove the discriminative power of the suggested metric (see \autoref{app:chemcensor_validation}).  

CC knowledge base is grounded on the best-to-date publicly available \textbf{USPTO-full} dataset covering reactions ($N = 1.4M$) from the US Patents from 1976 to 2016 \cite{Lowe2017usptofull}. CC versions based on this reference set are denoted with the -U2 postfix. To validate the approach within extended synthetic precedent space ($N = 4.8M$ reactions), a commercially available Pistachio set \cite{pistachio_nextmove} was added to \textbf{USPTO-full}. The CC versions based on this set are denoted with the -U2P2 postfix. 

ChemCensor is a part of a larger
benchmarking system \textbf{URSA} (\textbf{U}tilitarian \textbf{R}etro\textbf{S}ynthesis
\textbf{A}ssessment), which is designed to score not only outputs
from SSRS models, but also MSRS solutions and to be reported in detail separately.

\begin{figure*}[ht]
  \centering
  \includegraphics[
    width=0.8\linewidth,
    trim=20 200 100 120,
    clip
  ]{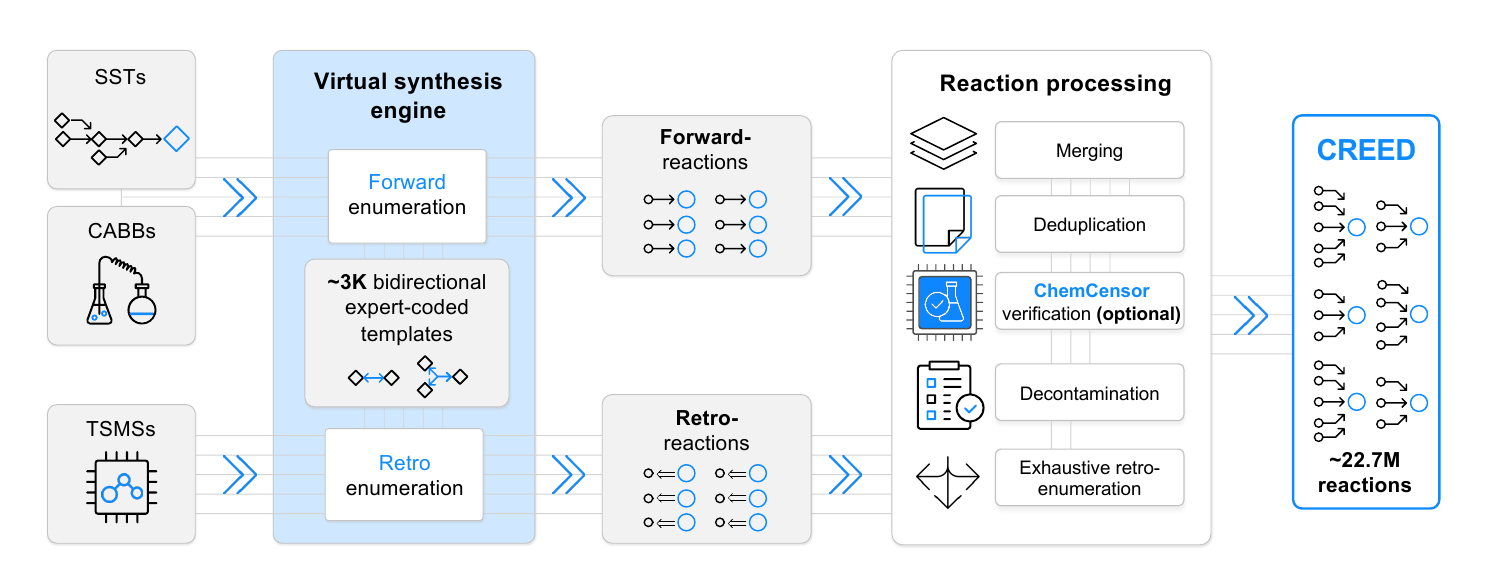}
  \caption{
  CREED (Comprehensive Reactant Exhaustive Enumeration Dataset) design process.
}
  \label{fig:CREED_constructing}
\end{figure*}

\subsection{Benchmarking Sets} 
\label{sec:expert_set}

To evaluate the SSRS task outputs from LLMs, we proposed a new dataset, \textbf{\expertset{}}. It is a benchmark dataset consisting of $100$ synthetically plausible, machine-generated target molecules. They were generated within the generative chemistry pipeline \cite{Ivanenkov2023chemistry42} as potential inhibitors of novel drug targets \cite{Oprea2021targets2020, Oprea2021targets2021, Oprea2021targets2022}. The synthetic accessibility status of these molecular structures is confirmed by expert synthetic chemists, who provided validated theoretical synthetic schemes (see example in \autoref{app:expert_set_app}) supported by relevant references for reactions and leading to the curated CABBs (\textit{commercially available building blocks}). All molecular structures in this dataset are novel and non-overlapping with publicly available reaction datasets, ensuring that they have not been used for training retrosynthesis models. The \expertset{} set is intended as a benchmarking set for evaluating generalization, robustness, and exploratory capabilities of SSRS models under conditions free from training-set memorization or data leakage.

We additionally report results on the widely adopted \textbf{USPTO-50K-test} \cite{liu2017uspto50ktest} to ensure direct comparability with prior SSRS literature. For consistency with our evaluation protocol and to improve experimental reliability, we use a curated version of USPTO-50K-test, removing a small subset of reactions according to criteria described in \autoref{app:uspto_50_k_test_mods}. This dual-benchmark setup (\expertset{} as the main generalization benchmark and USPTO-50K-test as the community-standard reference) enables evaluation both on a leakage-resistant target pool and on an established baseline used across SSRS studies.

\subsection{Aggregated Metrics} 
\label{sec:integral_metrics}
We use ChemCensor Score (CC) as the per-reaction plausibility metric. For model-level aggregation, we define the following metrics:

\textbf{Av. PT-Max CC} is calculated as the average of the $\mathrm{Max\ CC}$ values over all target molecules $t$ in the benchmark set of size $N$, where $\mathrm{Max\ CC}$ for a target denotes the highest $\mathrm{CC}$ value among all $\mathrm{CC}$ scores obtained for the set of possible retro-reactions $\mathcal{R}_t$ predicted for that target by a model:

\begin{eqnarray}
\mathrm{Max\ CC}(t) &=& \max_{r \in \mathcal{R}_t} \mathrm{CC}(t,r)
\label{eq:max-cc}\\
\mathrm{Av.~PT\text{-}Max\ CC} &=& \frac{1}{N}\sum_{t=1}^{N} \mathrm{Max\ CC}(t)
\label{eq:apt-max-cc}
\end{eqnarray}

\textbf{Av. PT-Top-K} is calculated as the average of $\mathrm{CC@K}(t)$ over all target molecules in the benchmark dataset of size $N$, where $\mathrm{CC@K}(t)$ is the mean $\mathrm{CC}$ score across the Top-$K$ \emph{unique} predicted retro-reactions for target molecule $t$. To explicitly account for prediction diversity, the ranked list of model outputs is first \emph{deduplicated} (exact duplicates are removed). If, after deduplication, fewer than $K$ unique predictions remain, the missing slots are padded with zeros. This prevents artificially inflating the metric by repeating the same high-scoring prediction multiple times.

\begin{eqnarray}
\mathrm{CC@K}(t) &=& \frac{1}{K}\sum_{k=1}^{K} \mathrm{CC}\!\left(t, r^{\text{uniq}}_{t,k}\right)
\label{eq:cc-at-k}\\
\mathrm{Av.~PT\text{-}Top\text{-}K} &=& \frac{1}{N}\sum_{t=1}^{N} \mathrm{CC@K}(t)
\label{eq:apt-top-k}
\end{eqnarray}

\subsection{CREED dataset}
\label{sec:train_dataset}

In this work, we also introduce \textbf{CREED} (\textbf{C}omprehensive \textbf{R}eactant \textbf{E}xhaustive \textbf{E}numeration \textbf{D}ataset), a large-scale and quality-controlled dataset of chemical reactions. CREED was built using two generation paradigms supported by an in-house virtual synthesis engine and the CC verification framework (see Fig.~\ref{fig:CREED_constructing}). First, in a forward route-generation regime, we performed construction of multistep synthetic pathways using the forward synthesis engine fueled with 1) $\sim$3K expert-coded bidirectional reaction templates, 2) CABBs as \textit{starting materials} and 3) synthetic scheme templates (SSTs) derived from real-world syntheses of approved drugs (e.g., \cite{Flick2022fda2020}). Individual elementary reactions were then extracted from the forward-generated routes. This approach covered $\sim$$67\%$ of reaction templates. The remaining $\sim$$33\%$ templates were not covered within drug-aware SSTs. Thus, we had to generate plausible template-specific molecular structures (TSMSs), which could be applied to the remaining templates in the retro-regime. After generating such molecular structures using the generative chemistry approach \cite{Ivanenkov2023chemistry42}, the remaining templates were applied to them; and the resulting retro-reactions covered the missing templates. Afterwards, both the retro-reactions and forward-reactions were merged. After this step and decontamination against the USPTO-50K-test set, CREED comprises $22,712,526$ unique reactions for $1,493,715$ unique products.

CREED is designed to prioritize \textit{chemical plausibility} over “single-ground-truth” convention. Multiple plausible reaction options for a single product are provided to enable diverse plausible SSRS model outputs, contrastingly to the USPTO-full, which mostly contains a single set of reactants per product. See such examples of MA (Multiple Answer) options and other details of CREED in \autoref{app:CREED_stats}.

An alternative dataset version, CREED-CCV (\textbf{C}hem\textbf{C}ensor \textbf{V}erification), was prepared using ChemCensor (v.0.5.2) as a verification framework for reaction plausibility. After the merging step of retro-reactions and forward-reactions, the candidate reactions were subsequently verified with CC, and only those with CC Score $>0$ were retained. Finally, CREED-CCV comprises $6,368,986$ unique reactions with theoretically supported plausibility by the CC module for $698,765$ unique products.

\section{Experiments}
\label{sec:experiments}

In this section, we conduct an evaluation of recent foundation models and chemical specialists on the proposed \benchmark{} benchmark to investigate the capabilities of these models in the single-step retrosynthesis task. Moreover, we train a family of novel \textbf{\ourmodel{}} (\textbf{C}hemistry \textbf{C}onstraint–\textbf{C}onsistent \textbf{L}anguage \textbf{M}odel) models on the proposed \traindataset{} and show that carefully curated train corpora can substantially improve LM performance on the SSRS task. Additional training, resources, optimization, and other details can be found in \autoref{app:training_details}.

\subsection{Baseline Models}
\label{sec:baselines}

As baselines for our benchmark, we adopt the set of proprietary and open-weight LMs as well as a set of SSRS conventional models. Our selection is focused on (i) widely accessible general-purpose LLMs, (ii) recent LMs with general chemistry or synthesis knowledge and (iii) conventional SSRS models, specifically:

\textbf{Proprietary Foundation Models:} Grok 4.1 \citeyear{xai2025grok41}; Gemini 2.5 Flash \citeyear{google2025gemini25flash} and 3 Flash preview \citeyear{google2025gemini3flash};  GPT 5.1 \citeyear{openai2025gpt51} and 5.2 \citeyear{openai2025gpt52}; Claude Sonnet 4.5 \citeyear{anthropic2025claudesonnet45} and Opus 4.5 \citeyear{anthropic2025claudeopus45}.

\textbf{Open-weight Foundation Models:} DeepSeek 3.2 \citeyear{deepseekai2025deepseekv32}; Qwen 3 \citeyear{qwen2025qwen3technicalreport} and Kimi K2 \citeyear{kimiteam2025kimik2openagentic}.

\textbf{Open-weight Chemical Specialist Models:} ether0 \citeyear{narayanan2025ether0}; NatureLM \citeyear{Xia2025NatureLM} and RetroDFM-R \citeyear{zhang2025retrodfm}.

\textbf{Conventional SSRS models:} the following models were evaluated via Syntheseus: LocalRetro \cite{chen2021localretro}, R-SMILES \cite{Zhong2022-root-aligned-smiles}, MEGAN \cite{sacha2021megan}, Chemformer \cite{irwin2022chemformer}, Graph2Edits \cite{Zhong2023-graph2edits}, GLN \cite{coley_gln}, MHNreact \cite{Seidl2022mhnreact} and RetroKNN \cite{xie2023retrosknn}.

\subsection{\ourmodel{} Supervised Fine-Tuning}
\label{sec:sft_train_details}

\textbf{Training setups:}
We perform supervised fine-tuning (SFT) of the \ourmodel{} model under three data configurations. Specifically, we train on either \traindataset{}, USPTO-full~\cite{Lowe2017usptofull}, or a concatenation of \traindataset{} and USPTO-full, where USPTO-full is decontaminated with respect to the benchmark sets and upsampled to match the size of \traindataset{}.

We initialize \ourmodel{}s with \texttt{LFM2 2.6B} checkpoint \cite{amini2025lfm2} and train \textit{with basic reasoning} (see \autoref{app:training_details}) for $50{,}000$ steps:
\texttt{C3LM-LFM2-USPTO} on USPTO-full,
\texttt{C3LM-LFM2-CREED} on CREED, and
\texttt{C3LM-LFM2-CREED+USPTO} trained on CREED+USPTO-full, respectively. Alternatively, \texttt{C3LMs-Qwen3} based on the \texttt{Qwen3 8B} checkpoint (\citeyear{qwen2025qwen3technicalreport}) were trained on CREED-CCV w/ and w/o USPTO--full (see \autoref{app:ablation_studies}). The full inventory of the C3LM family of models (both SFT and RFT) can be found in \autoref{app:model_inventory}.

\begin{figure*}[h]
  \centering
  \includegraphics[
    width=\linewidth,
    clip,
    trim=20 255 150 60,
  ]{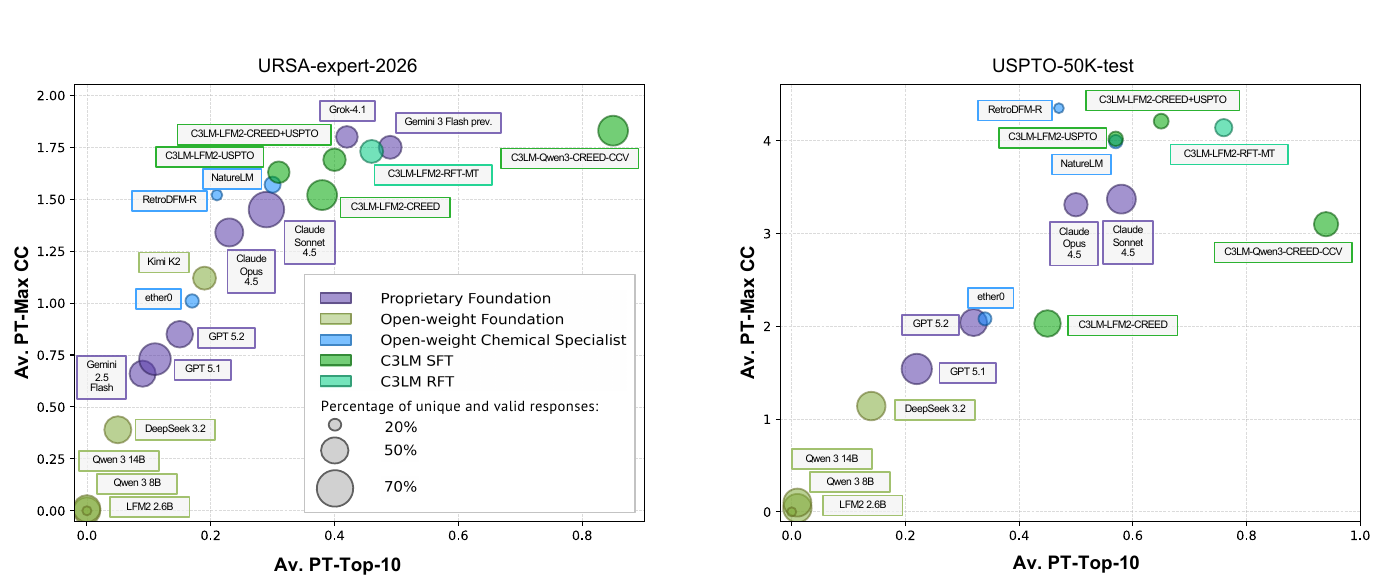}
  \caption{
  LLMs' performance in URSA benchmark on \expertset{} (left) and USPTO-50K-test (right) sets.
  Higher values along both axes indicate better performance (top-right is best).}
  \label{fig:Scatter}
\end{figure*}

\textbf{Training procedure: } On each training step, we utilize a total $524,288$ tokens context windows across all GPUs and pack \cite{brown2020packing} multiple training sequences into each GPU available context window. 

\textbf{Training Data Preprocessing:} Similarly to other chemical language models \cite{livne2024nacho, pei2023biot5}, we extend the base vocabulary with SMILES format \cite{weininger1988smiles} specific tokens; this is aimed at isolating chemical tokens from natural-language tokens and providing a consistent representation of SMILES entities. During the training, we tokenize SMILES into specialized tokens always in model outputs, and with $0.5$ probability for user input; we adopt this strategy to align LFM2 base checkpoint chemical knowledge and new tokens. To improve chemical generalization, we augment SMILES entities during training by applying non-canonical random traversal. Examples of training prompts can be found in \autoref{app:templates}.

\begin{table*}[th!]
\centering
\begin{tabular}{@{}l|c|cccc|c|cccc@{}}
\toprule
\multirow{2}{*}{\textbf{Model}} &
\multicolumn{5}{c|}{\textbf{URSA-expert-2026}} &
\multicolumn{5}{c}{\textbf{USPTO-50K-test}} \\
\cline{2-6}\cline{7-11}
& \multirow{2}{*}{\textbf{Unique}} & \multirow{2}{*}{\textbf{Max}} & \multicolumn{3}{c|}{\textbf{Av. PT-Top-K CC}}
& \multirow{2}{*}{\textbf{Unique}} & \multirow{2}{*}{\textbf{Max}} & \multicolumn{3}{c}{\textbf{Av. PT-Top-K CC}} \\
&  &  & \textbf{@3} & \textbf{@5} & \textbf{@10}
&  &  & \textbf{@3} & \textbf{@5} & \textbf{@10} \\
\midrule
\multicolumn{11}{c}{\textit{Proprietary Foundation Models}} \\
\midrule
Grok-4.1                   & 43\% & 1.75 & \textbf{1.29} & \textbf{0.93} & \textbf{0.49} & --   & --   & --   & --   & --   \\
Gemini 2.5 Flash           & 50\% & 0.66 & 0.30 & 0.18 & 0.09 & --   & --   & --   & --   & --   \\
Gemini 3 Flash preview     & 40\% & \textbf{1.80} & 1.21 & 0.83 & 0.42 & --   & --   & --   & --   & --   \\
GPT 5.1                    & 62\% & 0.73 & 0.35 & 0.22 & 0.11 & 59\% & 1.54 & 0.72 & 0.45 & 0.22 \\
GPT 5.2                    & 51\% & 0.85 & 0.47 & 0.29 & 0.15 & 52\% & 2.04 & 1.01 & 0.63 & 0.32 \\
Claude 4.5 Sonnet          & 70\% & 1.45 & 0.88 & 0.58 & 0.29 & 56\% & 3.37 & 1.81 & 1.15 & 0.58 \\
Claude 4.5 Opus            & 54\% & 1.34 & 0.73 & 0.46 & 0.23 & 44\% & 3.31 & 1.62 & 1.00 & 0.50 \\
\midrule
\multicolumn{11}{c}{\textit{Open-weight Foundation Models}} \\
\midrule
DeepSeek 3.2               & 52\% & 0.39 & 0.16 & 0.10 & 0.05 & 55\% & 1.14 & 0.46 & 0.27 & 0.14 \\
Qwen3 8B \                 & 52\% & 0.00 & 0.00 & 0.00 & 0.00 & 55\% & 0.04 & 0.03 & 0.02 & 0.01 \\
Qwen3 14B \                & 52\% & 0.01 & 0.00 & 0.00 & 0.00 & 54\% & 0.10 & 0.03 & 0.02 & 0.01 \\
Kimi K2                    & 43\% & 1.12 & 0.62 & 0.38 & 0.19 & --   & --   & --   & --   & --   \\
LFM2 2.6B                  & 0\%  & 0.00 & 0.00 & 0.00 & 0.00 & 0\%  & 0.00 & 0.00 & 0.00 & 0.00 \\
\midrule
\multicolumn{11}{c}{\textit{Open-weight Chemical Specialist Models}} \\
\midrule
ether0                     & 20\% & 1.01 & 0.55 & 0.35 & 0.17 & 19\% & 2.08 & 1.07 & 0.67 & 0.34 \\
NatureLM                   & 27\% & 1.57 & 0.97 & 0.61 & 0.30 & 20\% & 3.99 & 1.85 & 1.14 & 0.57 \\
RetroDFM-R                 & 12\% & 1.52 & 0.70 & 0.43 & 0.21 & 8\%  & \textbf{4.35} & 1.56 & 0.94 & 0.47 \\
\midrule
\multicolumn{11}{c}{\textit{\ourmodel{}}, Supervised Fine-Tuning} \\
\midrule
\ourmodel{}-LFM2-USPTO         & 40\% & 1.63 & 0.96 & 0.61 & 0.31 & 23\% & 4.02 & 1.83 & 1.13 & 0.57 \\ 
\ourmodel{}-LFM2-CREED         & 58\% & 1.52 & 1.05 & 0.75 & 0.38 & 51\% & 2.03 & 1.25 & 0.87 & 0.45 \\
\ourmodel{}-LFM2-CREED+USPTO         & 42\% & 1.69 & 1.14 & 0.78 & 0.40 & 24\% & \underline{4.21} & \underline{2.06} & \underline{1.29} & \underline{0.65} \\ 
\midrule
\multicolumn{11}{c}{\textit{\ourmodel{}}, Reinforcement Learning Fine-Tuning} \\
\midrule
\ourmodel{}-LFM2-RFT-MA         & 42\% & 1.72 & 1.14 & 0.78 & 0.40 & 26\% & 4.06 & 2.02 & 1.28 & 0.64 \\
\ourmodel{}-LFM2-RFT-MT         & 39\% & \underline{1.79} & \underline{1.22} & \underline{0.85} & \underline{0.43} & 23\% & 4.16 & \textbf{2.13} & \textbf{1.35} & \textbf{0.68} \\
\end{tabular}

\caption{Single-step retrosynthesis baselines on plausibility-based evaluation. \textbf{Unique}: fraction of unique valid reactant sets among samples. \textbf{Max}: per-target maximum ChemCensor score averaged over targets. \textbf{Av.\ PT-Top-K CC}: per-target average ChemCensor score over top-K unique predictions. We highlight best values with \textbf{bold} and second best with \underline{underline}.}
\label{tab:chemcensor_benchmark_results}
\end{table*}

\subsection{\ourmodel{} Reinforcement Learning Fine-Tuning}
\label{sec:rl_train_details}

\textbf{Training procedure: } Online Reinforcement Learning Fine-Tuning (RFT) of the \textit{no-reasoning} \ourmodel{} (\traindataset{}+USPTO-full) model was performed using single-reward Group Relative Policy Optimization (GRPO) \cite{shao2024deepseekmathpushinglimitsmathematical}. RFT used the same \traindataset{} training split as SFT with a GRPO group size of $8$, sampling temperature of $1$, and KL-regularization weight of $0.1$. The GRPO reward function is defined as a weighted sum combination of 3 reward components. Each component assigns a reward in the range $(-1, +1)$. First, the thinking-format component (weight $0.1$) verifies if the generated completion is correctly formatted, by returning $1$ for keeping the thinking in thinking tags, and $-1$ otherwise. Next, the molecular-syntax component (weight $0.5$) verifies if the answer is a valid SMILES string. Finally, the retrosynthesis correctness component (weight $1$) assesses if the generated answer is a valid reactant. For this, we use 2 different variants and train a separate model for each variant.

\textbf{\traindataset{} Multiple Answer (MA) Exact Matching: } Applies a binary reward of $+1$ if the generated reactants exist in the exhaustive enumerated list of \traindataset{} and $-1$ if not.

\textbf{Molecular Transformer Reward: } Applies a binary reward based on agreement with Molecular Transformer (MT) \cite{schwaller2019moltransformer}. The generated reactants from our model are passed to MT and a score of $+1$ is assigned if the MT-predicted product matches the original input to our model and $-1$ if not. MT was used as the forward synthesis reward function in ether0~\cite{narayanan2025ether0} and the same underlying model weights were used (MT trained to $40,000$ steps on USPTO480K \cite{jin_2017_uspto480k}).

Additionally, for the sake of end-user's practical utility and separately from the main experimental setup, the best C3LMs after SFT were subjected to RFT using CC as a reward function (see \autoref{app:ablation_studies}). For a generated reaction, the reward is the chemical plausibility between $0$ and $5$, scaled to $(0,+1)$. A reward of $-1$ is assigned when the generated reactant SMILES is invalid.

\subsection{Evaluation Protocol}
\label{sec:eval_protocol}

We perform a benchmark of all models from Sec. \ref{sec:baselines} and \ourmodel{}s on the proposed \expertset{} and revised USPTO-50K-test sets. For each product, we generate $15$ independent responses from each model within $15$ prompts (1 prompt to get 1 reaction) given in the \autoref{app:templates}.

All models were evaluated without external tools (no code execution or internet use) to isolate the core model capability. Reasoning was enabled in all models that support it with the default reasoning budget. We used a few-shot prompting method with five examples. The examples were chosen from the training set of the USPTO-full dataset at random for each data item in the test sets. The queries in the few-shot examples and the final question were sampled at random from a set of templates provided in the Appendix~\ref{app:templates}. For models released with task-specific retrosynthesis prompts by their authors, we used those prompts instead. 

We do not report USPTO-50K-test results for Grok-4.1, Gemini 2.5 Flash, Gemini 3 Flash preview and Kimi K2, because API request limits and/or long generation times prevented us from obtaining the full set of required completions on this larger dataset with 15 repetitions per example to reliably estimate the proposed metrics. Our proposed URSA-expert-2026 benchmark is a more reliable estimate of the generalizability of such LLMs that are usually prone to USPTO-50K-test set contamination. However, to have consistent side-by-side benchmarking, we have derived 10\% diverse subset of the USPTO-50K-test (the USPTO-50K-test-mini) to benchmark all models, including slow ones. These supplementary results can be found in \autoref{app:uspto_50k_sample}.

\section{Main Findings and Discussion}
\label{sec:main_results}

Table~\ref{tab:chemcensor_benchmark_results} reports \benchmark{} performance across LLM baselines and key C3LMs in the basic benchmark version (ChemCensor 0.5.2-U2). Other representations of these results are given as bar charts in \autoref{fig:chemcensor_categ_by_model} and as bubble plots in \autoref{fig:Scatter}. Overall, the majority of \textit{Proprietary Foundation} and \textit{Open-weight Chemical Specialist} language models generate a large fraction of valid and unique reactants and achieve strong ChemCensor scores on the USPTO-50K-test split (see the results on the shortened set, $N = 497$, for all models in \autoref{app:uspto_50k_sample}). However, \expertset{} appears to be a considerably harder setting: performance drops sharply for nearly all models. Surprisingly, \textit{Open-Weight Foundation} language models (with the exception of Kimi K2) tend to perform markedly worse: several models fail to produce valid and/or unique reactants, and even when they do, their predictions often receive low CC scores.

\begin{figure*}[ht]
  \centering

  \begin{minipage}{1\textwidth}
    \centering
    \includegraphics[width=0.36\textwidth]{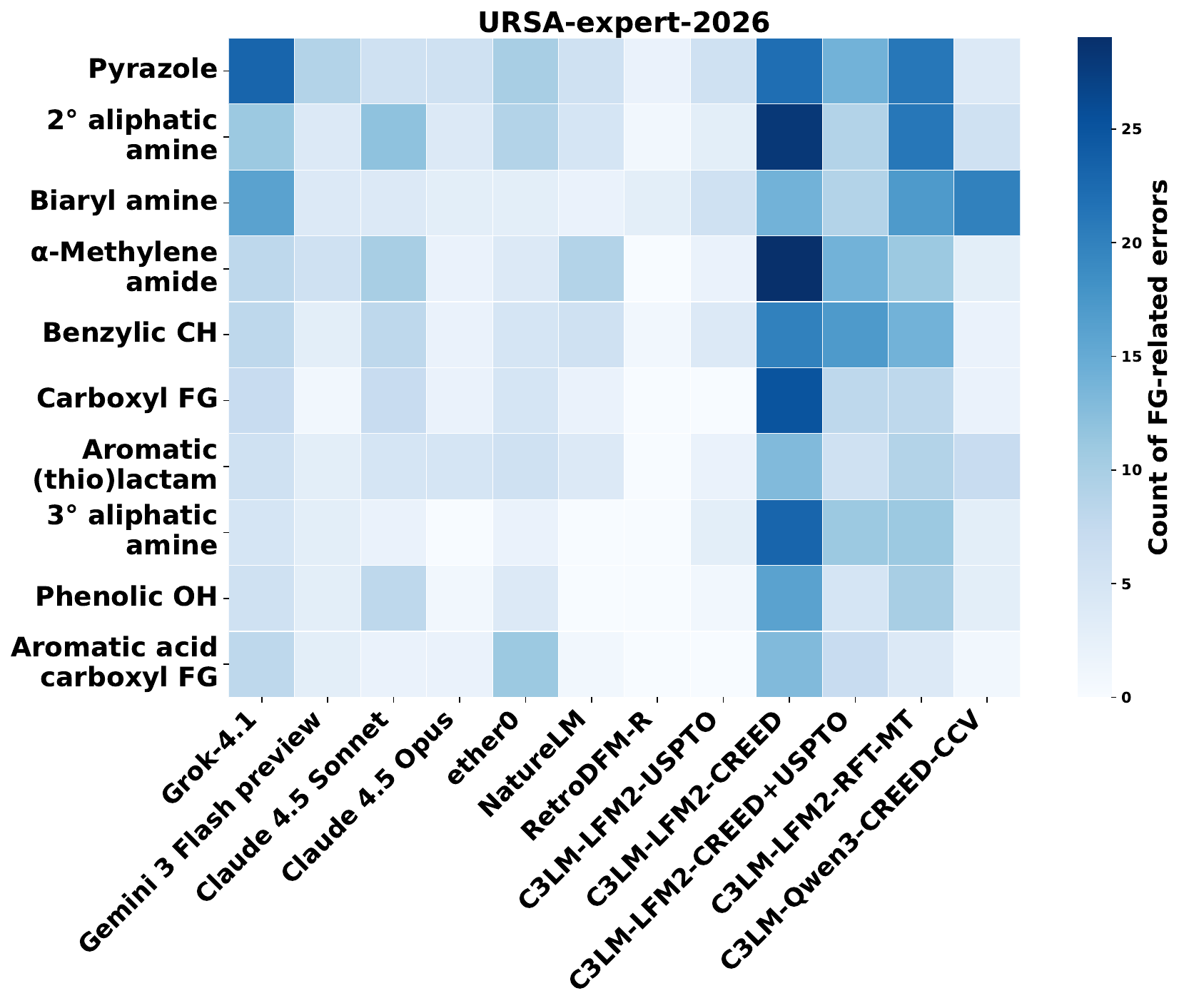}\hspace{1.3cm}
    \includegraphics[width=0.47\textwidth]{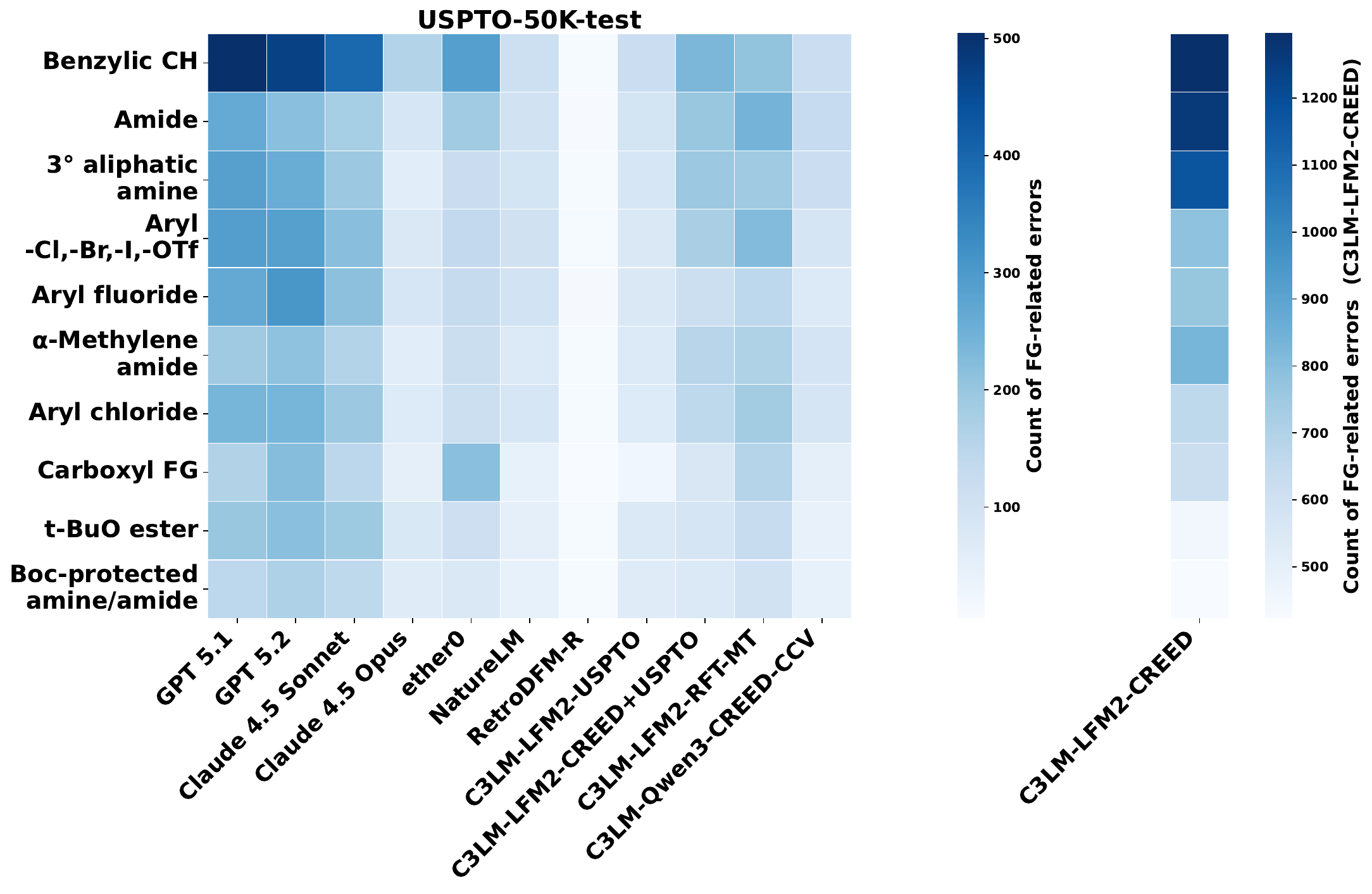}
  \end{minipage}
  \caption{FG–related error profiles across models on test sets. The heat map shows the count of CC FG incompatibility errors across the best $10$ models by \textbf{Av.~PT-Top-10 CC} (see \autoref{tab:chemcensor_benchmark_results}), darker cells indicate a higher error number. See FGs in \autoref{app:fg_smarts}.}
  \label{fig:fg_heatmap_by_model}
\end{figure*}

The ablation study shows that CREED, which is larger and more diverse than USPTO-full, leads to higher-scoring outputs on the OOD URSA-expert-2026 set than if a model is trained on USPTO-full alone. Adding USPTO-full to training reduces uniqueness, although it boosts performance on both benchmarking sets. This might be explained by the fact that USPTO-full mainly contains one set of reactants per product. The benefits of using CREED are even clearer when Pistachio, a larger reference set, is added to the ChemCensor relational DB (see \autoref{app:chemcensor_u2p2}). This finding supports the community's need for publicly available large datasets of reactions. We further show that training on the proposed large, diverse CREED enables models to reach the performance of all baselines. In particular, our C3LM, initialized from the \texttt{LFM2-2.6B}, which has one of the lowest baseline CC scores, achieves the best or 2nd-best results on both benchmarking sets after SFT and RFT.

\autoref{fig:chemcensor_categ_by_model} highlights differences in the diversity of output reactions across the models, rather than CC values, and shows the mechanics of CC-based evaluation. While red bars indicate reactions failed to align any \textit{synthetic precedent}, yellow ones show those with plausible RC, but failed the RC/FGs intercompatibility check. The usage of CREED for LLMs training is proven for both diverse (low grey bars) and plausible (high green bars) reactions outcomes. CREED also allows for a broader transition zone (yellow) due to higher diversity of reactions, however, not sufficiently supported by RC/FGs intercompatibility precedents. This transition might be positively resolved by using the C3LMs trained on CREED-CCV. While methodologically it would not be correct to compare these models to others in the main experimental setup due to the circularity of reward inherited from the CREED-CCV design, the practical utility of these models is supported by a significantly higher percentage of outcome reactions passing CC, therefore having stronger support from synthetic precedents (for more details see \autoref{app:ablation_studies}).  Interestingly, the models fail to satisfy RC/FGs intercompatibility criteria in different ways. As the heat map in \autoref{fig:fg_heatmap_by_model} shows, some models can self-confidently oversample reactions failing to handle aryl halides, various amines and other FGs selectively. Expectedly, \texttt{C3LM-LFM2-CREED} makes more RC/FGs errors than \texttt{C3LM-LFM2-USPTO}, since the relational database of the main ChemCensor version is derived from USPTO-full.

Although LLMs are promising SSRS tools, the baseline conventional SSRS models are still superior from both output plausibility (in terms of ChemCensor metrics, see \autoref{app:conventional_models}) and cost perspective. The challenge for LLMs to reach the Pareto frontier in solving the SSRS task is a matter of future research and significantly improved models.

\section{Conclusion}
\label{sec:conclusion}

We propose \benchmark{}, a plausibility-based metric for single-step retrosynthesis that utilizes synthetic precedents as the source of chemistry-aware constraints beyond exact-match Top-K accuracy. We release large and diverse \traindataset{} ($\sim$$22.7M$ reactions) and \expertset{}, an expert-annotated OOD benchmark of $100$ novel TMs.

Benchmarking a broad set of LMs shows that performance on the USPTO-50K-test does not reliably transfer to \expertset{}, which is substantially harder. Fine-tuning on \traindataset{} is highly effective: proposed C3LMs achieve best or second-best \benchmark{} scores on both benchmarks, while compared to general-purpose foundation and chemical specialist baselines.

\section{Code and Data Availability}
\label{sec:code_availability}

The following artifacts of this work are publicly available:
\begin{itemize}[nosep, topsep=0pt]
    \item ChemCensor source code -- \url{https://github.com/insilicomedicine/ChemCensor}.
    \item CREED -- \url{https://huggingface.co/datasets/insilicomedicine/CREED}.
    \item URSA-expert-2026 and curated USPTO-50K-test -- \url{https://huggingface.co/datasets/insilicomedicine/URSA-benchmarking-sets}
    \item C3LM -- \url{https://huggingface.co/insilicomedicine/C3LM}
\end{itemize}

\section*{Impact Statement}
This paper proposes ChemCensor, an evaluation metric and benchmarking protocol for single-step retrosynthesis, and introduces supporting datasets and models. If adopted, these tools could improve the reliability and reproducibility of research claims in computer-aided synthesis planning and help practitioners compare systems in a way that better reflects chemical plausibility. In the drug-discovery setting, more faithful evaluation reduces wasted effort on implausible routes and accelerates iteration by prioritizing synthetically feasible designs.

Potential negative impacts stem from dual use. Improved retrosynthesis models and benchmarking may potentially lower barriers to designing synthesis routes for harmful or controlled substances, and they may enable misuse if deployed without safeguards. The suggested CREED dataset was filtered by medicinal chemistry patterns and decontaminated to alleviate such risks, with the only focus on safe drug-like molecules as opposed to harmful substances. 

We also mitigate such risks by focusing the contribution on evaluation methodology, emphasizing out-of-domain and plausibility-aware assessment rather than providing an end-to-end automated synthesis capability, and by documenting dataset construction, decontamination, and limitations to support responsible use. We encourage future work to incorporate access controls and misuse-aware filtering when applying the resulting models in high-risk settings, and to audit benchmark coverage and biases as reaction data and model capabilities evolve.
\section*{Acknowledgments}
\label{sec:acknowledgments}

We thank the reviewers of ICML~2026 for their thoughtful and constructive feedback. Their comments motivated several substantial additions to the appendices of this work, which, together, meaningfully strengthened the rigor and practical utility of the paper.

\bibliography{bibliography}
\bibliographystyle{icml2026}


\newpage
\appendix
\onecolumn

\section{C3LM Inventory}
\label{app:model_inventory}

This appendix provides a complete inventory of the C3LM (Chemistry Constraint–Consistent Language Model) family of models evaluated in this work. The C3LM family comprises two branches built upon different base checkpoints: LFM2 2.6B and Qwen3 8B. Within each branch, individual models differ along three orthogonal axes, allowing us to systematically isolate the contribution of each design choice. Proprietary advanced reasoning traces were designed to algorithmically resemble the way chemists solve the reaction verification task, including decomposition of the product into meaningful parts (including functional groups, reaction centers, their reactivity from multiple aspects like electrophilicity, nucleophilicity, oxidative/reductive potential e.t.c), drafting the synthesis hypothesis, checking them and making step back if the hypothesis was chemically wrong or plausible.

\begin{table*}[th!]
\centering
\begin{tabular}{@{}l|lll@{}}
\toprule
\textbf{Model Name} & \textbf{Training data} & \textbf{Reasoning} & \textbf{Reward} \\
\midrule
\multicolumn{4}{c}{\textit{C3LM (LFM2 base model)}} \\ 
\midrule
C3LM-LFM2-USPTO      & USPTO         & Basic    & -- \\
C3LM-LFM2-CREED      & CREED         & Basic    & -- \\
C3LM-LFM2-CREED+USPTO      & CREED + USPTO & Basic    & -- \\
C3LM-LFM2-RFT-MT      & CREED + USPTO & Basic & Molecular Transformer \\
C3LM-LFM2-RFT-MA      & CREED + USPTO & Basic & Multi Answer Ground Truth \\
\midrule
C3LM-LFM2-CREED-CCV+USPTO      & CREED-CCV + USPTO & Basic    & -- \\
C3LM-LFM2-CREED-CCV+USPTO-AR      & CREED-CCV + USPTO & Advanced & -- \\
C3LM-LFM2-AR-RFT-CC      & CREED-CCV + USPTO & Advanced & ChemCensor \\
\midrule
\multicolumn{4}{c}{\textit{C3LM (Qwen3 base model)}} \\ 
\midrule
C3LM-Qwen3-CREED-CCV     & CREED-CCV         & Basic    & -- \\
C3LM-Qwen3-CREED-CCV+USPTO     & CREED-CCV + USPTO & Basic    & -- \\
C3LM-Qwen3-CREED-CCV+USPTO-AR     & CREED-CCV + USPTO & Advanced & -- \\
C3LM-Qwen3-RFT-CC     & CREED-CCV + USPTO & Basic & ChemCensor \\
C3LM-Qwen3-RFT-MA     & CREED-CCV + USPTO & Basic & Multi Answer Ground Truth \\
C3LM-Qwen3-RFT-MT    & CREED-CCV + USPTO & Basic & Molecular Transformer \\

\bottomrule
\end{tabular}
\caption{Model Inventory.}
\label{tab:ablation_model_inventory}
\end{table*}

\section{Glossary}
\label{app:gloassary}

\begin{description}

\item[\textbf{Target molecule}]
The target molecule is the desired chemical compound that represents the ultimate goal of synthesis planning. It is the molecule for which the system generates or evaluates synthetic routes, and it serves as the starting point for retrosynthetic disconnection, working backward from the target to identify precursor molecules.

\item[\textbf{Reactants}]
Reactants are the chemical compounds that undergo transformation during a chemical reaction and whose atoms are directly incorporated into the product structure. Technically, reactants are distinguished from reagents by the presence of atom mapping -- atoms in reactants have corresponding mapped atoms in the product(s).

\item[\textbf{Reagents}]
Reagents are chemical compounds that participate in a chemical reaction but whose atoms are not directly incorporated into the product structure. Reagents typically facilitate or enable the transformation (e.g., catalysts, bases, acids, solvents with reactive roles) with no or only minor atom contribution to the final product. For benchmarking, it is reasonable to extend the reacting species with reagents, since they may influence the correctness of atom--atom mapping and thereby the reaction-center extraction process.

\item[\textbf{Starting materials}]
Starting materials are the initial chemical compounds from which a synthetic route begins. They are molecules that exist at the terminal nodes (leaves) of a retrosynthetic tree and are not produced by any reaction step within the route. Starting materials serve as the input chemicals for the synthesis and are expected to be commercially available or otherwise accessible within the reported synthetic methods.

\item[\textbf{Building blocks}]
Building blocks (CABBs) are commercially available chemical compounds that can be found in vendor datasets and purchased from them.

\item[\textbf{Single-step retrosynthesis model}]
An SSRS model predicts one or several retrosynthetic disconnections by mapping a target product molecule to a set of precursor reactants corresponding to a single reaction step. The model does not perform recursive planning or multi-step route construction, focusing instead on identifying chemically plausible reactants.

\item[\textbf{Multi-step retrosynthesis model}]
An MSRS model is a system that applies retrosynthetic transformations (typically recursively) to decompose a target molecule into commercially available or otherwise accessible starting materials through a sequence of reaction steps.

\item[\textbf{Reaction center (RC)}]
An RC is the set of atoms (dynamic atoms) in one or more reactant molecules and the product molecule that undergoes change during a chemical transformation, including atoms/bonds that are formed, broken, created, destroyed, or whose connectivity, bond order, formal charge, or hybridization state differs between reactants and products.

\item[\textbf{Chemical plausibility}]
Chemical plausibility reflects alignment of a reaction with core principles of organic synthesis (e.g., chemoselectivity, regioselectivity, stereoselectivity). Operationally, it can be reduced to chemoinformatic concepts such as reaction centers, functional groups, their occurrence, and compatibility rules, potentially augmented with conditions (solvents, temperature, catalysts, auxiliary reagents). In this system, plausibility is assessed by comparing the reaction center and functional-group context against a reference dataset of verified transformations. If the extracted reaction center is absent from the reference library, the reaction is considered implausible; similarly, functional groups never observed for that reaction center negate plausibility. When both reaction-center and functional-group context are supported by precedents, the reaction is considered chemically plausible.

\item[\textbf{Level of confidence}]
The nominal degree of chemical plausibility is estimated via discrete levels of confidence (LC), depending on which reaction-center representation is matched among verified transformations. Higher LC indicates that a more specific (larger-context) reaction-center definition is supported, correlating with higher nominal plausibility and representativeness in terms of synthetic precedents. The LC value is used as the reaction score in ChemCensor.

\item[\textbf{Functional groups (FGs)}]
FGs are structural motifs that determine chemical reactivity and properties. In the present system, functional groups are represented as SMARTS patterns \cite{smarts} that can be matched to molecular structures via substructure search. Functional-group context annotated for each reaction center helps determine which patterns are tolerable for a transformation, supporting chemical plausibility.

\item[\textbf{Functional group (FG) signature}]
An FG signature is the ensemble of FGs present in reactant/product molecules that are not affected by the transformation. For a given reaction center, the signature is constructed by aggregating synthetic precedents from the reference dataset.

\item[\textbf{ChemCensor}]
ChemCensor Score (CC Score) is a quantitative metric evaluating the chemical plausibility of a reaction and/or a retrosynthetic route by measuring the proportion of steps that pass plausibility validation, weighted by their confidence levels.

\item[\textbf{Reference dataset}]
A reference dataset is a collection of verified reaction transformations extracted from sources including patents (e.g., USPTO), articles, preprints, and ELNs. It may include metadata such as conditions and yield. In this system, the reference dataset is used to validate analyzed reactions via reaction-center matching and functional-group signature comparison.

\item[\textbf{Synthetic precedent}]
A synthetic precedent is an elementary synthetic fact of a successful chemical reaction recorded in a reference dataset.

\end{description}
\section{When Single Answer Is Not Enough: Examples}
\label{app:when_single_answer_is_not_enough}

\begin{figure*}[!h]
  \centering
  \includegraphics[
    clip,
    trim=0 450 550 70,
    page=1,
    scale=0.95,
  ]{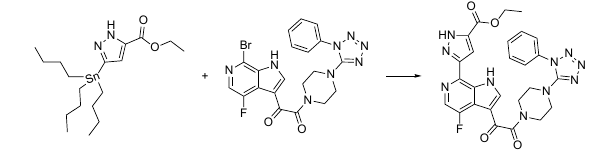}
  \caption{The ground truth for the example \#1 from the USPTO-50K-test set.
}
  \label{fig:ex1_gt}
\end{figure*}

To illustrate the weakness of ground truth-based metrics, the following target molecule (example \#1) from the USPTO-50K-test set was used to predict the possible reactant sets using the RetroKNN model \cite{xie2023retrosknn} via Syntheseus interface \cite{maziarz2025syntheseus}. In the USPTO-50K-test set, there is only one reaction, the ground truth, which is provided for model performance estimation. The reaction represents bromo-Stille cross-coupling C-C bond formation (\autoref{fig:ex1_gt}).

\begin{figure}[H]
  \centering
  \includegraphics[
    clip,
    trim=0 180 550 70,
    page=2,
    scale=0.95,
  ]{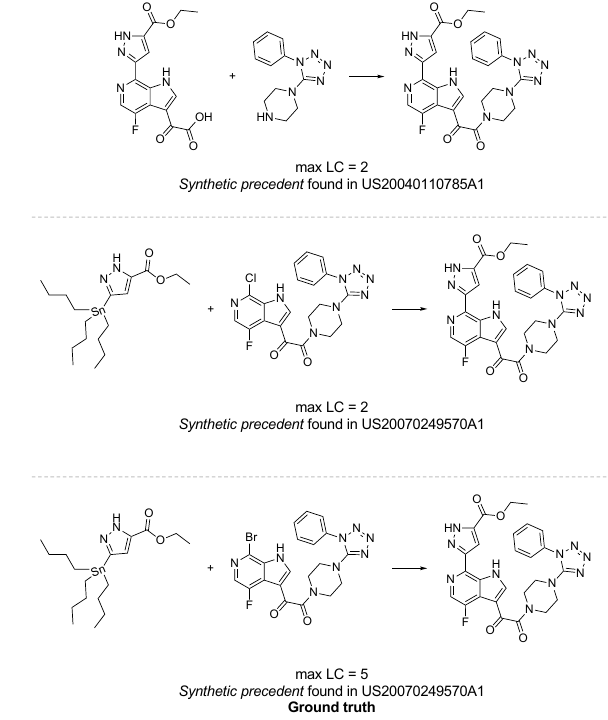}
  \caption{Scoring three different outcomes predicted by RetroKNN for example \#1.
}
  \label{fig:ex1}
\end{figure}

Besides the reaction shown in \autoref{fig:ex1_gt}, RetroKNN also predicts at least two other reactions since, obviously, other disconnections for the molecule are available. The first reaction proceeds as an amide synthesis, and the second one is also based on Stille cross-coupling, but for a chlorine atom. While these predictions are considered irrelevant for ground truth-based metrics, they are scored as chemically plausible by ChemCensor (see \autoref{fig:ex1}).

\begin{figure}[H]
  \centering
  \includegraphics[
    clip,
    trim=0 450 550 65,
    page=3,
    scale=0.95,
  ]{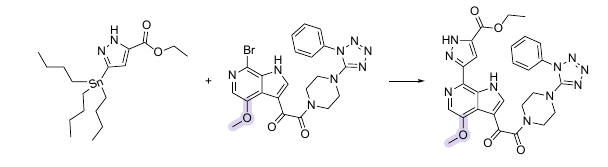}
  \caption{The additional reaction found in the patent application US20070249579A1, where the ground truth for example \#1 was extracted from.
}
  \label{fig:ex1_add}
\end{figure}

It is interesting to note that for the chloro-Stille cross-coupling, the algorithm provides the same patent application (US20070249579A1) as the ground truth. ChemCensor allows checking all reactions associated with the concrete reference document ID and if one checks other reactions present in the patent with the ground-truth reaction, the chloro-Stille coupling, very similar to the ground truth with a fluorine atom replaced with a methoxy group is found (\autoref{fig:ex1_add}). The Top-K accuracy-based benchmark rejects this reaction, because it is not the ground truth, while it differs from the ground truth only by a methoxy group non-participating in the transformation and halogen functionality.

\begin{figure}[H]
  \centering
  \includegraphics[
    clip,
    trim=0 450 550 70,
    page=4,
    scale=0.90,
  ]{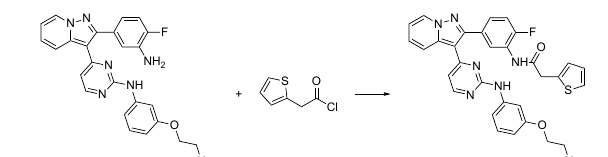}
  \caption{The ground truth for example \#2.
}
  \label{fig:ex2_gt}
\end{figure}

To additionally illustrate the difference between our scoring approach and commonly used metrics, another target molecule from the USPTO-50K-test, example \#2, was used to predict single-step retrosynthetic output within the LocalRetro model \cite{chen2021localretro}. The target molecule and its ground truth are shown in \autoref{fig:ex2_gt}.

\begin{figure}[H]
  \centering
  \includegraphics[
    clip,
    trim=0 180 500 70,
    page=5,
    scale=0.90,
  ]{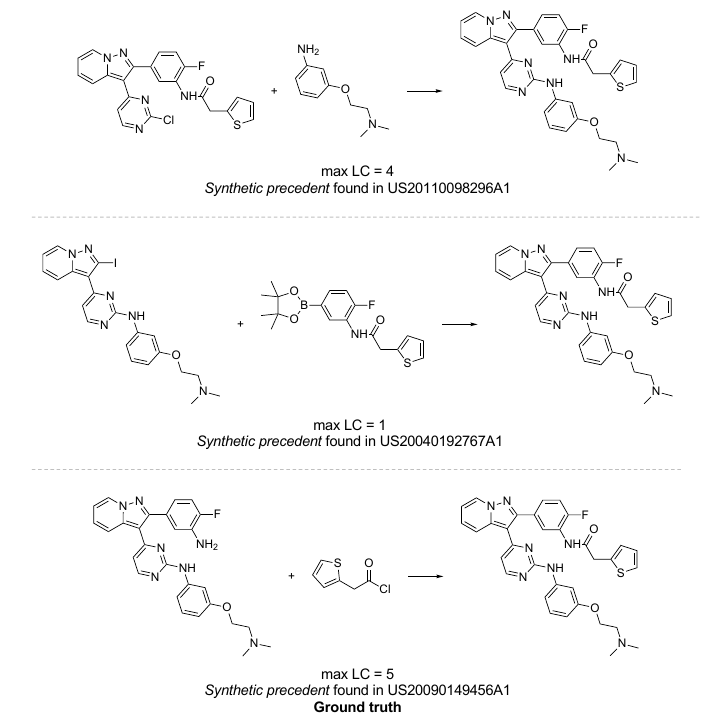}
  \caption{Scoring three different outcomes predicted by LocalRetro for example \#2.
}
  \label{fig:ex2}
\end{figure}

In addition to the reaction shown in \autoref{fig:ex2_gt}, LocalRetro also outputs at least two other transformations. The first reaction was obtained by the formation of a C-N amide bond between the core scaffold and the side chain. The second one represents iodo-Suzuki cross-coupling. Although these predictions are considered chemically plausible by ChemCensor, they are deemed irrelevant for the ground truth-based metrics (\autoref{fig:ex2}). Interestingly, for the second transformation, the ChemCensor Score is 1, highlighting the degree of exclusiveness for iodo-pyrazolopyridine substrate in the Suzuki reaction.

The examples provided illustrate the advantages of the present ChemCensor as a tool to assess SSRS models’ output. It is not biased to the ground truth concept and provides the user with a powerful engine for single reaction evaluation. Crucially, the failures shown above demonstrate a deeper methodological issue: Top-K accuracy linked to a specific benchmark set (such as USPTO-50K-test) accounts for only one "correct" reactant set (the one provided as the benchmark answer), even though multiple alternative reactants can be equally valid and should not be penalized. Strikingly, Top-K accuracy penalizes even reactants that are themselves reported in USPTO-full, the very corpus from which USPTO-50K-test is derived. This exposes the fundamental fallacy of the Top-K paradigm: it fails on previously reported reactions and would penalize truly novel but synthetically feasible disconnections by construction. ChemCensor, by contrast, accommodates multiple correct reactant sets by isolating the key reaction features (reaction center (RC) and functional groups (FGs))from the surrounding molecular context, rather than requiring an exact match of the entire reaction. Additional cases of reactions misinterpreted by Top-K accuracy on USPTO-50K-test but correctly handled by ChemCensor are presented in \autoref{tab:topk-critics}

\includepdf[
  pages=1,
  offset=0 -7mm,
  scale=0.92,
  trim={0mm 0mm 0mm 45mm},
  pagecommand={%
    \thispagestyle{plain}%
    \captionsetup{type=table}%
    \vspace*{0em}%
    \caption{Reactions misinterpreted by the Top-K Accuracy-based approach on the USPTO-50K-test benchmarking set and well-interpreted by ChemCensor}%
    \label{tab:topk-critics}%
    \vspace*{0em}%
  }
]{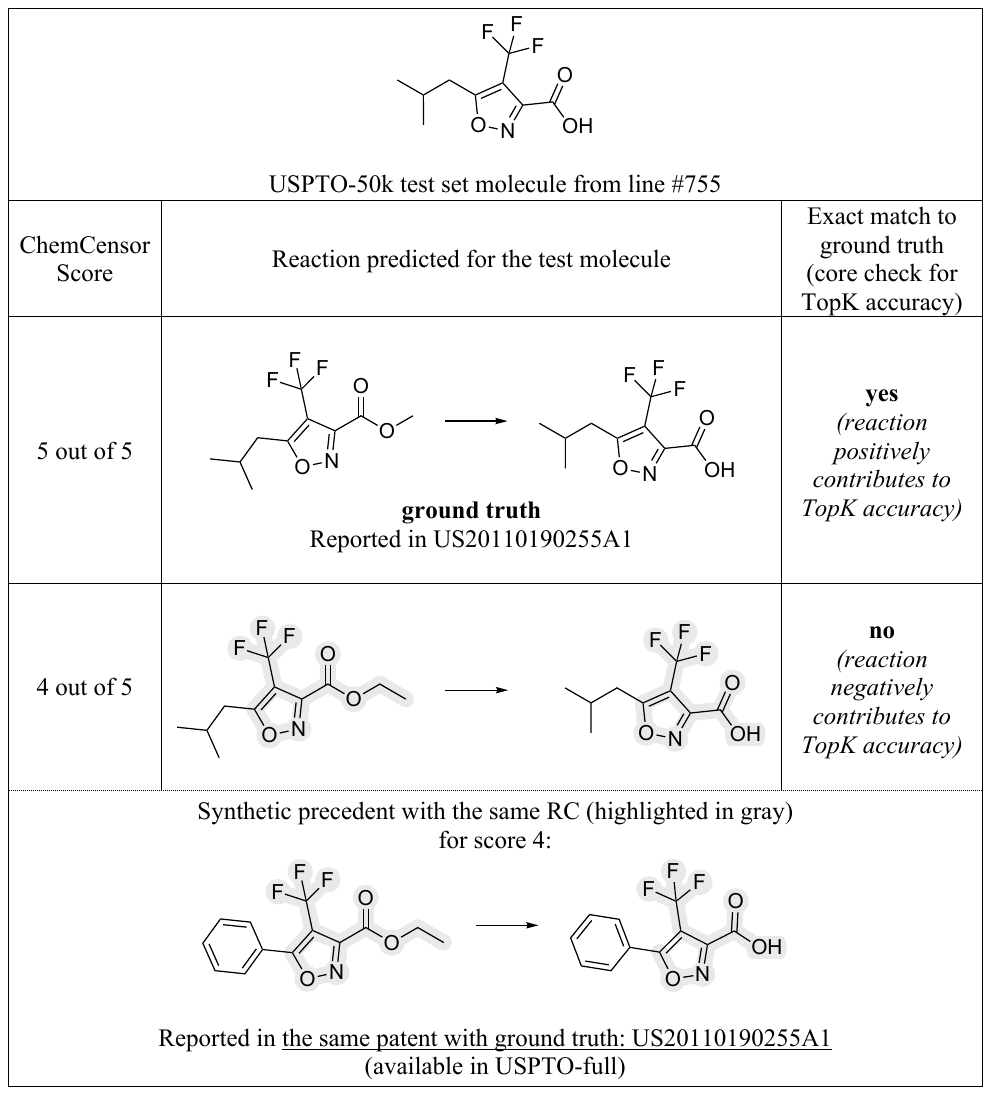}
\includepdf[
  pages=2-,
  scale=0.92,
  trim={0mm 0mm 0mm 20mm},
  pagecommand={%
    \thispagestyle{plain}%
    \captionsetup{type=table}%
    \caption*{Continuation of \autoref{tab:topk-critics}.}%
  }
]{figs/appendix/rev1-ans1.pdf}

\newpage
\section{ChemCensor Details}
\label{app:chemcensor_rc}

\begin{figure*}[ht]
  \centering
  \includegraphics[
    width=\linewidth,
    clip,
    trim=30 160 300 70,
  ]{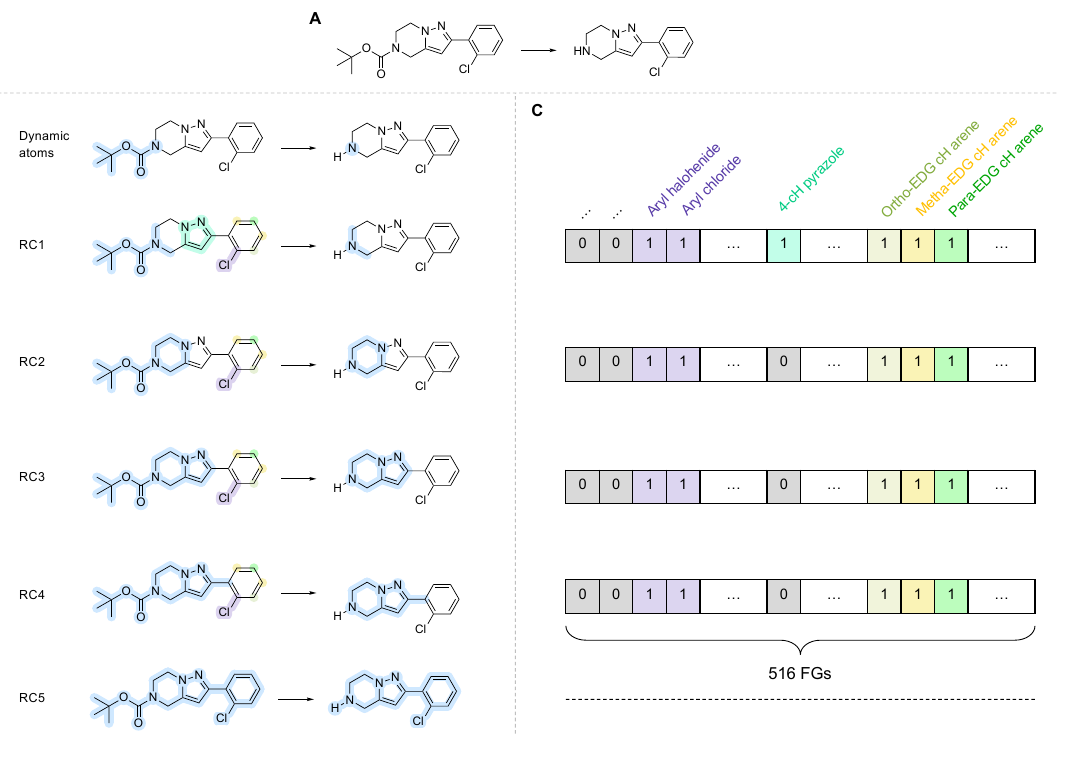}
  \caption{ChemCensor operands. \textbf{A}. Input reaction. \textbf{B}. RC\textbf{N}s (highlighted in blue). \textbf{C}. FG signatures (FGs are highlighted).
}
  \label{fig:RCFG}
\end{figure*}

ChemCensor estimates how closely a single reaction resembles a reference corpus of synthetic precedents and uses that resemblance as a proxy for chemical plausibility. Besides precedent estimation, it also performs basic checks, such as correct reaction syntax and SMILES validity. The pipeline standardizes the input, validates SMILES, and maps atoms between reactants and products (RxnMapper \cite{Schwaller2021rxnmapper} was used). Reactive atoms (atoms related to a dynamic part) are inferred from the mapped transform. Next, the processed reaction undergoes rigorous stereo validation: if there are any allowed stereo inversions (e.g., SN2 reactions), legitimate stereo-center emergence (a procedure for resolving stereoisomers), or stereo errors (inversions in the static part of the reaction). Also, to properly render regioselectivity of electrophilic aromatic substitutions, the separate module, SeAr annotator, determines if the reaction belongs to S\textsubscript{e}Ar class and annotates the functional group context surrounding the C-H center. Once processing is done, the reaction center extractor yields a nested family of centers (RC1–RC4) with increasing peripheral context. Each center is additionally annotated with non-participating functional groups located outside the reactive part of the transformation. Finally, the scorer compares the extracted centers with centers prepared in the same manner for the reference database of synthetic precedents. It accepts a center if its signature is contained in the reference (sub-signature match), with higher levels yielding higher scores, which implies the higher chemical plausibility. An exact canonical reaction hit can yield the top score. Additionally, SEAr scoring includes checking the SEAr context stored with each precedent reaction. The details on center extraction and FG annotation are provided below.

In the current version, there are five levels of reaction center annotation, denoted RC1, RC2, RC3, RC4 and RC5 corresponding to confidence levels and ChemCensor Scores 1, 2, 3, 4 and 5, respectively (see \autoref{fig:RCFG} \textbf{A}). The RC\textbf{N} is the representation of an RC, where \textbf{N} shows the portion of the structural context of an RC; the higher the \textbf{N}, the higher the context size (see \autoref{fig:RCFG} \textbf{B}):

\begin{itemize}[noitemsep]
    \item RC1 captures the minimal local environment of RC with the inclusion of dynamic atoms and atoms at topological distance of 1 from them. Also, atoms from predefined FGs (e.g., carbonyl group, nitrile, etc.) are included when at least one atom of a FG is dynamic.
    \item  RC2 includes atoms from RC1 with additional incorporation of the primary ring context (if present), stereochemical surroundings, and/or the atoms at topological distance of 2 around dynamic atoms.
    \item The next level, RC3, extends the atoms of RC2 by the addition of a fused-ring context around (if present) or the atoms at topological distance of 3 around dynamic atoms.
    \item Finally, RC4 is further extended by adding substituent context to aromatic systems and completing the predefined FGs, e.g., carbonyl group, nitrile, etc., or the atoms at topological distance of 4 from reactive atoms.
\end{itemize}

Scoring for S\textsubscript{e}Ar reactions also implies comparison of S\textsubscript{e}Ar functional group context surrounded by a C-H center with the context of the same center compiled for reported S\textsubscript{e}Ar reaction examples.

After extraction, each reaction center is annotated with functional groups outside the reaction center part (see \autoref{fig:RCFG} \textbf{B}). ChemCensor uses a curated database of 516 functional group patterns to cover basic chemical entities relevant to organic chemistry. Once matched, a non-reacting FG is considered tolerable during transformation if all the atoms of a FG are not overlapped with RC atoms. Besides that, FG signatures of centers derived from S\textsubscript{e}Ar reactions are additionally annotated with patterns describing different C-H patterns (e.g., presence of ortho-/para-EWG or EDG groups) which are normally not checked for centers from non-S\textsubscript{e}Ar transformations. This helps better evaluate the regioselectivity of S\textsubscript{e}Ar reactions without affecting other reaction classes by poorly reactive C-H patterns.

During the processing of the corpus of reference reaction data, an FG signature is prepared to store the cumulative ensemble of non-participating groups found for all reaction examples associated with the given RC. To expedite the handling and searching, each reaction center is treated as a canonicalized SMARTS string describing reacting patterns for both left and right sides of the reaction. The annotated FG signatures are encoded as numpy binary arrays in the form of 516-bit signatures (\autoref{fig:RCFG} \textbf{C}). Reaction centers and FG signatures prepared for the dataset of synthetic precedents are stored in a SQLite database together with the reaction examples from which they originate.

\newpage
\section{Reaction Examples and Detailed ChemCensor Output}
\label{app:detailed_chemcensor_output}

\begin{figure}[H]
  \centering
  \includegraphics[
    clip,
    trim=0 65 550 0,
    page=1,
    scale=0.9,
  ]{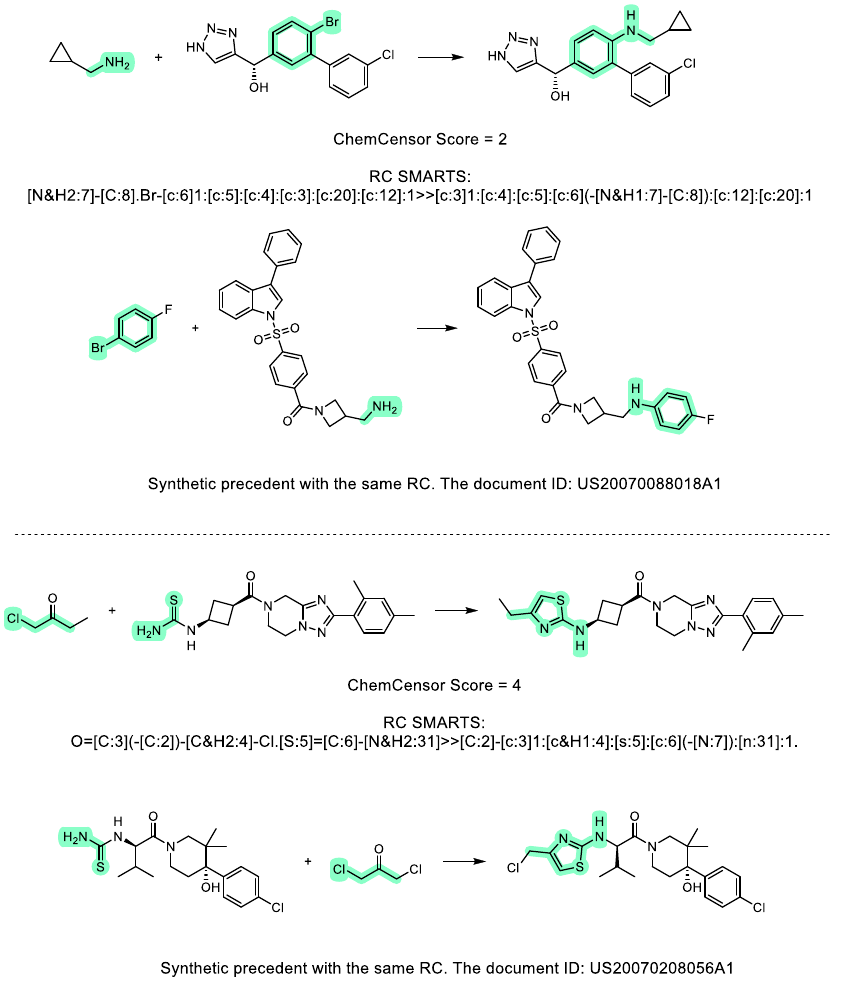}
  \caption{Example reactions predicted by the C3LM SFT model which passed ChemCensor.
}
  \label{fig:passed_ex}
\end{figure}

This section illustrates ChemCensor scoring and annotation for example reactions sampled from predictions of the C3LM SFT model, with reasoning for URSA-expert-2026. We picked 5 examples to highlight different pass and non-pass categories. In \autoref{fig:passed_ex}, two reactions that fully passed ChemCensor in terms of both the RC and FG combination precedence are outlined. For each valid reaction, the module provides extracted mapped SMARTS of the reaction center, reference document ID and the respective synthetic precedent example.

\FloatBarrier

\begin{figure}[H]
  \centering
  \includegraphics[
    clip,
    trim=0 240 550 0,
    page=2,
    scale=0.95,
  ]{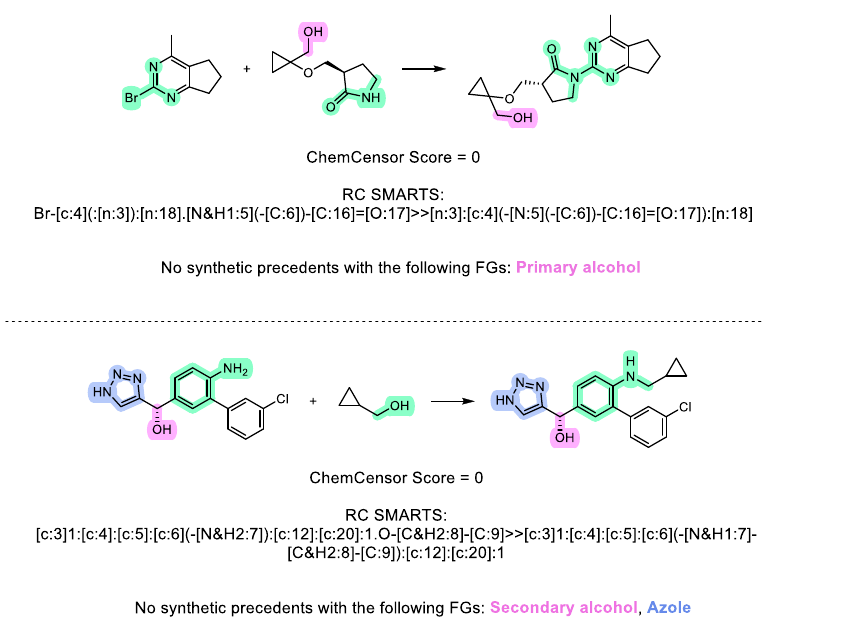}
  \caption{Example reactions which didn't pass ChemCensor due to functional group incompatibility.
}
  \label{fig:fg_failed_ex}
\end{figure}

If the input reaction is not reflected by the reference dataset of synthetic precedents in terms of its RC and FG context, ChemCensor provides a zero score and the reaction can be considered as chemically implausible. In that case, the module prints the exact reason for failure: (1) there are no precedents of RC representations or (2) the presence of specific non-participating functional groups is not reported for the found RC precedent. In the latter situation, ChemCensor returns the full list of functional groups that violate chemo-/regioselectivity knowledge extracted from the reference dataset (\autoref{fig:fg_failed_ex}).

\FloatBarrier

\begin{figure}[H]
  \centering
  \includegraphics[
    clip,
    trim=0 385 600 0,
    page=3,
    scale=0.95,
  ]{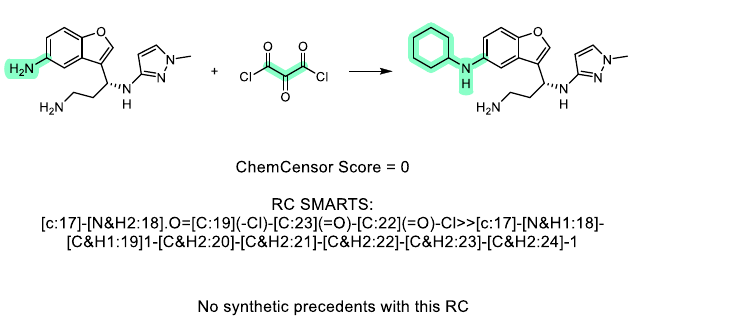}
  \caption{A reaction example for which RC representations are absent from the reference dataset.
}
  \label{fig:fg_no_rc_ex}
\end{figure}

Reactions for which reaction center representations are absent from the reference dataset are annotated with extracted mapped SMARTS only and a message of absence of synthetic precedents (\autoref{fig:fg_no_rc_ex}).

Beyond detecting the outright absence of precedents, ChemCensor can also discriminate between regio- and stereochemical outcomes of a given transformation. Representative examples are provided in \autoref{tab:regio} and \autoref{tab:stereo}, each containing both reactions that pass ChemCensor (matching the major or stereochemically correct product reported in precedents) and reactions that are penalized with a zero score (corresponding to minor regioisomers or unreported stereochemical configurations).

\includepdf[
  pages=1,
  offset=0 10mm,
  scale=0.80,
  pagecommand={%
    \thispagestyle{plain}%
    \captionsetup{type=table}%
    \vspace*{0em}%
    \caption{Examples of regioselectivity assessments by ChemCensor.}%
    \label{tab:regio}%
    \vspace*{0em}%
  }
]{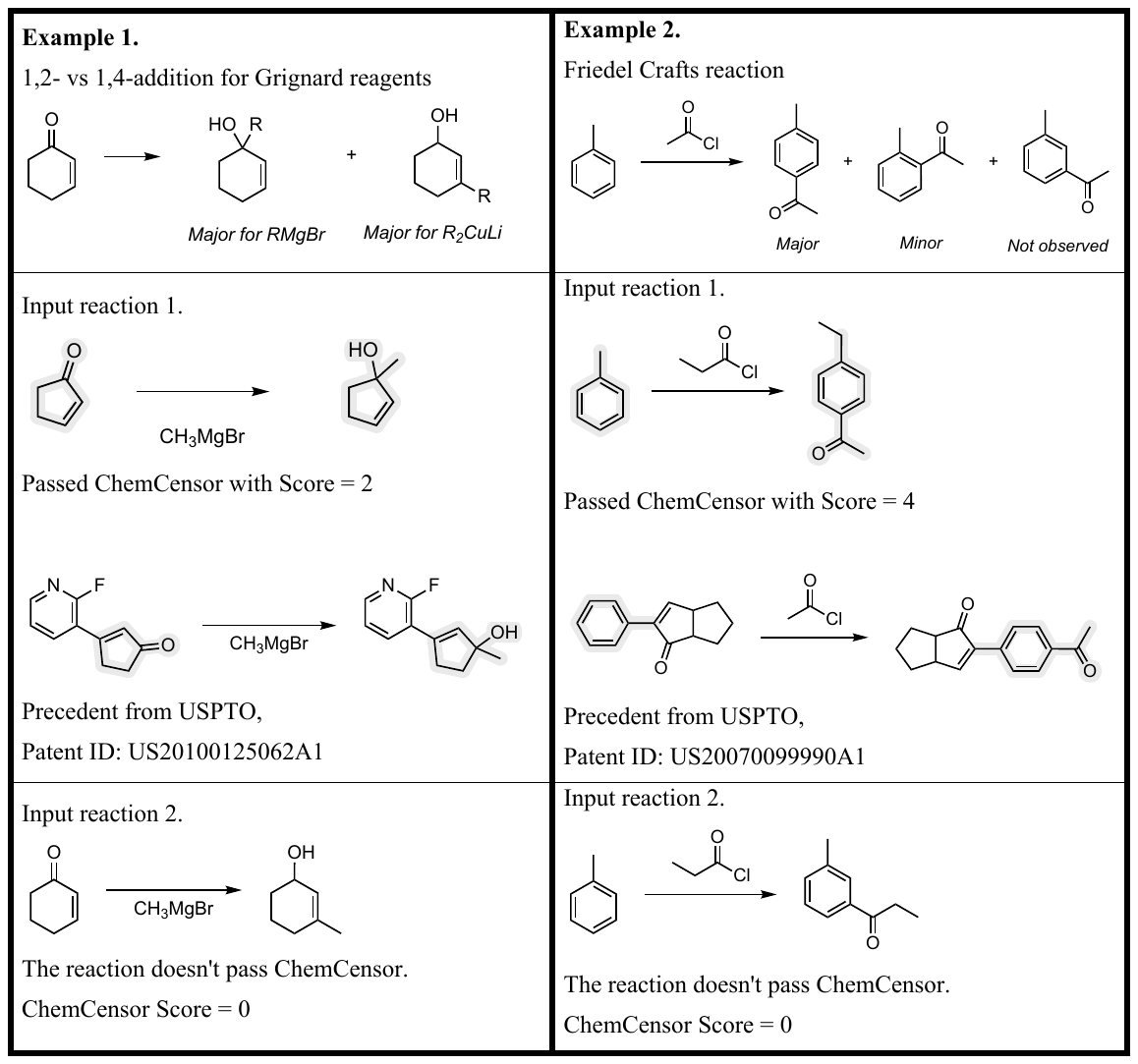}
\includepdf[
  pages=2,
  offset=0 10mm,
  scale=0.80,
  pagecommand={%
    \thispagestyle{plain}%
    \captionsetup{type=table}%
    \caption*{Continuation of ~\autoref{tab:regio}.}%
  }
]{figs/appendix/rev1-ans3.pdf}

\includepdf[
  pages=3,
  offset=0 10mm,
  scale=0.80,
  pagecommand={%
    \thispagestyle{plain}%
    \captionsetup{type=table}%
    \vspace*{0em}%
    \caption{Stereoselectivity assessments by ChemCensor exemplified by SN2 reactions.}%
    \label{tab:stereo}%
    \vspace*{0em}%
  }
]{figs/appendix/rev1-ans3.pdf}
\newpage
\section{Functional Groups With The Highest FG-related Error Rate}
\label{app:fg_smarts}

\begin{figure*}[ht]
  \centering
  \includegraphics[
    clip,
    trim=0 5 100 0,
    page=1,
    scale=0.55,
  ]{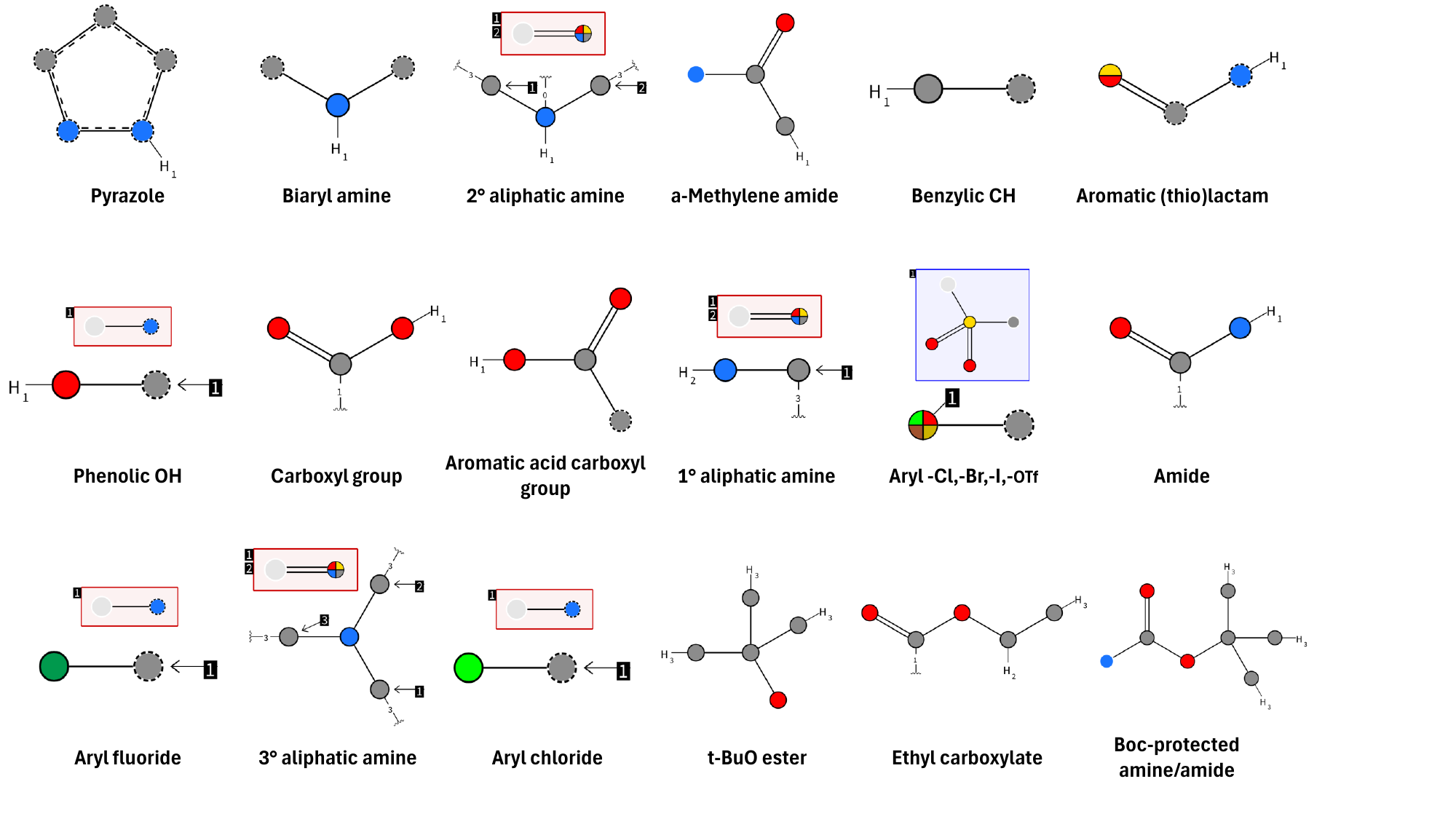}
  \caption{Visualized SMARTS patterns for FGs with the highest FG incompatibility error count across the Top-10 models by means of \textbf{Av. PT-Top-10 CC}. Individual figure preparation was performed in SMARTSview \cite{Ehrt2020SMARTSplus}.
}
  \label{fig:fig_fg_smarts}
\end{figure*}
\section{\traindataset{} Details}
\label{app:CREED_stats}

\paragraph{Public release.}
The CREED is released on Hugging Face (see \autoref{sec:code_availability}). We ship two parallel tables under the same logical layout, differing only in whether a per-reaction ChemCensor score is attached:
\begin{itemize}[noitemsep, topsep=2pt]
    \item \textbf{CREED} -- full candidate lists \emph{without} per-reaction ChemCensor scores, intended for users who only need product--reactant structure or wish to apply their own filtering.
    \item \textbf{CREED-CCV} -- ChemCensor-Verified subset (pipeline \texttt{chemcensor==0.5.2}); each candidate carries a numeric ChemCensor score, and only candidates with $\mathrm{CC}>0$ are retained.
\end{itemize}

Both tables follow a product-centric record layout: one row per product, with an ordered list of one-step disconnections (\texttt{reactant\_candidates}) attached as a struct field. Each candidate carries the product isomeric SMILES, the list of reactant/reagent isomeric SMILES, and (for CREED-CCV) the ChemCensor score. This layout makes the multi-solution nature of single-step retrosynthesis explicit at the storage level, in contrast to the one-row-per-reaction convention of USPTO-derived corpora.

\textbf{Summary statistics.}
\autoref{tab:creed_summary_stats} reports global statistics for both releases.

\begin{table}[H]
\centering
\small
\setlength{\tabcolsep}{8pt}
\renewcommand{\arraystretch}{1.15}
\begin{tabular}{@{}lrr@{}}
\toprule
\textbf{Statistic} & \textbf{CREED} & \textbf{CREED-CCV} \\
\midrule
\multicolumn{3}{@{}l@{}}{\textbf{Global}} \\
Unique products & 1{,}493{,}715 & 698{,}765 \\
Reaction candidates & 22{,}712{,}526 & 6{,}368{,}986 \\
Parquet shards & 8 & 4 \\
\midrule
\multicolumn{3}{@{}l@{}}{\textbf{Candidates per product}} \\
Min & 1 & 1 \\
Max & 152 & 2{,}364 \\
Mean & $\sim$15.21 & $\sim$9.11 \\
\bottomrule
\end{tabular}
\caption{Summary statistics of the two CREED releases. CREED contains all enumerated candidates from the virtual synthesis engine; CREED-CCV additionally requires a positive ChemCensor score.}
\label{tab:creed_summary_stats}
\end{table}

\paragraph{Splitting.}
For internal training and evaluation, we constructed train/validation/test partitions of CREED and CREED-CCV using random sampling in the 0.8/0.1/0.1 ratio, preserving the distribution of candidates per product and ensuring that each product appears in exactly one split.

\paragraph{Decontamination.}
The intersection between CREED-CCV and the publicly available USPTO-full dataset comprises $7{,}147$ reactions ($0.12\%$ of all reactions in CREED-CCV). Product molecules in CREED were verified to be disjoint from the target pools of both holdout sets used in the main paper: \expertset{} and USPTO-50K-test. To quantify any residual structural overlap, Tanimoto similarities (Morgan fingerprints, radius~$=2$, $2048$ bits) were computed between CREED product molecules and molecules from each holdout set.

For \expertset, the most similar CREED product has a Tanimoto coefficient of $0.523$, confirming that the holdout supports genuine out-of-distribution evaluation of models trained on CREED. For the USPTO-50K-test set, $30$ molecules show a similarity above $0.85$, of which $14$ have similarity exactly equal to $1.0$ and differ from the corresponding CREED entry only in stereochemistry. We retain these stereo-only collisions in CREED on the grounds that the disconnections themselves differ at the stereocenter level, but flag them here for transparency.

\begin{figure*}[h]
  \centering
  \includegraphics[
    width=\linewidth,
    clip,
    trim=0 200 50 70,
  ]{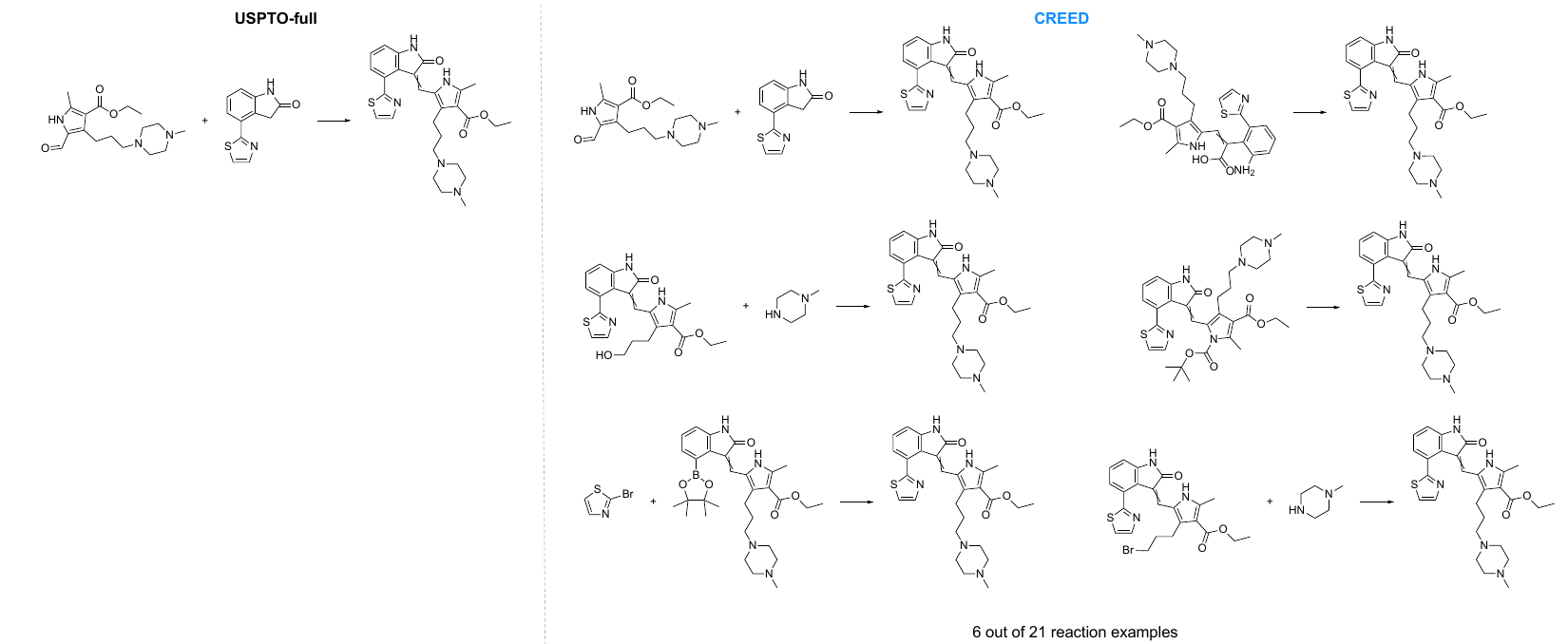}
  \caption{Reaction examples for a representative product molecule, drawn from USPTO-full (left) and CREED (right).}
  \label{fig:CREEDvsUSPTO1}
\end{figure*}

\begin{figure*}[h]
  \centering
  \includegraphics[
    width=0.9\linewidth,
    clip,
    trim=0 250 320 70,
  ]{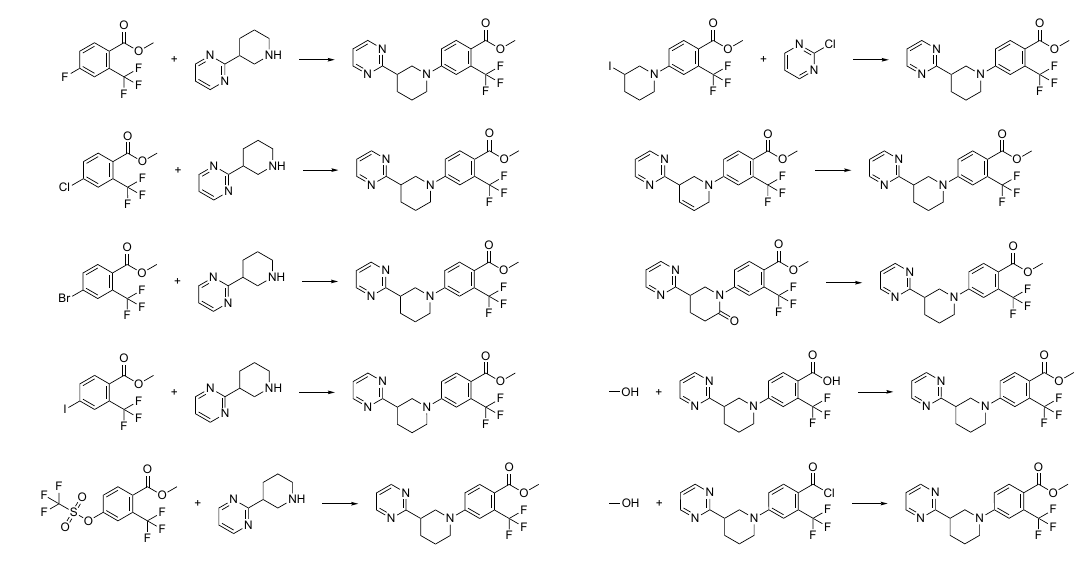}
  \caption{Examples of reactions from CREED for a product molecule that is not present in USPTO-full.}
  \label{fig:CREED10rxn}
\end{figure*}

\paragraph{Chemical capacity relative to USPTO-full.}
\autoref{fig:CREEDvsUSPTO1} illustrates the chemical capacity of CREED relative to USPTO-full for a representative product. Whereas USPTO-full typically provides only a small number of recorded transformations per product, CREED yields a substantially larger and more diverse set of plausibility-verified disconnections, reducing benchmark sparsity and enabling richer training and evaluation signals for a given target. \autoref{fig:CREED10rxn} shows a further example of disconnections from CREED for a product molecule that is absent from USPTO-full.

\section{\ourmodel{} Model Training Details}
\label{app:training_details}

\ourmodel{} is supervised fine-tuned on \traindataset{} using the \texttt{AdamW} optimizer with a learning rate of $1\times10^{-5}$, a linear warmup over the first $2{,}000$ steps, and a cosine learning-rate schedule thereafter.

\ourmodel{} Models were trained using GRPO with one of three verifiable reward functions. Each of the three RFT models shares all training hyperparameters. The prompts for the GRPO groups were sampled from the same CREED training set as SFT in groups of 16 with completions were generated with a temperature of 1.2. The number of updates per batch (number of iterations, $\mu$) was set to 1 and the KL penalty ($\beta$) was set to 0.05. The groups of 16 completions were sampled in batches of 8 unique prompts for a global batch size of 128. Each model was trained on a single node with 8 NVIDIA B200 GPUs. Deepspeed stage 3 \cite{rasley2020deepspeed} was used, with gradient checkpointing and BF16 precision. The AdamW optimizer was used for each model with a learning rate $1 \text{x} 10^{-6}$ at a constant schedule and no learning rate warmup, as well as a weight decay of $0.05$. The token throughput was approximately $20,000$ tokens per global step and each model was trained for approximately $5000$ steps or stopped early if suitable reward convergence or significant model deviation was observed.

\paragraph{Basic reasoning.}
For every training example that uses a chain-of-thought trace, we prepend a deterministic, task-agnostic reasoning block describing the chemical entities present in the user prompt. All molecular entities marked with \texttt{\textless /smiles\textgreater } tags are parsed and enriched with their canonical SMILES; multi-fragment structures are additionally decomposed via BRICS (Breaking of Retrosynthetically Interesting Chemical Substructures \cite{degen2008brics}, RDKit \texttt{FindBRICSBonds}~/~\texttt{FragmentOnBonds}), which provides a set of retrosynthetically motivated bond cleavages and the resulting synthon-sized fragments in SMILES form.

\begin{windowbox}[width=\linewidth]{Basic reasoning example}
    \begin{fieldbox}{}
\textless think\textgreater

I've found chemical entities in user input:
\textless smiles\textgreater n1cnc2c(C)nn(CC3CCC(c4[nH]c(=O)on4)CC3)c2c1Cl \textless /smiles\textgreater

Additional info: BRICS fragments \textless smiles\textgreater{Cc1n[nH]c2c(Cl)ncnc12}\textless /smiles\textgreater \textless smiles\textgreater{C}\textless /smiles\textgreater

\textless smiles\textgreater{C1CCCCC1}\textless /smiles\textgreater \textless smiles\textgreater{O=c1[nH]cno1}\textless /smiles\textgreater 

\textless /think\textgreater
    \end{fieldbox}
\end{windowbox}
\section{Templates Used for Benchmarking and \ourmodel{} training}
\label{app:templates}

To formulate retrosynthesis tasks in natural language, we use a set of modified templates from MolInstructions \cite{Fang2023MolInstructions}. We wrap each pair of a product and reactants into fifteen different templates, forming the final data representation. These templates in Jinja2 format are listed below.

\begin{windowbox}[width=\linewidth]{Templates for the retrosynthesis task}
    \begin{fieldbox}{Template 1}
Based on the given product, provide some plausible reactants that might have been utilized to prepare it: \textless smiles\textgreater\textcolor{blue}{\{\{source\}\}}\textless /smiles\textgreater.
    \end{fieldbox}

    \begin{fieldbox}{Template 2}
Can you identify some reactants that might result in the given product: \textless smiles\textgreater\textcolor{blue}{\{\{source\}\}}\textless /smiles\textgreater?
    \end{fieldbox}

    \begin{fieldbox}{Template 3}
Given the following product \textless smiles\textgreater\textcolor{blue}{\{\{source\}\}}\textless /smiles\textgreater, please provide possible reactants.
    \end{fieldbox}

    \begin{fieldbox}{Template 4}
Given the product provided \textless smiles\textgreater\textcolor{blue}{\{\{source\}\}}\textless /smiles\textgreater, propose some possible reactants that could have been employed in its formation.
    \end{fieldbox}

    \begin{fieldbox}{Template 5}
Given this product \textless smiles\textgreater\textcolor{blue}{\{\{source\}\}}\textless /smiles\textgreater, can you propose the corresponding reactants?
    \end{fieldbox}

    \begin{fieldbox}{Template 6}
Please suggest possible reactants for the given product: \textless smiles\textgreater\textcolor{blue}{\{\{source\}\}}\textless /smiles\textgreater
    \end{fieldbox}

    \begin{fieldbox}{Template 7}
Please suggest potential reactants for the given product: \textless smiles\textgreater\textcolor{blue}{\{\{source\}\}}\textless /smiles\textgreater
    \end{fieldbox}

    \begin{fieldbox}{Template 8}
Please suggest potential reactants used in the synthesis of the provided product: \textless smiles\textgreater\textcolor{blue}{\{\{source\}\}}\textless /smiles\textgreater
    \end{fieldbox}

    \begin{fieldbox}{Template 9}
Provide a list of potential reactants that may have produced the given product: \textless smiles\textgreater\textcolor{blue}{\{\{source\}\}}\textless /smiles\textgreater
    \end{fieldbox}

    \begin{fieldbox}{Template 10}
Provided the product below, propose some possible reactants that could have been used in the reaction: \textless smiles\textgreater\textcolor{blue}{\{\{source\}\}}\textless /smiles\textgreater
    \end{fieldbox}

    \begin{fieldbox}{Template 11}
What are the possible reactants that could have formed the following product \textless smiles\textgreater\textcolor{blue}{\{\{source\}\}}\textless /smiles\textgreater?
    \end{fieldbox}

    \begin{fieldbox}{Template 12}
What reactants could lead to the production of the following product \textless smiles\textgreater\textcolor{blue}{\{\{source\}\}}\textless /smiles\textgreater?
    \end{fieldbox}

    \begin{fieldbox}{Template 13}
Which reactants could have been used to generate the given product \textless smiles\textgreater\textcolor{blue}{\{\{source\}\}}\textless /smiles\textgreater?
    \end{fieldbox}

    \begin{fieldbox}{Template 14}
With the given product \textless smiles\textgreater\textcolor{blue}{\{\{source\}\}}\textless /smiles\textgreater, suggest some likely reactants that were used in its synthesis.
    \end{fieldbox}

    \begin{fieldbox}{Template 15}
With the provided product \textless smiles\textgreater\textcolor{blue}{\{\{source\}\}}\textless /smiles\textgreater, recommend some probable reactants that were likely used in its production.
  \end{fieldbox}
\end{windowbox}

\clearpage
\section{USPTO-50K-test Set} 
\label{app:uspto_50_k_test_mods}

The original \textbf{USPTO-50K} splits for train, validation and test subsets were introduced in \cite{liu2017uspto50ktest} and can be accessed on the GitHub repository:
\begin{itemize}
    \item Sources (SMILES strings of products): \url{https://github.com/pandegroup/reaction_prediction_seq2seq/blob/master/processed_data/test_sources}
    \item Targets (SMILES strings of reactants): \url{https://github.com/pandegroup/reaction_prediction_seq2seq/blob/master/processed_data/test_targets}
\end{itemize}

The original test set contains $5004$ reactant--product pairs. We removed $32$ entries that were clearly corrupted or ill-posed for retrosynthesis evaluation (see \autoref{tab:uspto50k-deleted}).

These exclusions target unambiguous dataset artifacts that undermine the benchmark semantics. In particular, we found cases where molecules (e.g., trifluoroacetic acid or maleic/fumaric acid) appear as the \emph{product}, which is most consistent with preprocessing or role-assignment errors (e.g., reagent/solvent leakage into the product field) rather than meaningful synthetic targets. We also removed entries with obvious processing artifacts and targets with molecular weight below 100~Da. Overall, this conservative curation reduces evaluation noise and makes reported metrics more representative of performance on chemically meaningful targets.

\begin{table*}[ht]
  \centering
  \includegraphics[
    width=\linewidth,
    clip,
    trim=0 515 0 70,
  ]{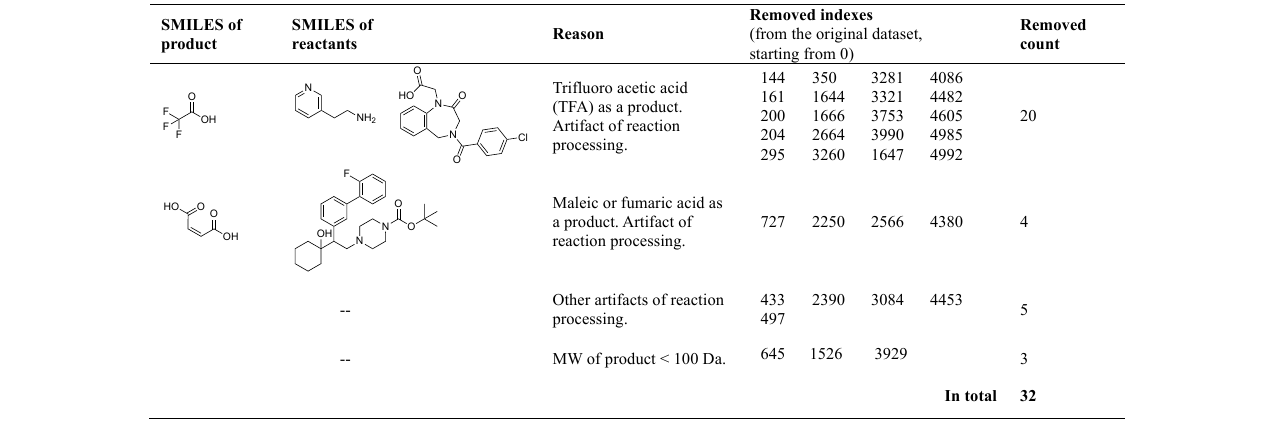}
  \caption{Items removed from the USPTO-50K test set.
}
  \label{tab:uspto50k-deleted}
\end{table*}

The modified \textbf{USPTO-50K-test} was used as a holdout set. The \textbf{CREED} and \textbf{USPTO-full} were strictly decontaminated with \textbf{USPTO-50K-test}, so the training datasets for our models don't contain reactions (and product molecules) from holdout at all.
\clearpage
\section{\expertset{} Set} 
\label{app:expert_set_app}

The SMILES strings of molecular structures from the \expertset{} set are provided in \autoref{tab:expert}. The 2D structures are provided in \autoref{tab:ursa-expert-2d}.

The dataset diversity is $0.85886$. Dataset diversity is defined as the average pairwise dissimilarity between all distinct molecule pairs:
\begin{equation}
\mathrm{Diversity} = \frac{1}{N(N-1)} \sum_{i \ne j} \left( 1 - \mathrm{Similarity}(i,j) \right),
\label{eq:diversity}
\end{equation}
where $N$ is the number of molecules in the dataset.

Molecular similarity in \autoref{eq:diversity} is computed as the cosine similarity between binary vectors of structural screens, where each vector element indicates the presence or absence of a specific atom-centered structural fragment derived using the Chemosoft software (ChemDiv Inc. chemical database software, \url{https://www.chemdiv.com/}). The set of screens is constructed dynamically across the dataset, capturing atom types, bond types, and ring environments without using hash-based compression.

\scriptsize
\begin{longtable}{lp{0.9\textwidth}}
\caption{\expertset{} set.}\label{tab:expert}\\
\toprule ID & SMILES\\
\midrule
\endfirsthead

\toprule ID & SMILES\\
\midrule
\endhead

\bottomrule
\endfoot

\begin{filecontents*}{tables/appendix/expert_set_100_canonical.csv}
Structure ID,canonical_smiles
X404-223-1421,O=C1Cc2ccc(Nc3ccc(N[C@@H]4CN(C(=O)c5cccc(NCCO)c5)C[C@@H]4c4cncs4)cc3)cc2N1
X404-133-8064,CCCc1nc2cc(Nc3c(F)cc(-c4ccc(OC)c5cn[nH]c45)cc3F)ccc2c(=O)[nH]1
X404-223-1462,CNc1ncnc2c1CCN(Cc1ccc(Sc3ccc(O)cc3)cc1-c1cc(C)n(CCC\#N)c1C)C2
X404-135-1949,CC(=O)c1ccccc1-c1nnc(-c2cc(=O)n(-c3nnc(NCc4ccc(Cl)cc4)n3C)nc2C2CCCCC2)o1
X404-219-2456,O=C1c2nc(CCC3(F)CC3)cn2CCN1c1nnc(C(F)(F)F)n1Cc1cncc2ccccc12
X404-135-0036,Cc1ccc(Nc2noc(-c3cnn4ccccc34)n2)c(COc2cnn(C(=O)C3CC3)c(=O)c2C2CCOCC2)c1
X404-135-2016,COCCc1ccc(NC(=O)Nc2ccc(C(F)(F)F)cc2-c2cnc(N3CCC(C)(O)CC3)nc2C)cn1
X404-223-6672,CNc1[nH]nc(N2CCC[C@H](Oc3ccc(-c4ccccc4C)cn3)C2)c1C(C)(C)O
X404-223-10866,CC(C)Nc1nc(-c2nnc3c(N4CCC[C@@](O)(c5ccc(CO)cc5)C4)cccn23)cs1
X404-133-6064,COCc1csc(N2CCn3nc(Nc4ccccc4Cl)cc3C2)n1
X404-135-2456,COc1cc2[nH]c(N(C)c3nc4c(s3)CN(CC3CCN(S(C)(=O)=O)CC3)CC4)nc2cc1F
X404-135-2479,CC[C@@H]1CCC[C@@H](C(=O)O)N1c1cccc(-n2cc(Oc3ncccc3-n3c(N4CCC(O)CC4)nc4ccccc43)cn2)c1
X404-216-0669,Cc1nnc(-c2cccc(Nc3cc(N4CCCC4=O)c(N4CCO[C@@H](C)[C@@H]4C\#N)cc3F)c2)s1
X404-208-1766,C[C@@H](Nc1nc(-c2ccc3[nH]c(=O)cnc3c2)nn1[C@@H]1CCNC1)C(F)(F)F
X404-133-0167,COc1cc(-c2ccc(OC)c3[nH]ccc23)cc(F)c1NC1CCN(c2cncc(C)n2)CC1
X404-223-11593,CN(C)c1cc(NCc2ccccc2N2CC3(C[C@@H](C(N)=O)CO3)C2)n(-c2cscn2)n1
X404-219-0927,C(\#C[C@H]1CCNC1)c1cnc(Oc2cccnc2-c2ncno2)cn1
X404-208-1309,CN1[C@H](CNc2ccccc2Nc2ccc(-n3nccn3)nc2)CCS1(=O)=O
X404-208-2728,N\#Cc1ccc(N2C(=O)NC[C@H]2c2nnc3n2CCN(c2nccnc2C2CC2)C3)cc1
X404-234-0753,COc1cc(-c2n[nH]cc2N(C)C[C@H]2CC[C@@H](c3ccccn3)CC2)on1
X404-133-0570,Cc1ccc(-c2nc3cnc(N[C@H]4CNC(=O)N(C)C4)cc3n2C2CC(F)(F)C2)cc1
X404-208-3152,CN(C)c1ccc(F)cc1-n1c(-c2cnn(C)c(=O)c2-c2cc(CO)n[nH]2)cnc1S(=O)(=O)c1ccccc1
X404-219-2015,CNC(=O)[C@H]1CCCC[C@@H]1NCc1cc(-c2cc(-c3cccnc3)n[nH]c2=O)[nH]n1
X404-234-0313,Cc1nonc1-c1n[nH]cc1N(C)C[C@@H]1CCCOCC1
X404-234-1944,Cc1ccc(S(=O)(=O)Cc2nc([C@@H](C)Nc3nn(C)cc3C3CCOCC3)no2)cc1
X404-1769-0103,Cc1ccc(-c2cnc3nc(NCc4ccon4)[nH]c3c2)cc1CC1CCNCC1
X404-1769-0655,COc1ccccc1-c1cccc(N2CCC(CN(C)c3cc(C4(O)COC4)cc(=O)[nH]3)CC2)c1
X404-1769-3494,CN(C)c1c(Nc2ccc(S(=O)(=O)N3CCC(F)CC3)cc2)[nH]c2ccccc12
X404-1769-3011,O=C(CC(F)(F)F)Nc1nc(-c2c(N3CCN(CC4CC4)CC3)sc3ccccc23)no1
X404-1769-1627,Cc1cc(-n2cc(C[C@H](C)C(=O)O)c3nccnc32)cc(C2CC2)c1NC(=O)C[C@H]1CCNC1
X404-1769-3571,CN(CC1(C)CCC1)C(=O)CCn1cnc2c(C(=O)O)ncnc21
X404-1769-4280,CO[C@H]1CN(Cc2ccc(=O)n(-c3c(C)noc3C)c2)C[C@@H]1F
X404-1768-0001,FC(F)C1(CNc2cnoc2)CCN(c2ccc3c(c2)C[C@@H](Oc2cc(N4CCc5cccnc5C4)ccn2)CO3)CC1
X404-1768-6593,Cn1ccc(N[C@H](CCN)c2coc3ccc(NC4CCCCC4)cc23)n1
X404-1768-0722,COc1cc(-n2ccnc2Nc2ccncc2N2C[C@H](C3CCCC3)NC2=O)ccc1O
X404-1768-1481,CS(=O)(=O)c1ccc(-c2cc[nH]c(=O)c2C[C@H]2CCN(C(=O)c3nncs3)C2)cc1
X404-1768-3854,COCC1(Nc2nc(C)c(-c3nc(C)nc4sccc34)c([C@@H]3CCCN3)n2)CCCC1
X404-1768-3154,Cn1nc(Oc2cccc(-c3ccccc3C3COC3)c2)cc1CCO
X404-1768-2834,O=C(CC(F)(F)F)Nc1ncncc1-c1nn(C2CCC2)c2ccccc12
X404-1767-0002,Oc1cc(Cl)cc(-n2cc(Oc3cccc(NC4CCC4)c3)cn2)c1Cl
X404-1767-1147,O[C@@H](CNC1CCCC1)c1cc(-c2cccc(NCc3ccncc3)n2)n[nH]1
X404-1767-0220,COc1ccc(Oc2ccc(N3C[C@@H](C)[C@@H](N)C3)n3cncc23)c(Cl)c1
X404-1767-2533,CCC(C)(C)c1nc(-c2c(N(C)Nc3cc(F)cc(F)c3)sc3c2CCC3)no1
X404-1767-2293,N[C@@H](CO)Cn1ccnc1-c1nccc(C2=CCCCC2)n1
X404-1767-1670,O=C(O)CN1CCC[C@H]1C(=O)Nc1csc(Nc2cccc3[nH]ccc23)n1
X404-1769-2917,Cc1[nH]c(NC[C@H](Nc2ccc(F)cc2)c2ccc(F)cc2)nc(=O)c1C
X404-1766-2225,C[C@@](O)(CNc1nc2n(n1)CCCC2)C1CCCC1
X404-1766-0360,O=C(Nc1cc(CO)cnc1-c1nc2ccccn2n1)C1CCNCC1
X404-1766-0764,Cc1nc(N2CC[C@H](COC3(CO)CC3)C2=O)nc2c1CCC2
X404-1766-0001,O[C@@H](c1ccc(NCC2CC2)c(-c2cccc(Cl)c2)c1)c1c[nH]nn1
X404-1766-0908,COC1(c2nnc([C@H](C)C3CCN(c4ccnn4C)CC3)o2)CC1
X404-1766-2761,Cc1cc(-c2nc(=O)c3cc(N4CCC[C@@H](C(=O)O)C4)ccc3[nH]2)cc(C)c1O
X404-1766-1829,Oc1ccc(CNC2CN(c3nc(Nc4ccc([C@@H](O)C(F)(F)F)cc4)cs3)C2)cc1
X404-1765-6158,CC(C)[C@@](C)(CN)NC(=O)Nc1nc(-c2ccc(O)cc2)c2ccsc2n1
X404-1765-2441,Cn1c(N2CC[C@H](C(=O)NC3COC3)C2)nc(=O)c2ccccc21
X404-1765-2945,NC(=O)c1ccccc1N1CCn2c(nnc2Nc2ccccc2CO)C1
X404-1765-3286,CCNc1cc(-c2nccc(O)c2C)ccc1Cc1nc(C(=O)O)no1
X404-1765-0440,CC1(C)CN(C[C@@H](Nc2noc3ccccc23)c2c[nH]cn2)C1
X404-1765-0653,CCOC[C@@H]1CC(=O)N(c2cc(Cl)cc(-c3ncnc4ccsc34)c2)C1
X404-1765-1623,N\#Cc1ccccc1C(=O)Nc1csc(N[C@H](C(=O)O)C2(C(F)(F)F)CCC2)n1
X404-1763-4916,Cn1cnc(C(=O)N2CCC(OC[C@H]3CCNCCO3)CC2)n1
X404-1763-2878,N[C@@H]1CCCN([C@@H](c2ccccc2)c2nnc(C(F)(F)F)o2)C1
X404-1763-4338,Cc1cccc(N2C[C@H](Nc3ccc(=O)n(C)c3)C[C@H]2C(=O)O)c1
X404-1767-4790,CCc1ccc(O)c([C@@H](C)Oc2cc3n(n2)CCN(c2ccc(C(=O)O)cc2)C3)c1
X404-1763-2737,CC(C)[C@H]1CCCN1c1nnc(C(=O)NC[C@H]2CC(F)(F)CN2)o1
X404-1763-3528,C[C@]1(F)C[C@H](NCc2cc(-c3cc(C(=O)O)no3)[nH]n2)C1
X404-1764-0248,Cc1ccc(N(c2ncc(-c3ccsc3)c(C)n2)C2CCN(c3cccc(Cl)c3)CC2)c(=O)[nH]1
X404-1762-2412,C[C@H](C(=O)O)N(C)CCc1nc2ccccc2n1-c1nnc(C2CC2)[nH]1
X404-1762-1236,C[C@@H](NCc1cncc2c1[nH]c(=O)n2C1CCOCC1)c1ccc(-c2cnccn2)s1
X404-1762-1239,C[C@@H](O)c1nc(NC(=O)CN2CC[C@H](C(=O)N3CCCCC3)C2)c2ccccn12
X404-1762-2518,CCn1nncc1-c1cnc([C@H](O)CN2CCC(C3(O)CC3)CC2)cn1
X404-1762-3779,C[C@H](CC(F)F)n1c(-c2nnc([C@H]3CCCCO3)o2)nc2ccccc21
X404-1762-5054,Cc1cccc(-n2cccc(-c3ccc([C@@H](OC(C)(F)F)C(N)=O)c(C)c3)c2=O)c1
X404-1762-1516,Cc1ccc(Nc2ccccc2F)cc1N(C[C@@H]1CNCCO1)c1cccnn1
X404-1761-0534,Cc1ccc(Cc2cnoc2)c(-c2ccc(=O)n(-c3cc4c(nn3)CCN(CCO)C4)c2)c1
X404-1761-2168,O=C(O)[C@@H]1CCN(c2ccc3nc(C(F)(F)F)cc(-n4cccn4)c3c2)C[C@@H]1O
X404-1761-6535,COC1CCN(C(=O)c2nn(C[C@@H](C)OC)c3c2CN(Cc2nc(C)nc4ccccc24)CC3)CC1
X404-1761-0006,O=C(O)c1cnc(-c2cnc(N3CCc4cc(CC[C@H]5CCCNC5)ccc43)c3[nH]ccc23)nc1
X404-1761-3479,FC(F)(F)c1c(N2CCOCC2)cccc1N1CCn2nc(-c3ccccc3Cl)cc2C1
X404-1761-3491,COc1ccc(-n2ncc3c(C4CC4)cc(O[C@H]4C[C@@H](C)C4)nc32)nc1
X404-1761-3689,CCc1csc(N[C@H]2C[C@@H](C(=O)N3CCn4nc(-c5ccc(C)cc5C)nc4C3)C2)n1
X404-1764-1035,N[C@@H](Cc1ccnc(Oc2ccccc2-c2ccccc2-c2ccc(=O)n([C@H]3C[C@@H](F)C3)c2)c1)C(=O)O
X404-1764-4001,Cn1c(C(=O)O)nnc1[C@H]1CCC[C@](O)(c2ccc(O)cc2)C1
X404-1764-2345,Cc1cc(CNc2ccccc2N2C[C@H](c3cnc([C@H]4CC[C@H](C(=O)O)CC4)s3)CC2=O)[nH]n1
X404-1764-0025,Oc1ccc(-c2cc(Nc3cc(Cc4cnccn4)on3)ccc2N2CCC(O)(C(F)(F)F)CC2)cc1
X404-1764-2402,NCc1ccc(-c2ncnc3scc(CN4CCN(S(=O)(=O)C5CC5)CC4)c23)cc1
X404-1764-3071,FC(F)(F)OC1(CNc2ccnc(-c3cc4ccncc4[nH]3)n2)CCCCC1
X404-1764-4281,C[C@@H]1CN(Cc2nnc3ccc(NCC(C)(C)CC\#N)cn23)CC2(CCCCC2)O1
X404-1760-0042,O=S1(=O)Nc2c(-c3ccc(O[C@@H]4CNCC4(F)F)nc3)cccc2N1Cc1ccccc1C(F)(F)F
X404-1760-1189,O=C(O)c1ccccc1-c1ccc(CN[C@@H]2CCC[C@@H]2NCC(F)(F)F)[nH]1
X404-1760-0185,C[C@@H](NC(=O)CCC(F)(F)F)c1ccc(N2CC[C@H](Nc3csc(-c4ccc(O)cc4)n3)C2)cc1
X404-1760-0023,O=C1CC2(CCC2)CCN1C(=O)c1cc(-c2cc(C3CC(F)(F)C3)c3cn[nH]c3n2)cc2cccnc12
X404-1761-3537,Cc1ccn2ncc(-c3cnc(N4CCC(N5CCC[C@@H]5C)CC4)o3)c2c1
X404-1760-0572,CNC[C@@H](O)c1ccc(C2(CNC)CCCC2)cc1
X404-1760-2143,COc1cc2n(n1)CCN(CC(=O)Nc1cccc3c1cc(N1CCOC[C@@H](C)C1)n3C)C2
X404-1768-5005,CC[C@H](Nc1cnn(C(C)C)c(=O)c1)c1ccc(-c2cccc(NC(=O)c3cscn3)c2)cc1
X404-1760-1425,Cc1cc(N2CCC(c3cc(-c4ccc(F)cc4)on3)CC2)n(CCCS(C)(=O)=O)n1
X404-1761-2220,CCCc1nc2c(-c3nccn3[C@@H]3CN(C(=O)C4CCC4)C[C@H]3F)ccnc2[nH]1
X404-1761-4128,CCN(c1cccc([C@@H](C)Nc2nc(=O)c(-c3ccccc3)c[nH]2)c1)S(C)(=O)=O
X404-1768-3704,Cc1cc(=O)nc(N(Cc2ccc(N3CC(C(F)(F)F)C3)cc2)c2cccc(C\#N)c2)[nH]1
\end{filecontents*}
\csvreader[
  separator=comma,
  late after line=\\
]{tables/appendix/expert_set_100_canonical.csv}{}{%
  \texttt{\csvcoli} & \texttt{\csvcolii}
}
\end{longtable}
\normalsize

\includepdf[
  pages=1,
  offset=0 -7mm,
  scale=0.92,
  pagecommand={%
    \thispagestyle{plain}%
    \captionsetup{type=table}%
    \vspace*{0em}%
    \caption{URSA-expert-2026 set 2D structures.}%
    \label{tab:ursa-expert-2d}%
    \vspace*{0em}%
  }
]{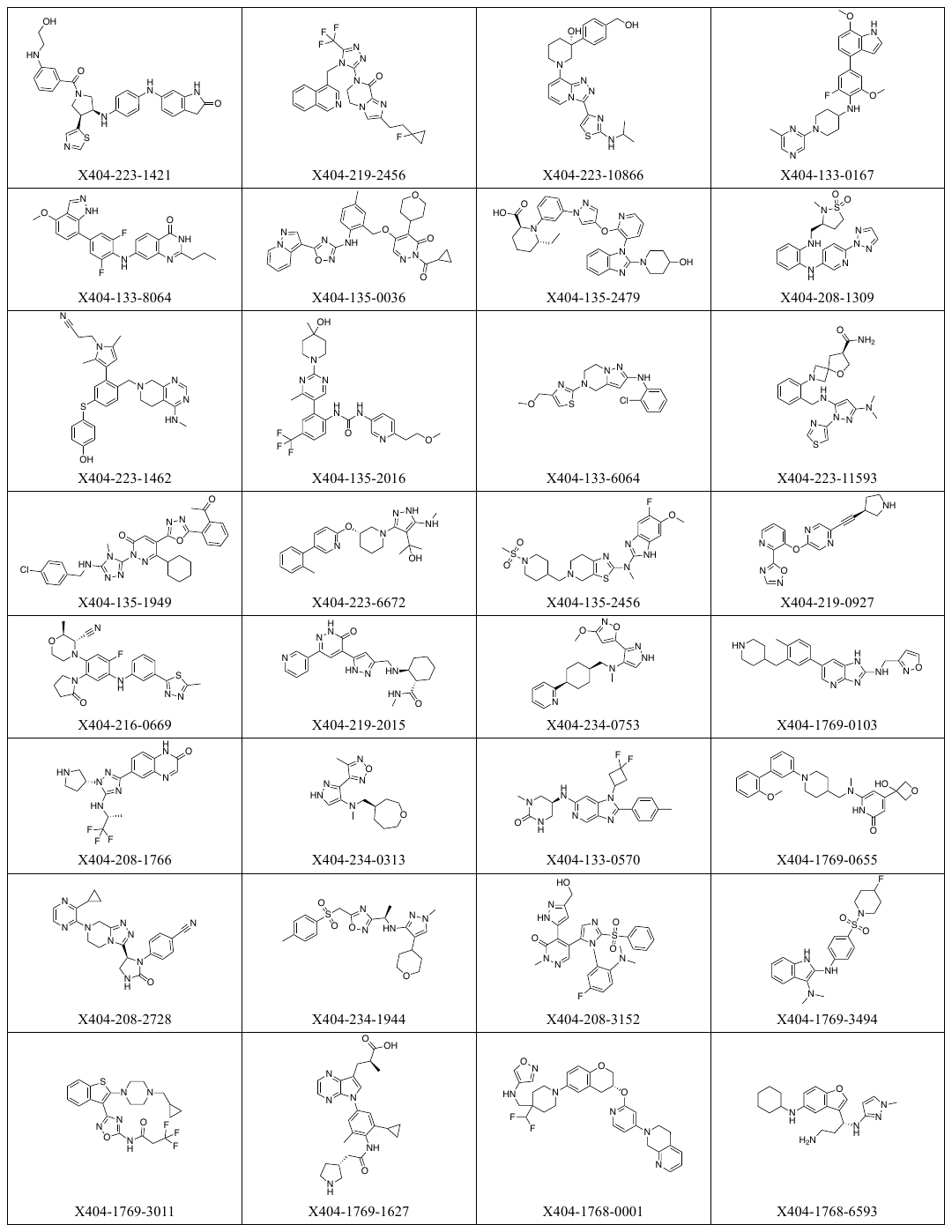}
\includepdf[
  pages=2-,
  offset=0 -7mm,
  scale=0.92,
  pagecommand={
  \thispagestyle{plain}
    \captionsetup{type=table}%
    \caption*{Continuation of \autoref{tab:ursa-expert-2d}.}%
  }
]{figs/appendix/URSA-expert-2026/Expert_Data_Set.pdf}

\begin{figure*}[ht]
  \centering
  \includegraphics[
    width=\linewidth,
    clip
  ]{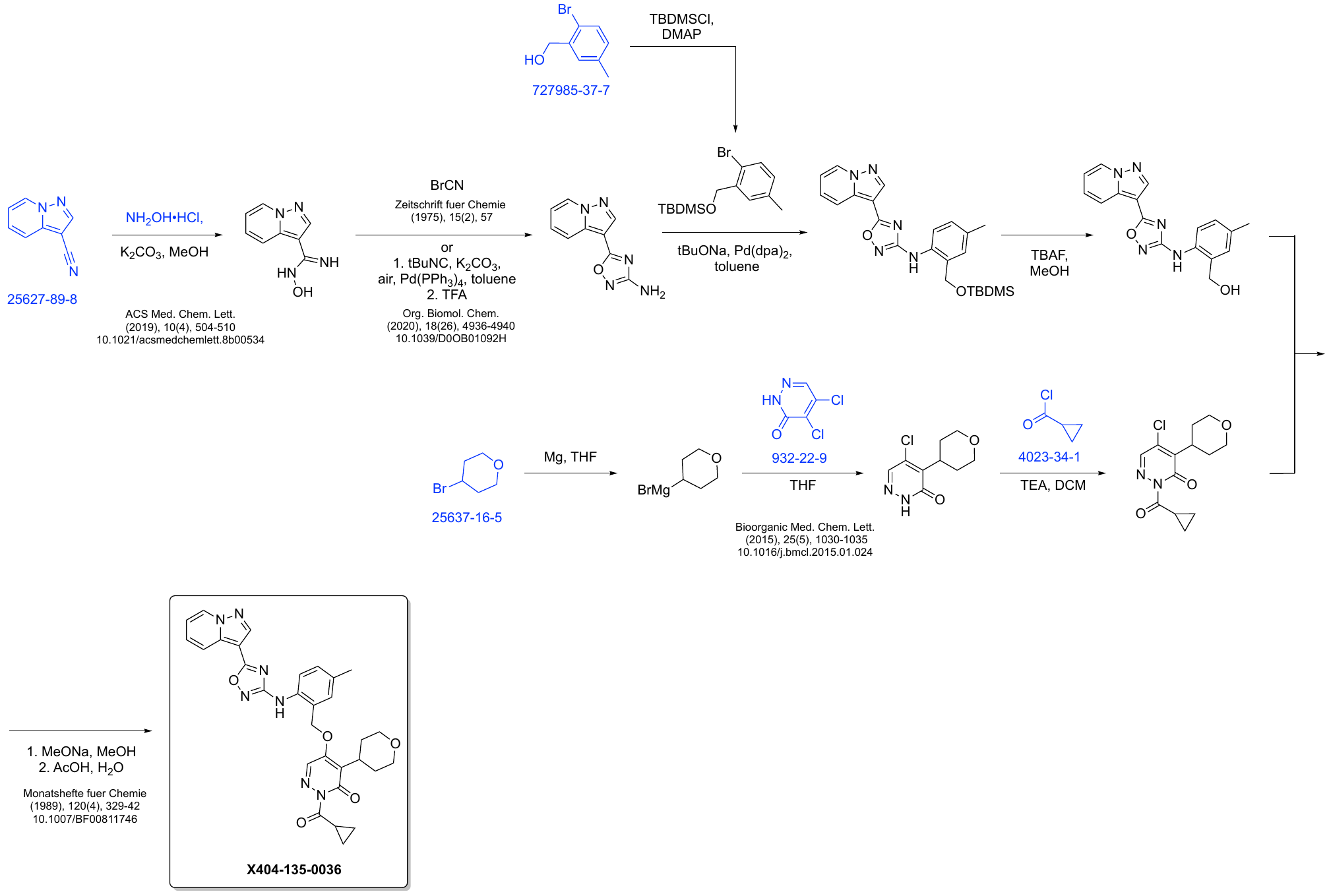}
  \caption{Theoretical synthesis scheme for the compound X404-135-0036 from \expertset{} set.
}
  \label{fig:Expert_scheme}
\end{figure*}

As proof of the confirmed synthetic accessibility status of molecular structures in the \expertset{}, we provide the example of a theoretical synthetic scheme proposed by synthetic organic chemistry experts (see \autoref{fig:Expert_scheme}).

The \expertset{} was used as a holdout set. The CREED and USPTO-full were decontaminated within \expertset{}, so the training datasets for our models don't contain reactions and product molecules from the holdout at all.
\clearpage
\section{Benchmark Models Completion Examples}
\label{app:completions_examples}

Most of the general-purpose models we tested support reasoning, but the
availability of reasoning traces differs. GPT\cite{openai2025gpt51, openai2025gpt52} models do not expose 
reasoning traces. Grok\cite{xai2025grok41} exposes encrypted reasoning traces, so ordinary
users cannot easily analyze them. DeepSeek 3.2 \cite{deepseekai2025deepseekv32} does not support explicit reasoning. 
Kimi-K2-thinking \cite{kimiteam2025kimik2openagentic} has thinking traces, but does not expose them through API.

\begin{windowbox}[width=\linewidth]{User input}
\ttfamily\footnotesize
 \textbf{Task:} You are a helpful assistant in chemistry and biology. Your task is to predict the reactants that synthesize the given product. Carefully study previous conversation to match the exact output format. The SMILES strings in the final answer should be wrapped in \textless smiles\textgreater...\textless /smiles \textgreater tags. 

<Few-shot examples>. 

Based on the given product, provide some plausible reactants that might have been utilized to prepare it.

\textbf{Product:} \texttt{Cc1ccc(Nc2noc(-c3cnn4ccccc34)n2)c(COc2cnn(C(=O)C3CC3)c(=O)c2C2CCOCC2)c1}\\
\textbf{Return:} Reasoning + Answer.
\end{windowbox}
\vspace{0.8em}

\begin{windowbox}[width=\linewidth]{Model: Grok-4.1}
  \begin{fieldbox}{Reasoning}
Not available
  \end{fieldbox}
  \begin{fieldbox}{Answer}
\textless smiles\textgreater Cc1ccc(Nc2noc(-c3cnn4ccccc34)n2)c(COc2cnnc(=O)c2C2CCOCC2)c1.
O=C(Cl)C1CC1 \textless /smiles \textgreater
  \end{fieldbox}
\end{windowbox}

\vspace{0.8em}

\begin{windowbox}[width=\linewidth]{Model: Gemini 2.5 Flash}
  \begin{fieldbox}{Reasoning}
Alright, let's break this down. My primary goal here is to identify potential synthetic routes to this complex molecule, focusing on plausible starting materials and key bond formations. The product is a beast! It's clear that I need to dissect this structure into manageable pieces.

First, I'm thinking about the central benzene ring. It acts like a hub, connecting several important fragments. This ring contains an amine group (N-aryl linkage) and an ether linkage (-CH2O-). That amine is likely formed via Buchwald-Hartwig amination or Ullmann condensation. The ether, an SN2 reaction or a Williamson ether synthesis or something similar, and I'll need to figure out which reaction is most suitable.

Now, let's look at the fragments. The left side of the nitrogen - the `Nc2noc(-c3cnn4ccccc34)n2` fragment – looks like an oxadiazole ring fused to a benzotriazole-like system – an imidazobenzotriazole derivative. That amino-oxadiazole ring, `Nc2noc(-c3cnn4ccccc34)n2` may be formed from an amidoxime and an acyl chloride derivative. The oxadiazole itself can be built from amidoximes and then coupled to the complex heterocycle.

The right side, `COc2cnn(C(=O)C3CC3)c(=O)c2C2CCOCC2`, is a pyrazolone with a cyclopropane carbonyl and a tetrahydropyran.  This is a pyrazolone derivative.  This structure contains a carbonyl and likely made via standard pyrazolone synthesis using a beta-keto ester and a hydrazine derivative.

To summarize, for the N-aryl bond, I could see using a Buchwald-Hartwig amination approach. For example, a bromo-benzene derivative and the oxadiazole-containing amine.  For the ether linkage, I anticipate the reaction between an alcohol and an aryl halide, or the alcohol version of the central aryl fragment and the pyrazolone derivative that is being attached.
  \end{fieldbox}
  \vspace{0.4em}
  \begin{fieldbox}{Answer}
The complex structure of the given product suggests a multi-step synthesis involving several key transformations. We can deduce plausible reactants by breaking down the product into simpler, recognizable fragments and considering common synthetic reactions for their formation and subsequent coupling.
Product:
\textless smiles\textgreater

Cc1ccc(Nc2noc(-c3cnn4ccccc34)n2)c(COc2cnn(C(=O)C3CC3)c(=O)c2C2CCOCC2)c1
\textless /smiles \textgreater%

Let's break down the molecule:

1.  **Central Phenyl Core:** `Cc1ccc(N...)c(CO...)c1`
2.  **Left Arm (N-aryl-oxadiazole-heterocycle):** `Nc2noc(-c3cnn4ccccc34)n2`
3.  **Right Arm (Ether-linked pyrazolone derivative):** `COc2cnn(C(=O)C3CC3)c(=O)c2C2CCOCC2`

Based on this, we can propose the following disconnections:

**Disconnection 1: N-aryl bond formation**
The N-aryl bond (`Ar-NH-Ar'`) is commonly formed via a Buchwald-Hartwig amination or Ullmann condensation. This would involve an aryl halide and an amine.

\textless skipped reasoning \textgreater

Let's define the two major fragments:

**Fragment A (Amine component):**
`H2N-c1noc(-c2cnn3ccccc23)n1`
This is a 5-amino-1,2,4-oxadiazole derivative.
The `-c2cnn3ccccc23` group is likely a `[1,2,4]triazolo[4,3-a]quinoline` or similar bicyclic nitrogen heterocycle. Let's call this `R\_A`.
A common synthesis for 5-amino-1,2,4-oxadiazoles involves reacting an amidoxime (`R-C(=NOH)NH2`) with cyanogen halide (`Cl-CN`) or similar reagent, or reacting a nitrile (`R-CN`) with hydroxylamine to form the amidoxime, followed by cyclization.
So, `R\_A-CN` (e.g., `[1,2,4]triazolo[4,3-a]quinoline-CN`) and `NH2OH` (hydroxylamine) could be precursors to the `R\_A-C(=NOH)NH2` which then forms `H2N-c1noc-R\_A`.

**Fragment B (Phenolic component):**
`HO-c1cnn(C(=O)C2CC2)c(=O)c1C1CCOCC1`
This is a `1H-pyrazole-5-ol` or `5-hydroxypyrazole` derivative. The keto-enol tautomerism of pyrazolones is relevant here.
The core pyrazolone ring (`N-N=C-C(=O)-C`) is often formed by the condensation of a beta-keto ester with a hydrazine derivative.
In this case, the pyrazolone has:
*   An N-substituent: `C(=O)C2CC2` (cyclopropylcarbonyl)
*   A C-substituent: `C1CCOCC1` (tetrahydropyran-4-yl)
So, a possible formation route for Fragment B could involve:
1.  **Hydrazine derivative:** `H2N-NH-C(=O)C2CC2` (N-cyclopropylcarbonyl hydrazine)
2.  **Beta-keto ester:** `R-CO-CH2-COOEt` where `R` is `C1CCOCC1` (ethyl 3-oxo-3-(tetrahydropyran-4-yl)propanoate)

\textless skipped reasoning \textgreater

**Overall Proposed Reactants:**

The most direct disconnection leads to three main fragments that are then coupled.

*   **Fragment 1 (central core with leaving group for N-arylation and a nucleophilic alcohol for etherification):**
`\textless smiles\textgreater Cc1ccc(Br)c(CO)c1\textless /smiles \textgreater`

*   **Fragment 2 (amino-oxadiazole):**
`\textless smiles\textgreater Nc1noc(-c2cnn3ccccc23)n1\textless /smiles \textgreater`
(This compound itself might be synthesized from `[1,2,4]triazolo[4,3-a]quinoline-3-carbonitrile` and hydroxylamine, followed by cyanation and cyclization).

*   **Fragment 3 (hydroxypyrazolone):**
`\textless smiles\textgreater Oc1cnn(C(=O)C2CC2)c(=O)c1C1CCOCC1\textless /smiles \textgreater`
(This compound can be synthesized from `H2N-NH-C(=O)C2CC2` and `ethyl 3-oxo-3-(tetrahydropyran-4-yl)propanoate`).

Given the complexity, a multi-step synthesis would be required. One possible sequence:

1.  Buchwald-Hartwig coupling of `Cc1ccc(Br)c(CO)c1` with `Nc1noc(-c2cnn3ccccc23)n1` to form the N-aryl bond. This would yield `\textless smiles\textgreater Cc1ccc(Nc2noc(-c3cnn4ccccc34)n2)c(CO)c1\textless /smiles \textgreater`.
2.  Then, a Mitsunobu reaction or a Williamson ether synthesis between the benzylic alcohol (`-CH2OH`) from the product of step 1 and the phenolic pyrazolone (`Oc1cnn(C(=O)C2CC2)c(=O)c1C1CCOCC1`). For Williamson, the alcohol would first be converted to a bromide, then reacted with the phenol.

Therefore, the likely direct reactants for the final coupling steps are:

\textless smiles \textgreater
Cc1ccc(Nc2noc(-c3cnn4ccccc34)n2)c(CO)c1
\textless /smiles \textgreater\

\textless smiles \textgreater
Oc1cnn(C(=O)C2CC2)c(=O)c1C1CCOCC1
\textless /smiles \textgreater
\end{fieldbox}
\end{windowbox}

\vspace{0.8em}

\begin{windowbox}[width=\linewidth]{Model: Gemini 3 Flash preview}
  \begin{fieldbox}{Reasoning}
**Parsing the Structure**

Now, I'm carefully dissecting the pyrazolone-containing fragment, attempting to identify each atom's bonding. It seems that the key to understanding the structure lies in correctly interpreting the ring closure within the SMILES string. I'm focusing on the position of the tetrahydropyran substituent.
  \end{fieldbox}
  \vspace{0.4em}
  \begin{fieldbox}{Answer}
\textless smiles \textgreater Cc1ccc(Nc2noc(-c3cnn4ccccc34)n2)c(CO)c1.
O=C(c1cnn(C(=O)C2CC2)c(=O)c1C1CCOCC1)O \textless /smiles \textgreater
  \end{fieldbox}
\end{windowbox}

\vspace{0.8em}

\begin{windowbox}[width=\linewidth]{Model: GPT 5.1}
  \begin{fieldbox}{Reasoning}
Not available
  \end{fieldbox}
  \vspace{0.4em}
  \begin{fieldbox}{Answer}
\textless  smiles \textgreater Cc1ccc(N)c(COc2cnn(C(=O)C3CC3)c(=O)c2C2CCOCC2)c1.c1noc(-c2cnn3ccccc23)n1Cl\textless  /smiles \textgreater 
  \end{fieldbox}
\end{windowbox}

\vspace{0.8em}

\begin{windowbox}[width=\linewidth]{Model: GPT 5.2}
  \begin{fieldbox}{Reasoning}
Not available
  \end{fieldbox}
  \vspace{0.4em}
  \begin{fieldbox}{Answer}
\textless  smiles \textgreater Cc1ccc(N)c(COc2cnn(C(=O)C3CC3)c(=O)c2C2CCOCC2)c1.Clc1noc(-c2cnn3ccccc23)n1\textless  /smiles \textgreater 
  \end{fieldbox}
\end{windowbox}

\vspace{0.8em}

\begin{windowbox}[width=\linewidth]{Model: Claude Sonnet 4.5}
  \begin{fieldbox}{Reasoning}
Let me analyze this complex molecule step by step to identify potential reactants.

The product is: \textless smiles \textgreater Cc1ccc(Nc2noc(-c3cnn4ccccc34)n2)c(COc2cnn(C(=O)C3CC3)c(=O)c2C2CCOCC2)c1 \textless /smiles \textgreater

Let me break down the structure:
1. There's a methylated benzene ring with an NH group connected to an oxadiazole
2. The oxadiazole has a 1H-indazole substituent
3. The benzene also has a CH2O- linker connecting to a pyridazinone
4. The pyridazinone has a cyclopropylcarbonyl group on the nitrogen
5. The pyridazinone also has a tetrahydropyran substituent

Looking at the structure, this appears to be formed from multiple coupling reactions. Let me identify potential disconnection points:

1. The NH-oxadiazole bond could be formed via nucleophilic aromatic substitution
2. The CH2O ether linkage could be formed via Williamson ether synthesis
3. The cyclopropylcarbonyl on the nitrogen could be an acylation

Key fragments I can identify:
- The oxadiazole with indazole: -c1noc(-c2cnn3ccccc23)n1
- The aromatic amine with CH2OH: Cc1ccc(N)c(CO)c1 or similar
- The pyridazinone core with substituents

Let me think about the most likely disconnections:

The most straightforward synthesis would involve:
1. A substituted benzene with NH2 and CH2O- groups
2. An oxadiazole electrophile (likely with a leaving group like Cl)
3. A pyridazinone with an OH group that can react with the CH2 group

The most probable route involves coupling an aniline bearing the CH2O-pyridazinone moiety with a chloro-oxadiazole intermediate. Alternatively, the oxadiazole-aniline coupling could occur first, followed by attachment of the pyridazinone through the CH2 linker. about which is more likely. The first option seems more straightforward: the aniline nitrogen would perform a nucleophilic aromatic substitution on the chloro-oxadiazole, displacing the chlorine and forming the N-aryl bond directly.
  \end{fieldbox}
  \vspace{0.4em}
  \begin{fieldbox}{Answer}
Looking at this complex molecule, I can identify key disconnection points. The most likely synthesis involves:

1. A nucleophilic aromatic substitution between an aniline and a chloro-oxadiazole
2. The aniline already contains the CH2O-pyridazinone substituent

**Plausible reactants:**

\textless smiles \textgreater Clc1noc(-c2cnn3ccccc23)n1 \textless /smiles \textgreater

\textless smiles \textgreater Cc1ccc(N)c(COc2cnn(C(=O)C3CC3)c(=O)c2C2CCOCC2)c1 \textless /smiles \textgreater

The reaction would proceed via nucleophilic aromatic substitution where the aniline nitrogen attacks the chloro-oxadiazole, displacing the chloride leaving group.
  \end{fieldbox}
\end{windowbox}

\vspace{0.8em}

\begin{windowbox}[width=\linewidth]{Model: Claude Opus 4.5}
  \begin{fieldbox}{Reasoning}
Let me analyze the product structure to determine possible reactants.

Product SMILES: Cc1ccc(Nc2noc(-c3cnn4ccccc34)n2)c(COc2cnn(C(=O)C3CC3)c(=O)c2C2CCOCC2)c1

This is a complex molecule. Let me break it down:

1. There's a methylphenyl group (toluene derivative) with:
- A -Nc2noc(-c3cnn4ccccc34)n2 substituent (an oxadiazole ring with an indazole)
- A -COc2cnn(C(=O)C3CC3)c(=O)c2C2CCOCC2 substituent (a pyridazinone with cyclopropyl amide and tetrahydropyran)

Looking at this structure, I can see several key components:

1. An oxadiazole ring (-noc-n-) connected to:
- An aniline nitrogen
- An indazole (-c3cnn4ccccc34)

2. A pyridazinone system with:
- An ether linkage (COc)
- A cyclopropyl carbonyl group C(=O)C3CC3
- A tetrahydropyran ring (C2CCOCC2)

Possible disconnections:
1. The ether bond (C-O-C) between the benzyl and pyridazinone
2. The C-N bond between the aniline and oxadiazole

For the synthesis, plausible reactants could be:

1. A halogenated pyridazinone (like a chloropyridazinone):
Clc1cnn(C(=O)C2CC2)c(=O)c1C1CCOCC1

2. A hydroxymethyl-substituted aniline bearing the oxadiazole group: HOCc1cc(C)ccc1Nc2noc(-c3cnn4ccccc34)n2

The key disconnections appear to be the ether linkage between the benzyl carbon and pyridazinone oxygen, and potentially the C-N bond connecting the aniline to the oxadiazole. A nucleophilic aromatic substitution or Williamson ether synthesis could couple these fragments together. =O) at the ring, and a tetrahydropyran substituent attached. For an ether formation reaction, I need an alcohol group and a halide leaving group. The benzyl alcohol from the aniline-oxadiazole compound would serve as the nucleophile, while the chlorine on the pyridazinone ring would act as the leaving group in a Williamson ether synthesis.

Let me map out the reactants more carefully. The first component contains the methylated aromatic ring with the aniline linker and indazole moiety, terminating in a primary alcohol. The second reactant is the chlorinated pyridazinone with the cyclopropyl carbamate and tetrahydropyran groups. Now I need to properly encode these structures—let me verify the connectivity by working through the product structure systematically, starting from the central benzene ring and tracing each substituent to ensure the SMILES representation is accurate. =O)c1C1CCOCC1

Converting to SELFIES format requires careful parsing of the SMILES notation. The chloropyridazinone reactant has a pyridazinone core with a cyclopropyl carbamate substituent and a morpholine ring attached. I'll work through the SMILES systematically to ensure accurate conversion, starting with the alcohol component and then the chlorinated pyridazinone to map out the ether bond formation properly. The indazole is fused to a benzene ring, and there's a hydroxymethyl substituent on the benzene. For the chloropyridazinone component, I have chlorine on the ring itself, an N-cyclopropylcarbonyl group, and a tetrahydropyran moiety attached. Converting these structures to SELFIES notation for the final answer. )n2)c(CO)c1 for the benzyl alcohol derivative. For the chloropyridazinone, I need to verify the structure matches the product fragment where the oxygen connects to the pyridazinone ring. The reactant should have a chloride at that position: Clc1cnn(C(=O)C3CC3)c(=O)c1C1CCOCC1. I'm reconsidering the ring numbering to ensure the connectivity is correct for the coupling reaction. The chlorinated pyridazinone can be written more concisely as Clc1cnn(C(=O)C2CC2)c(=O)c1C1CCOCC1, which clearly shows the chlorine substituent, the cyclopropyl carbonyl group on one nitrogen, the carbonyl making it a pyridazinone, and the tetrahydropyran ring attached to the aromatic core.
  \end{fieldbox}
  \vspace{0.4em}
  \begin{fieldbox}{Answer}
Looking at the product structure, I can identify a key ether linkage (C-O-C) between the benzyl group and the pyridazinone ring. This suggests a nucleophilic substitution reaction between a benzyl alcohol and a halogenated pyridazinone.

The product can be disconnected into:
1. A benzyl alcohol containing the oxadiazole-indazole moiety
2. A chloropyridazinone with the cyclopropyl amide and tetrahydropyran substituents

\textless smiles \textgreater Cc1ccc(Nc2noc(-c3cnn4ccccc34)n2)c(CO)c1 \textless /smiles \textgreater

\textless smiles \textgreater Clc1cnn(C(=O)C2CC2)c(=O)c1C1CCOCC1 \textless /smiles \textgreater
  \end{fieldbox}
\end{windowbox}

\vspace{0.8em}

\begin{windowbox}[width=\linewidth]{Model: DeepSeek 3.2}
  \begin{fieldbox}{Reasoning}
Not available
  \end{fieldbox}
  \vspace{0.4em}
  \begin{fieldbox}{Answer}
Based on the structure analysis, the product likely results from a Buchwald-Hartwig amination coupling between an aniline derivative and a halogenated heterocycle. The plausible reactants are:

\textless smiles \textgreater Cc1ccc(NH2)c(COc2cnn(C(=O)C3CC3)c(=O)c2C2CCOCC2)c1 \textless /smiles \textgreater
\textless smiles \textgreater Clc2noc(-c3cnn4ccccc34)n2 \textless /smiles \textgreater
  \end{fieldbox}
\end{windowbox}

\vspace{0.8em}

\begin{windowbox}[width=\linewidth]{Model: Kimi K2}
  \begin{fieldbox}{Reasoning}
Not available
  \end{fieldbox}
  \vspace{0.4em}
  \begin{fieldbox}{Answer}
\textless  smiles\textgreater Cc1ccc(Nc2noc(-c3cnn4ccccc34)n2)c(CBr)c1 \textless /smiles\textgreater

\textless  smiles \textgreater Oc1cnn(C(=O)C2CC2)c(=O)c1C1CCOCC1 \textless  /smiles \textgreater 
  \end{fieldbox}
\end{windowbox}

\vspace{0.8em}

\begin{windowbox}[width=\linewidth]{Model: RetroDFM-R}
  \begin{fieldbox}{Reasoning}
The target molecule, 2-(cyclopropanecarbonyl)-5-[[5-methyl-2-[(5-pyrazolopyridinyl-1,2,4-oxadiazol-3-yl)amino]phenyl]methoxy]-4-(tetrahydropyran-4-yl)pyridazin-3-one, features several critical functional groups: (1) a pyridazinone ring with a tetrahydropyran substituent, (2) a substituted benzene ring bearing an oxadiazole-linked anilino group, and (3) a cyclopropanecarbonylamino group on the pyridazinone nitrogen. Retrosynthetically, the acyl group (cyclopropanecarbonyl) on the pyridazinone nitrogen is disconnected via amide bond formation. This suggests the precursor pyridazinone compound with a free NH group reacts with cyclopropanecarbonyl chloride under basic conditions to form this acylation step. The ether linkage (methoxy group) on the benzene-pyridazinone backbone may arise earlier from a Williamson ether synthesis. However, the ground-truth reactants indicate the ether structure is preassembled in one precursor. Thus, the synthesis likely prioritizes installing the ether via alkylation/arylation before introducing the acyl group on the nitrogen. However, the ground-truth reactants already incorporate this ether, implying this strategy was pre-installed during their assembly.

The proposed disconnection focuses on the acylation step. Amide formation via acylation with cyclopropanecarbonyl chloride under Schotten-Baumann-type conditions (e.g., pyridine) ensures efficient coupling with nucleophilic attack. This pathway aligns well with the ground-truth reactants provided. Steric and electronic factors (e.g., pyridazinone’s resonance structure influencing NH reactivity) are mitigated using coupling reagents or elevated temperatures to drive selectivity. This retrosynthetic approach avoids side reactions (e.g., over-chloroformylation) due to steric hindrance around the pyridazinone ring. This two-component approach (preassembled ether intermediate and cyclopropanecarbonyl electrophile) provides a robust, modular strategy with compatibility to aromatic ether formation, while maintaining regiochemical control. The approach aligns with the provided ground-truth reactants, validating this disconnection’s feasibility.
  \end{fieldbox}
  \vspace{0.4em}
  \begin{fieldbox}{Answer}
Cc1ccc(Nc2noc(-c3cnn4ccccc34)n2)c(COc2cn[nH]c(=O)c2C2CCOCC2)c1.C(=O)(Cl)C1CC1
  \end{fieldbox}
\end{windowbox}

\vspace{0.8em}

\begin{windowbox}[width=\linewidth]{Model: NatureLM}
  \begin{fieldbox}{Reasoning}
Not available
  \end{fieldbox}
  \vspace{0.4em}
  \begin{fieldbox}{Answer}
\textless  reactants\textgreater Cc1ccc(Nc2noc(-c3cnn4ccccc34)n2)c(COc2cn[nH]c(=O)c2C2CCOCC2)c1.

C(=O)(Cl)C1CC1\textless  /reactants\textgreater 
  \end{fieldbox}
\end{windowbox}

\vspace{0.8em}

\begin{windowbox}[width=\linewidth]{Model: ether0}
  \begin{fieldbox}{Reasoning}
Okay, let's see... The user is asking about a chemical molecule and wants a one-step synthesis from purchasable precursors. The SMILES for the target molecule is: Cc1ccc(N-c2noc(-c3cnn4ccccc34)n2)c(CO-c2cnn(C(=O)C3CC3)c(=O)c2C2CCOCC2)c1.

The SMILES notation is complex, indicating multiple rings connected together. The target molecule is likely a substituted benzimidazole or similar heterocycle with functional groups, including an ether, ketone, and cyclopropane rings. The user is suggesting a reaction involving multiple components. The challenge is to find a reaction that can produce the specified compound from purchasable materials in one step. Common reactions include nucleophilic substitutions, amide couplings, or condensation reactions. The presence of nitrogen-containing rings suggests a amine-based coupling reaction. The molecule already has N attached directly to benzene rings, suggesting a aniline or diaryl amine. The structure could be a coupling between an aryl halide and a complex amine via a coupling reaction. The target has a complex structure with multiple rings and substituents, including an N-substituted aryl ether or amide linkage. The main benzene ring (c1cccc1) has a methyl substituent, a nitro group, and a substituent with amine linked to another ring. The benzene ring is Cc1ccc(...)c(CO...), implying position 4 with a methyl group and an activating groups like CCO or O connected through an ether linkage at the 4-position. Considering reactions:
1. **SNAr (substituent on aromatic rings)**:
    - Benzene substituents can undergo coupling reactions, like substitution of an existing leave group with an amine subsequent from another carbonyl or bromide source.
    - Multiple electron-withdrawing groups activate the ring and allow substitution.

2. **Benzimidazole derivatives can be formed via a C-N bond formation:**
    - The amine and aryl chloride or halide could attach to form a substituting amine.
    - Perhaps a Buchwald–Hartwig amination?

Possible precursors: The first part is the source molecule, and the second component is the amine. Given the complexity, the reaction likely links these parts. **Reaction Component:** The most plausible reaction is forming an ether linkage (COgene) or a condensation between a ketone and amine groups: The original reaction components might be: CO or COc2cnn(C(=O)C3CC3)c(=O)c2C2CCOCC2 succinctly suggesting a one possibility of linking substituents via coupling.
The succinct reaction SMILES: Cc1ccc(Br)c(CO-c2cnn(C(=O)C3CC3)c(=O)c2C2CCOCC2)c1.N-c2noc(-c3cnn4ccccc34)n2\textgreater[Cu]\textgreater Cc1ccc(N-c2noc(...)n2)c(CO-c2...).
  \end{fieldbox}
  \vspace{0.4em}
  \begin{fieldbox}{Answer}
Cc1ccc(Br)c(COc2cnn(C(=O)C3CC3)c(=O)c2C2CCOCC2)c1.N-c2noc(-c3cnn4ccccc34)n2\textgreater[Cu]\textgreater Cc1ccc(N-c2noc(-c3cnn4ccccc34)n2)c(CO-c2cnn(C(=O)C3CC3)c(=O)c2C2CCOCC2)c1
  \end{fieldbox}
\end{windowbox}

\vspace{0.8em}

\begin{windowbox}[width=\linewidth]{Model: C3LM-LFM2-CREED}
  \begin{fieldbox}{Reasoning}
I've found chemical entities in user input:

\textless  smiles\textgreater C1(C(=O)n2c(=O)c(C3CCOCC3)c(OCc3cc(C)ccc3Nc3nc(-c4c5n(nc4)cccc5)on3)cn2)CC1\textless  /smiles\textgreater 

Additional info: BRICS fragments

\textless  smiles\textgreater C1CC1\textless  /smiles\textgreater  \textless  smiles\textgreater O=Cn1ncccc1=O\textless  /smiles\textgreater 

\textless  smiles\textgreater C1CCOCC1\textless  /smiles\textgreater  \textless  smiles\textgreater O\textless  /smiles\textgreater  \textless  smiles\textgreater C\textless  /smiles\textgreater 

\textless  smiles\textgreater Cc1ccccc1\textless  /smiles\textgreater  \textless  smiles\textgreater N\textless  /smiles\textgreater  \textless  smiles\textgreater c1ncon1\textless  /smiles\textgreater 

\textless  smiles\textgreater c1ccn2nccc2c1\textless  /smiles\textgreater 

  \end{fieldbox}
  \vspace{0.4em}
  \begin{fieldbox}{Answer}
\textless  smiles\textgreater C1CC1C(=O)n1ncc(OCc2c(Nc3nc(-c4c5ccccn5nc4)on3)ccc(C)c2)c(C2(O)CCOCC2)c1=O\textless  /smiles\textgreater 
  \end{fieldbox}
\end{windowbox}

\section{Ablation Studies}
\label{app:ablation_studies}

\begin{table*}[th!]
\centering
\begin{tabular}{@{}l|c|cccc|c|cccc@{}}
\toprule
\multirow{2}{*}{\textbf{Model}} &
\multicolumn{5}{c|}{\textbf{URSA-expert-2026}} &
\multicolumn{5}{c}{\textbf{USPTO-50K-test}} \\
\cline{2-6}\cline{7-11}
& \multirow{2}{*}{\textbf{Unique}} & \multirow{2}{*}{\textbf{Max}} & \multicolumn{3}{c|}{\textbf{Av. PT-Top-K CC}}
& \multirow{2}{*}{\textbf{Unique}} & \multirow{2}{*}{\textbf{Max}} & \multicolumn{3}{c}{\textbf{Av. PT-Top-K CC}} \\
&  &  & \textbf{@3} & \textbf{@5} & \textbf{@10}
&  &  & \textbf{@3} & \textbf{@5} & \textbf{@10} \\
\midrule

\multicolumn{11}{c}{\textit{\ourmodel{} (LFM2 base model), Supervised Fine-Tuning}} \\
\midrule

C3LM-LFM2-CREED-CCV+USPTO             & 42\% & 1.63 & 1.08 & 0.74 & 0.39 & 31\% & 4.05 & 2.08 & 1.35 & 0.70 \\
\makecell[l]{C3LM-LFM2-CREED-CCV+\\USPTO-AR}              & 70\% & 1.81 & 1.38 & 1.12 & 0.67 & 58\% & 3.50 & 2.29 & 1.74 & 1.01 \\

\midrule
\multicolumn{11}{c}{\textit{\ourmodel{} (LFM2 base model), Reinforcement Learning Fine-Tuning}} \\
\midrule

C3LM-LFM2-AR-RFT-CC               & 50\% & 1.95 & 1.53 & 1.19 & 0.66 & 41\% & 3.79 & 2.46 & 1.79 & 0.97 \\

\midrule
\multicolumn{11}{c}{\textit{\ourmodel{} (Qwen3 base model), Supervised Fine-Tuning}} \\
\midrule

C3LM-Qwen3-CREED-CCV            & 58\% & 1.83 & 1.50 & 1.28 & 0.85 & 46\% & 3.10 & 2.12 & 1.64 & 0.94 \\ 
C3LM-Qwen3-CREED-CCV+USPTO             & 39\% & 1.73 & 1.20 & 0.88 & 0.51 & 25\% & 4.09 & 1.99 & 1.26 & 0.64 \\ 
\makecell[l]{C3LM-Qwen3-CREED-CCV+\\USPTO-AR}             & 55\% & 1.88 & 1.48 & 1.18 & 0.66 & 40\% & 4.26 & 2.50 & 1.68 & 0.86 \\ 

\midrule
\multicolumn{11}{c}{\textit{\ourmodel{} (Qwen3 base model), Reinforcement Learning Fine-Tuning}} \\
\midrule

C3LM-Qwen3-RFT-CC             & 60\% & 2.01 & 1.30 & 0.86 & 0.44 & 59\% & 3.98 & 1.95 & 1.22 & 0.61 \\ 
C3LM-Qwen3-RFT-MA             & 65\% & 1.82 & 1.44 & 1.12 & 0.60 & 59\% & 2.97 & 1.89 & 1.31 & 0.67 \\ 
C3LM-Qwen3-RFT-MT             & 67\% & 1.40 & 1.00 & 0.72 & 0.38 & 55\% & 3.43 & 1.71 & 1.08 & 0.54 \\

\bottomrule
\end{tabular}

\caption{Ablation studies on the C3LM family. We vary three orthogonal design axes: (\textit{i}) training corpus (CREED vs.\ CREED-CCV, both optionally combined with decontaminated USPTO-full), (\textit{ii}) reasoning mode (Basic vs.\ Advanced, see Appendix~\ref{app:model_inventory}), and (\textit{iii}) reinforcement-learning reward (ChemCensor, Multi-GT Exact Match, or Molecular Transformer). The configuration reported in the main paper (Table~\ref{tab:chemcensor_benchmark_results}) corresponds to the leakage-safe subset of this grid; here we extend the analysis to all C3LM-LFM2 and C3LM-Qwen3 variants.}
\label{tab:chemcensor_benchmark_results_app}
\end{table*}

The main-paper benchmark (\autoref{sec:experiments}) is intentionally restricted to C3LM variants that avoid any circular dependence between the training signal and the evaluation metric. Concretely, the reported C3LM models are trained on CREED (verified by an in-house virtual synthesis engine but \emph{not} filtered by ChemCensor) rather than on CREED-CCV, and the reinforcement-learning rewards considered there exclude ChemCensor itself. This conservative setup is designed to answer a single question: \emph{can a model trained on a large, plausibility-aware reaction corpus outperform foundation and chemistry-specialist baselines under a plausibility-based metric, without the metric appearing anywhere in the training pipeline?} The results in the main paper show that the answer is positive. These results are additionally supported by even better performance of \texttt{C3LM-LFM2-CREED} on the extended version of ChemCensor benchmark (ChemCensor-U2P2, see \autoref{app:chemcensor_u2p2}) on the OOD URSA-expert-2026 set as compared to \texttt{C3LM-LFM2-USPTO}. 

In this section, we go beyond that conservative regime and analyze the full C3LM design space, including configurations that deliberately couple training and evaluation through ChemCensor. The motivation is twofold: first, to quantify how much additional headroom a plausibility-aware training signal provides; second, to make the resulting limitations explicit, so that practitioners can decide which checkpoint is appropriate for their downstream use case.

\paragraph{Logical progression of the ablation.}
We structure the ablation as a four-step argument.

\textbf{Step~1. ChemCensor as a plausibility proxy.}
In \autoref{app:chemcensor_validation} we show that ChemCensor agrees with expert chemist judgment on a stratified sample of model outputs at $87\%$ accuracy (\autoref{tab:conf_matrix}). This level of agreement justifies treating ChemCensor as a reasonable, albeit imperfect, proxy for human-perceived chemical plausibility, and as a meaningful target for both evaluation and training.

\textbf{Step~2. Training on CREED, evaluating with ChemCensor (no circularity).}
The main-paper C3LM checkpoints are trained on CREED, whose construction relies on the in-house virtual synthesis engine and expert-coded templates but does \emph{not} use ChemCensor as a filter. Evaluating these models with ChemCensor therefore provides an independent test of the training signal: improvements over the baselines (\autoref{tab:chemcensor_benchmark_results}) cannot be attributed to the model having seen the metric during training. This step establishes that a plausibility-aware corpus alone, without any explicit exposure to the metric, already moves models substantially closer to the upper region of the ChemCensor scoreboard.

\textbf{Step~3. Training on CREED-CCV for practical utility.}
Given that ChemCensor is a useful proxy (Step~1), filtering CREED with ChemCensor to obtain CREED-CCV produces a smaller but more uniformly plausible training corpus ($6.38$M vs.\ $22.71$M reactions). C3LM models trained on CREED-CCV in \autoref{tab:chemcensor_benchmark_results_app}) achieve the highest ChemCensor scores on both URSA-expert-2026 and USPTO-50K-test. Two observations follow. First, the gains over CREED-only training confirm that the model is, in fact, learning the plausibility signal that ChemCensor measures, rather than memorizing arbitrary regularities of the corpus. Second, these checkpoints are the most practically useful members of the C3LM family for downstream synthesis planning, because the very property a chemist cares about, that proposed disconnections be backed by documented precedents, is the property the model has been optimized for.

\textbf{Step~4. Acknowledging and bounding the circularity.}
Because CREED-CCV is constructed using ChemCensor, and because some C3LM variants additionally use ChemCensor as an RL reward, the scores of CREED-CCV-trained checkpoints in \autoref{tab:chemcensor_benchmark_results_app} are partially self-referential: the metric is, to a degree, evaluating its own training signal. We make this circularity explicit rather than hiding it, and we mitigate its interpretive impact in three ways:
(i) the main-paper benchmark uses only the non-circular CREED-trained subset, so the central claim of the paper does not depend on CREED-CCV at all;
(ii) URSA-expert-2026 is constructed from novel, decontaminated targets with no overlap with USPTO, so the test molecules are out-of-domain relative to the ChemCensor reference dataset, limiting trivial leakage;
(iii) the expert validation in \autoref{app:chemcensor_validation} provides an independent, metric-free sanity check on representative outputs of both CREED- and CREED-CCV-trained variants.

\paragraph{Toward a metric-independent re-evaluation.}
The current best-available plausibility proxy is ChemCensor itself, and so the practically strongest C3LM checkpoints (those trained on CREED-CCV, optionally with the ChemCensor reward) cannot be fully decoupled from it within this paper. We see this as an open problem rather than a closed one. When a stronger or independently developed plausibility metric becomes available (for example, one grounded in quantum chemistry), every row of \autoref{tab:chemcensor_benchmark_results_app} can be re-scored, and the relative merits of the CREED-only and CREED-CCV-trained checkpoints can be re-assessed without the circularity caveat. \autoref{app:chemcensor_u2p2} reports a first step in this direction, in which a subset of reactions is re-scored against a Pistachio-derived reference database; we encourage future work to extend this kind of cross-reference evaluation systematically.

\paragraph{Take-aways.}
From a methodological standpoint, the main-paper setup is the correct one for evaluating the contribution of CREED as a training resource. From a deployment standpoint, the CREED-CCV-trained variants in this appendix are the more capable models, and chemists who treat ChemCensor as a useful (if imperfect) proxy for plausibility should prefer them. The two perspectives are complementary, and \autoref{tab:chemcensor_benchmark_results_app} reports the data needed to support both.
\newpage

\section{Reference-Database Sensitivity of ChemCensor}
\label{app:chemcensor_u2p2}

ChemCensor is a precedent-based metric: a reaction is judged plausible if its reaction center and functional-group context have been observed in a reference database of confirmed transformations. The version used throughout the main paper, denoted \emph{ChemCensor-U2}, is grounded on USPTO-full (1976--2016), the largest publicly available reaction corpus. USPTO-full, however, is a strict subset of the documented chemical literature, and a reaction that is well established in proprietary databases such as Reaxys or SciFinder, or in more recent patent corpora, can in principle receive a zero score under ChemCensor-U2 simply because the corresponding RC/FG combination has no precedent in USPTO-full. In this section, we quantify the magnitude of this coverage effect and characterize its impact on the benchmark.

\textbf{Extended reference database.}

To probe the sensitivity of ChemCensor to the choice of reference corpus, we rebuilt the relational RC/FG database using Pistachio (Q3~2023) in addition to USPTO-full, yielding an extended version \emph{ChemCensor-U2P2} (USPTO-full $\cup$ Pistachio). Pistachio adds patent reactions from 2017--2023 that lie outside the USPTO-full coverage window, and contributes additional functional-group annotations for reaction centers that are already present in USPTO-full. Re-scoring is performed on the same set of model predictions used in the main benchmark, so the only variable between U2 and U2P2 is the reference database.

\textbf{Magnitude of the coverage effect.}

Of the reactions generated by C3LM that receive a score of $\mathrm{CC}=0$ under ChemCensor-U2, $8.6\%$ are reassigned $\mathrm{CC}>0$ under ChemCensor-U2P2. For the strongest non-C3LM baseline (Gemini~3~Flash), the corresponding figure is $6.0\%$. Inspection of the reassigned reactions shows that the increase in pass rate is driven primarily by the annotation of additional non-reacting functional groups in the Pistachio-derived FG signatures, rather than by the appearance of entirely new reaction centers. Representative cases are shown in \autoref{tab:examples_u2p2}: for example, an aromatic amine in a bromo-Suzuki coupling, or a tertiary alcohol in a urea synthesis from a carbamate substrate — combinations that are absent from the USPTO-full FG annotations for the corresponding RCs but well represented in Pistachio.

A symmetric experiment using Reaxys or CAS SciFinder as the extending corpus is not feasible within the licensing terms of these providers, which preclude the large-scale export of reaction records required to rebuild a relational database. The Pistachio extension, therefore, captures the incremental gain from adding a further public-domain corpus and provides a lower bound on the total residual coverage gap of ChemCensor-U2 relative to the full body of documented synthetic chemistry.

\textbf{Effect on benchmark conclusions.}

\autoref{tab:u2p2} reports the complete set of ChemCensor metrics for all benchmarked models re-scored under ChemCensor-U2P2, directly comparable to Table~\ref{tab:chemcensor_benchmark_results} in the main paper. Two properties of the U2 $\to$ U2P2 transition are notable. First, the relative ranking of models is preserved: checkpoints that score well under the publicly reproducible U2 setting also score well under the extended U2P2 setting, and the ordering between proprietary foundation, open-weight foundation, chemical specialist, and C3LM groups is unchanged. Second, the absolute gap between the best C3LM checkpoint and the strongest non-C3LM baselines persists, indicating that the improvements reported in the main paper reflect a genuine difference in the quality of generated reactions rather than a particular alignment between C3LM outputs and USPTO-full coverage.

\textbf{Practical guidance.}

For practitioners deploying ChemCensor as a plausibility filter, the U2 $\to$ U2P2 comparison suggests a simple rule of thumb: a single-zero score on the public ChemCensor-U2 should be read as ``no precedent found in USPTO-full'' rather than as a categorical claim of chemical implausibility, and is most informative when interpreted together with the specific failure mode reported by the module (missing RC vs.\ FG incompatibility). When access to a richer reference corpus is available, rebuilding the relational database with that corpus closes part of the coverage gap, as demonstrated by the U2P2 results, without requiring any change to the underlying RC/FG extraction pipeline.

\clearpage
\vspace*{\fill}
\begin{table}[H]
\centering
\begin{tabular}{@{}l|c|cccc|c|cccc@{}}
\toprule
\multirow{2}{*}{\textbf{Model}} &
\multicolumn{5}{c|}{\textbf{URSA-expert-2026}} &
\multicolumn{5}{c}{\textbf{USPTO-50K-test}} \\
\cline{2-6}\cline{7-11}
& \multirow{2}{*}{\textbf{Unique}} & \multirow{2}{*}{\textbf{Max}} & \multicolumn{3}{c|}{\textbf{Av. PT-Top-K CC}}
& \multirow{2}{*}{\textbf{Unique}} & \multirow{2}{*}{\textbf{Max}} & \multicolumn{3}{c}{\textbf{Av. PT-Top-K CC}} \\
&  &  & \textbf{@3} & \textbf{@5} & \textbf{@10}
&  &  & \textbf{@3} & \textbf{@5} & \textbf{@10} \\
\midrule
\multicolumn{11}{c}{\textit{Proprietary Foundation Models}} \\
\midrule
Grok-4.1                   & 43\% & 1.97 & 1.45 & 1.09 & 0.58 & --   & --   & --   & --   & --   \\
Gemini 2.5 Flash           & 50\% & 0.76 & 0.34 & 0.20 & 0.10 & --   & --   & --   & --   & --   \\
Gemini 3 Flash preview     & 40\% & 2.03 & 1.39 & 0.97 & 0.49 & --   & --   & --   & --   & --   \\
GPT 5.1                    & 62\% & 0.82 & 0.41 & 0.26 & 0.13 & 59\% & 1.74 & 0.87 & 0.55 & 0.28 \\
GPT 5.2                    & 51\% & 0.93 & 0.52 & 0.33 & 0.16 & 52\% & 2.24 & 1.19 & 0.75 & 0.38 \\
Claude 4.5 Sonnet          & 70\% & 1.59 & 0.99 & 0.66 & 0.33 & 56\% & 3.55 & 2.04 & 1.33 & 0.68 \\
Claude 4.5 Opus            & 54\% & 1.51 & 0.82 & 0.53 & 0.27 & 44\% & 3.48 & 1.85 & 1.17 & 0.59 \\
\midrule
\multicolumn{11}{c}{\textit{Open-weight Foundation Models}} \\
\midrule
DeepSeek 3.2               & 52\% & 0.50 & 0.21 & 0.12 & 0.06 & 55\% & 1.42 & 0.64 & 0.39 & 0.20 \\
Qwen3 8B \                 & 52\% & 0.00 & 0.00 & 0.00 & 0.00 & 55\% & 0.06 & 0.04 & 0.03 & 0.02 \\
Qwen3 14B \                & 52\% & 0.01 & 0.00 & 0.00 & 0.00 & 54\% & 0.15 & 0.05 & 0.03 & 0.02 \\
Kimi K2                    & 43\% & 1.52 & 0.98 & 0.65 & 0.34 & --   & --   & --   & --   & --   \\
LFM2 2.6B                  & 0\%  & 0.00 & 0.00 & 0.00 & 0.00 & 0\%  & 0.00 & 0.00 & 0.00 & 0.00 \\
\midrule
\multicolumn{11}{c}{\textit{Open-weight Chemical Specialist Models}} \\
\midrule
ether0                     & 20\% & 1.20 & 0.68 & 0.43 & 0.22 & 19\% & 2.29 & 1.21 & 0.76 & 0.38 \\
NatureLM                   & 27\% & 1.76 & 1.12 & 0.71 & 0.36 & 20\% & 4.11 & 1.97 & 1.23 & 0.62 \\
RetroDFM-R                 & 12\% & 1.76 & 0.79 & 0.48 & 0.24 & 8\%  & 4.44 & 1.60 & 0.96 & 0.48 \\
\midrule
\multicolumn{11}{c}{\textit{\ourmodel{}, Supervised Fine-Tuning}} \\
\midrule
\ourmodel{}-LFM2-USPTO            & 40\% & 1.63 & 0.96 & 0.61 & 0.31 & 23\% & 4.14 & 1.94 & 1.20 & 0.60 \\ 
\ourmodel{}-LFM2-CREED            & 58\% & 1.72 & 1.23 & 0.93 & 0.50 & 51\% & 2.03 & 1.25 & 0.87 & 0.45 \\ 
\ourmodel{}-LFM2-CREED+USPTO      & 42\% & 1.91 & 1.35 & 0.95 & 0.49 & 24\% & 4.31 & 2.18 & 1.39 & 0.70 \\
\midrule
\multicolumn{11}{c}{\textit{\ourmodel{}, Reinforcement Learning Fine-Tuning}} \\
\midrule
\ourmodel{}-LFM2-RFT-MA      & 42\% & 1.93 & 1.32 & 0.93 & 0.48 & 26\% & 4.17 & 2.17 & 1.39 & 0.70 \\
\ourmodel{}-LFM2-RFT-MT      & 39\% & 2.06 & 1.41 & 1.03 & 0.53 & 23\% & 4.27 & 2.25 & 1.44 & 0.73 \\
\end{tabular}

\caption{ChemCensor metrics under the extended ChemCensor-U2P2 reference database (USPTO-full $\cup$ Pistachio~Q3~2023). Columns and aggregated metrics are defined as in \autoref{tab:chemcensor_benchmark_results}.}
\label{tab:u2p2}
\end{table}
\vspace*{\fill}
\clearpage

\includepdf[
  pages=1,
  offset=0 -14mm,
  scale=0.92,
  pagecommand={%
    \thispagestyle{plain}%
    \captionsetup{type=table}%
    \vspace*{0em}%
    \caption{Representative reactions that receive $\mathrm{CC}=0$ under ChemCensor-U2 (USPTO-full reference) but $\mathrm{CC}\geq 1$ under ChemCensor-U2P2 (USPTO-full~$\cup$~Pistachio Q3~2023 reference). Examples \#1--3 are from C3LM; examples \#4--6 are from Gemini~3~Flash. In each case, the reaction center is shared between the model output and a Pistachio precedent, and the score increase is driven by the annotation of additional non-reacting functional groups in the extended FG signature.}%
    \label{tab:examples_u2p2}%
    \vspace*{0em}%
  }
]{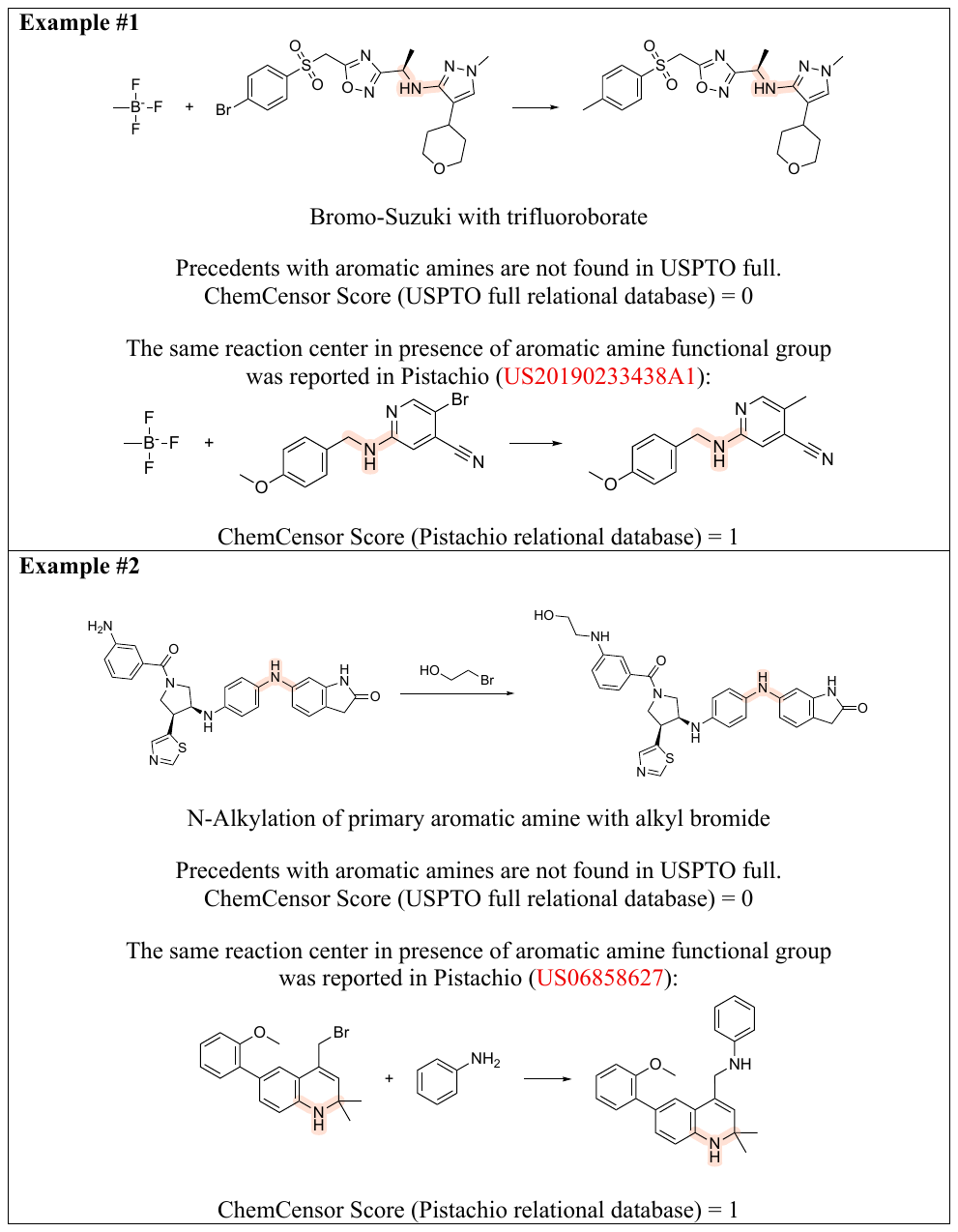}
\includepdf[
  pages=2-,
  scale=0.92,
  offset=0 -4mm,
  pagecommand={%
    \thispagestyle{plain}%
    \captionsetup{type=table}%
    \caption*{Continuation of \autoref{tab:examples_u2p2}.}%
  }
]{figs/appendix/rev2-ans2_1.pdf}
\section{Evaluation of Conventional Single-Step Retrosynthesis Models}
\label{app:conventional_models}

The main paper focuses on LLM-based approaches to single-step retrosynthesis, both general-purpose and chemistry-specialized. A complementary body of work, predating the LLM era and still widely deployed in production synthesis-planning pipelines, comprises template-based, graph-based, and sequence-to-sequence models trained specifically on USPTO-derived reaction sets (\autoref{sec:related_work}). To place the LLM benchmark of the main paper in a broader methodological context, we evaluate the principal representatives of this conventional family under the same ChemCensor protocol used elsewhere in this work.

\paragraph{Models and evaluation protocol.}
We benchmark eight conventional SSRS models accessible through Syntheseus~\citep{maziarz2025syntheseus}: LocalRetro, GLN, MEGAN, Chemformer, Graph2Edits, MHNreact, RetroKNN, and R-SMILES. Each model was used in its publicly released configuration, with predictions generated for both URSA-expert-2026 and USPTO-50K-test.

\paragraph{Results under ChemCensor-U2.}
\autoref{tab:conv_u2} reports performance under the publicly reproducible ChemCensor-U2 reference database (USPTO-full, 1976--2016). Conventional models achieve uniformly high fractions of unique valid predictions, which is expected given that they are trained explicitly on USPTO-derived reaction data and produce structured outputs by construction: R-SMILES reaches $100\%$ uniqueness on both benchmarks, and MHNreact, RetroKNN, LocalRetro, and Graph2Edits exceed $85\%$ on both. The Av.~PT-Max~CC values cluster tightly between $2.0$ and $2.2$ on URSA-expert-2026, with R-SMILES and Graph2Edits leading at $2.10$ and $2.08$ respectively; on USPTO-50K-test, the same models reach Av.~PT-Max~CC values in the $4.77$--$4.84$ range. Chemformer is a clear outlier, with markedly lower uniqueness ($24\%$ on URSA-expert-2026, $10\%$ on USPTO-50K-test) and correspondingly reduced Av.~PT-Top-K metrics, consistent with the mode-collapse behavior reported for that architecture in prior work.

\paragraph{Results under ChemCensor-U2P2.}
\autoref{tab:conv_u2p2} reports the same models re-scored against the extended ChemCensor-U2P2 reference database (USPTO-full $\cup$ Pistachio Q3~2023; see Appendix~\ref{app:chemcensor_u2p2}). As in the LLM case, the relative ordering of conventional models is preserved between U2 and U2P2, and absolute scores increase modestly across all checkpoints. The pattern of increase mirrors that observed for the LLM benchmark: gains accrue primarily through additional FG annotations rather than through newly recognized reaction centers, indicating that conventional models, like their LLM counterparts, occasionally produce reactions whose RC is well established in USPTO-full but whose FG context is documented only in more recent or broader corpora.

\paragraph{Conventional models in context.}
Two observations are useful for interpreting the conventional benchmark alongside the LLM benchmark of the main paper. First, the high uniqueness rates of conventional models reflect the structured nature of their output spaces (template inventories, graph edit grammars, constrained SMILES decoders), and not necessarily a higher chemical quality of individual predictions; the per-target Max~CC and Av.~PT-Top-K metrics provide the more directly comparable signal across architectures. Second, the strongest conventional models attain Av.~PT-Max~CC values on URSA-expert-2026 that are comparable to those of the best LLM checkpoints reported in the main paper (Table~\ref{tab:chemcensor_benchmark_results}), with the LLM checkpoints showing an advantage on the per-target Top-K averages — consistent with the broader exploratory behavior expected from generative LLMs versus the more concentrated output distributions of template- and graph-based models. The two model families are therefore best viewed as complementary tools within a synthesis-planning pipeline rather than as direct competitors, and the ChemCensor framework provides a common evaluation surface on which both can be measured.

\begin{table*}[th!]
\centering
\begin{tabular}{@{}l|c|cccc|c|cccc@{}}
\toprule
\multirow{2}{*}{\textbf{Model}} &
\multicolumn{5}{c|}{\textbf{URSA-expert-2026}} &
\multicolumn{5}{c}{\textbf{USPTO-50K-test}} \\
\cline{2-6}\cline{7-11}
& \multirow{2}{*}{\textbf{Unique}} & \multirow{2}{*}{\textbf{Max}} & \multicolumn{3}{c|}{\textbf{Av. PT-Top-K CC}}
& \multirow{2}{*}{\textbf{Unique}} & \multirow{2}{*}{\textbf{Max}} & \multicolumn{3}{c}{\textbf{Av. PT-Top-K CC}} \\
\cline{4-6}\cline{9-11}
&  &  & \textbf{@3} & \textbf{@5} & \textbf{@10}
&  &  & \textbf{@3} & \textbf{@5} & \textbf{@10} \\
\midrule
\multicolumn{11}{c}{\textit{SSRS Conventional Models}} \\
\midrule
LocalRetro          & 90\% & \textbf{2.14} & \underline{1.87} & 1.61 & 1.23 & 85\% & 4.79 & 3.35 & 2.73 & 1.89 \\
GLN                 & 71\% & 1.97 & 1.73 & 1.50 & 1.04 & 69\% & 4.80 & 3.25 & 2.60 & 1.68 \\
MEGAN               & 83\% & 2.03 & 1.76 & 1.50 & 1.01 & 81\% & 4.77 & 3.25 & 2.58 & 1.68 \\
Chemformer          & 24\% & 1.77 & 1.02 & 0.68 & 0.35 & 10\% & 4.66 & 1.73 & 1.05 & 0.52 \\
Graph2Edits         & 88\% & 2.08 & 1.86 & 1.63 & 1.28 & 72\% & \underline{4.81} & 3.11 & 2.41 & 1.52 \\
MHNreact            & \underline{99\%} & 2.08 & 1.86 & \underline{1.63} & \textbf{1.28} & \underline{98\%} & \textbf{4.83} & 3.35 & \underline{2.75} & \textbf{1.99} \\
RetroKNN            & 91\% & \underline{2.13} & 1.86 & 1.62 & 1.23 & 87\% & \underline{4.82} & \underline{3.36} & 2.73 & 1.90 \\
R-SMILES         & \textbf{100\%} & 2.10 & \textbf{1.88} & \textbf{1.64} & \underline{1.24} & \textbf{100\%} & \textbf{4.84} & \textbf{3.45} & \textbf{2.81} & \underline{1.93} \\
\bottomrule
\end{tabular}
\caption{Performance of conventional single-step retrosynthesis models under the ChemCensor-U2 reference database (USPTO-full, 1976--2016). Columns and aggregated metrics are defined as in \autoref{tab:chemcensor_benchmark_results}. All models were evaluated through the Syntheseus interface \cite{maziarz2025syntheseus} using publicly released checkpoints. Best values per column are shown in \textbf{bold}, second-best are \underline{underlined}.}
\label{tab:conv_u2}
\end{table*}

\begin{table*}[th!]
\centering
\begin{tabular}{@{}l|c|cccc|c|cccc@{}}
\toprule
\multirow{2}{*}{\textbf{Model}} &
\multicolumn{5}{c|}{\textbf{URSA-expert-2026}} &
\multicolumn{5}{c}{\textbf{USPTO-50K-test}} \\
\cline{2-6}\cline{7-11}
& \multirow{2}{*}{\textbf{Unique}} & \multirow{2}{*}{\textbf{Max}} & \multicolumn{3}{c|}{\textbf{Av. PT-Top-K CC}}
& \multirow{2}{*}{\textbf{Unique}} & \multirow{2}{*}{\textbf{Max}} & \multicolumn{3}{c}{\textbf{Av. PT-Top-K CC}} \\
\cline{4-6}\cline{9-11}
&  &  & \textbf{@3} & \textbf{@5} & \textbf{@10}
&  &  & \textbf{@3} & \textbf{@5} & \textbf{@10} \\
\midrule
\multicolumn{11}{c}{\textit{SSRS Conventional Models}} \\
\midrule
LocalRetro          & 90\% & \textbf{2.37} & \underline{2.12} & 1.85 & 1.41 & 85\% & 4.84 & 3.52 & 2.92 & 2.09 \\
GLN                 & 71\% & 2.24 & 1.93 & 1.68 & 1.21 & 69\% & 4.86 & 3.42 & 2.79 & 1.87 \\
MEGAN               & 83\% & \underline{2.34} & 1.99 & 1.69 & 1.18 & 81\% & 4.83 & 3.41 & 2.77 & 1.86 \\
Chemformer          & 24\% & 1.96 & 1.14 & 0.77 & 0.39 & 10\% & 4.72 & 1.78 & 1.08 & 0.54 \\
Graph2Edits         & 88\% & 2.37 & 2.10 & 1.78 & 1.29 & 72\% & \underline{4.87} & 3.28 & 2.59 & 1.69 \\
MHNreact            & \underline{99\%} & 2.31 & 2.08 & \underline{1.82} & \textbf{1.45} & \underline{98\%} & \textbf{4.89} & 3.52 & \underline{2.94} & \textbf{2.19} \\
RetroKNN            & 91\% & \underline{2.34} & 2.11 & 1.83 & 1.42 & 87\% & \underline{4.87} & \underline{3.53} & 2.93 & 2.11 \\
R-SMILES         & \textbf{100\%} & 2.36 & \textbf{2.14} & \textbf{1.89} & \underline{1.42} & \textbf{100\%} & \textbf{4.89} & \textbf{3.61} & \textbf{3.00} & \underline{2.13} \\
\bottomrule
\end{tabular}
\caption{Performance of conventional single-step retrosynthesis models under the extended ChemCensor-U2P2 reference database (USPTO-full $\cup$ Pistachio Q3~2023). Columns and aggregated metrics are defined as in Table~\ref{tab:chemcensor_benchmark_results}; the reference-database extension is described in Appendix~\ref{app:chemcensor_u2p2}. Best values per column are shown in \textbf{bold}, second-best are \underline{underlined}.}
\label{tab:conv_u2p2}
\end{table*}
\newpage
\clearpage
\section{A Cost-Efficient Subset of USPTO-50K-test for LLM Evaluation}
\label{app:uspto_50k_sample}

Running the full ChemCensor benchmark on USPTO-50K-test requires generating $15$ independent completions per target across $4{,}972$ curated targets (\autoref{app:uspto_50_k_test_mods}), which amounts to roughly $7.5 \cdot 10^{4}$ LLM completions per evaluated model. For proprietary frontier models, this scale is prohibitive in both wall-clock time and API cost -- and indeed, as noted in the main paper, this is precisely the reason we could not report USPTO-50K-test results for Grok-4.1, Gemini~2.5~Flash, or Gemini~3~Flash~preview. To make the benchmark practical for routine LLM evaluation without sacrificing statistical resolution, we additionally release a $10\%$ random sample of the curated USPTO-50K-test.

\paragraph{Construction.}
The sample, denoted \textbf{USPTO-50K-test-mini}, consists of $497$ targets drawn uniformly at random and without replacement from the $4{,}972$ targets of the curated USPTO-50K-test (\autoref{app:uspto_50_k_test_mods}). The dataset diversity is 0.85793 (see \autoref{eq:diversity}).

\paragraph{Usage.}
We recommend \textbf{USPTO-50K-test-mini} as the default USPTO-derived benchmark for LLM-based single-step retrosynthesis evaluation under typical API-cost or wall-clock constraints, in combination with the full URSA-expert-2026 set as the primary out-of-distribution benchmark. Reporting both numbers: the cost-efficient in-distribution score on USPTO-50K-test-mini and the leakage-resistant OOD score on URSA-expert-2026, gives a more complete picture of model behavior than either benchmark alone, at a fraction of the compute required by the full USPTO-50K-test.

\paragraph{Availability.} \textbf{USPTO-50K-test-mini} is available on Hugging Face: \url{https://huggingface.co/datasets/insilicomedicine/URSA-benchmarking-sets}.

\paragraph{Benchmark results on USPTO-50K-test-mini.}
\autoref{tab:uspto_50k_mini} reports ChemCensor scores on USPTO-50K-test-mini for every model evaluated in this work. 

\begin{table*}[th!]
\centering
\begin{tabular}{@{}l|c|cccc@{}}
\toprule
\multirow{2}{*}{\textbf{Model}} &
\multicolumn{5}{c}{\textbf{USPTO-50K-test-mini}} \\
\cline{2-6}
& \multirow{2}{*}{\textbf{Unique}} & \multirow{2}{*}{\textbf{Max}} & \multicolumn{3}{c}{\textbf{Av. PT-Top-K CC}} \\
&  &  & \textbf{@3} & \textbf{@5} & \textbf{@10} \\
\midrule
\multicolumn{6}{c}{\textit{Proprietary Foundation Models}} \\
\midrule
Grok-4.1                   & 37\% & 4.04 & 2.36 & 1.59 & 0.82 \\
Gemini 2.5 Flash           & 14\% & 1.16 & 0.49 & 0.30 & 0.15 \\
Gemini 3 Flash preview     & 28\% & 4.04 & 2.18 & 1.38 & 0.70 \\
GPT 5.1                    & 61\% & 1.46 & 0.69 & 0.42 & 0.21 \\
GPT 5.2                    & 51\% & 2.11 & 1.03 & 0.64 & 0.32 \\
Claude 4.5 Sonnet          & 56\% & 3.42 & 1.77 & 1.12 & 0.56 \\
Claude 4.5 Opus            & 44\% & 3.39 & 1.61 & 0.99 & 0.50 \\
\midrule
\multicolumn{6}{c}{\textit{Open-weight Foundation Models}} \\
\midrule
DeepSeek 3.2               & 56\% & 1.23 & 0.49 & 0.30 & 0.15 \\
Qwen3 8b                   & 18\% & 0.01 & 0.00 & 0.00 & 0.00 \\
Qwen3 14b                  & 20\% & 0.01 & 0.00 & 0.00 & 0.00 \\
Kimi K2                    & 44\% & 3.54 & 1.87 & 1.19 & 0.60 \\
LFM2 2.6b                  & 0\%  & 0.00 & 0.00 & 0.00 & 0.00 \\
\midrule
\multicolumn{6}{c}{\textit{Open-weight Chemical Specialist Models}} \\
\midrule
ether0                     & 18\% & 2.08 & 1.06 & 0.67 & 0.34 \\ 
NatureLM                   & 19\% & 4.15 & 1.93 & 1.19 & 0.60 \\ 
RetroDFM-R                 & 8\%  & 4.39 & 1.56 & 0.94 & 0.47 \\ 
\midrule
\multicolumn{6}{c}{\textit{\ourmodel{}, Supervised Fine-Tuning}} \\
\midrule
\ourmodel{}-LFM2-USPTO         & 22\% & 4.13 & 1.87 & 1.15 & 0.57 \\ 
\ourmodel{}-LFM2-CREED         & 50\% & 2.03 & 1.21 & 0.84 & 0.44 \\ 
\ourmodel{}-LFM2-CREED+USPTO   & 23\% & 4.27 & 2.05 & 1.29 & 0.65 \\
\midrule
\multicolumn{6}{c}{\textit{\ourmodel{}, Reinforcement Learning Fine-Tuning}} \\
\midrule
\ourmodel{}-LFM2-RFT-MA         & 26\% & 4.10 & 2.05 & 1.30 & 0.65 \\
\ourmodel{}-LFM2-RFT-MT         & 22\% & 4.23 & 2.10 & 1.33 & 0.67 \\

\midrule
\multicolumn{6}{c}{\textit{SSRS Conventional Models}} \\
\midrule

LocalRetro          & 85\% & 4.82 & 3.35 & 2.73 & 1.88 \\
GLN                 & 69\% & 4.81 & 3.23 & 2.58 & 1.66 \\
MEGAN               & 82\% & 4.82 & 3.26 & 2.59 & 1.57 \\
Chemformer          & 10\% & 4.70 & 1.73 & 1.05 & 0.53 \\
Graph2Edits         & 71\% & 4.80 & 3.12 & 2.40 & 1.51 \\
MHNreact            & 98\% & 4.84 & 3.34 & 2.75 & 1.98 \\
RetroKNN            & 87\% & 4.84 & 3.37 & 2.75 & 1.90 \\
R-SMILES            & 100\% & 4.87 & 3.51 & 2.86 & 1.95 \\

\midrule
\multicolumn{6}{c}{\textit{\ourmodel{}, Ablation Studies}} \\
\midrule

C3LM-LFM2-CREED-CCV+USPTO              & 31\% & 4.15 & 2.17 & 1.42 & 0.74 \\
C3LM-LFM2-CREED-CCV+USPTO-AR              & 57\% & 3.48 & 2.29 & 1.73 & 1.01 \\
C3LM-LFM2-AR-RFT-CC             & 41\% & 3.81 & 2.45 & 1.77 & 0.96 \\
C3LM-Qwen3-CREED-CCV             & 46\% & 3.07 & 2.12 & 1.63 & 0.93 \\
C3LM-Qwen3-CREED-CCV+USPTO             & 24\% & 4.12 & 1.99 & 1.26 & 0.64 \\
C3LM-Qwen3-CREED-CCV+USPTO-AR             & 38\% & 4.28 & 2.54 & 1.68 & 0.86 \\
C3LM-Qwen3-RFT-CC             & 59\% & 4.08 & 2.00 & 1.25 & 0.63 \\
C3LM-Qwen3-RFT-MA            & 58\% & 3.01 & 1.88 & 1.29 & 0.66 \\
C3LM-Qwen3-RFT-MT             & 55\% & 3.54 & 1.74 & 1.10 & 0.55 \\

\end{tabular}

\caption{USPTO-50K-test-mini benchmarking results (ChemCensor v.0.5.2-U2).}
\label{tab:uspto_50k_mini}
\end{table*}

The reduced compute footprint of the subset makes it tractable to extend the in-distribution benchmark to the proprietary frontier models (Grok-4.1, Gemini~2.5~Flash, Gemini~3~Flash~preview) whose API-cost and rate-limit profiles precluded full USPTO-50K-test evaluation in the main paper, placing them on a common scoreboard with the rest of the field. The model rankings observed on USPTO-50K-test-mini are consistent with those on the full USPTO-50K-test for the subset of models evaluated on both, confirming that the $10\%$ random sample preserves the discriminative structure of the parent benchmark at group level.
\section{Statistical Robustness of the ChemCensor Benchmark}
\label{app:cis}

The main ChemCensor benchmark results reported in \autoref{tab:chemcensor_benchmark_results} are averages over a finite number of target molecules: $n=100$ for URSA-expert-2026 and $n=4{,}972$ for the curated USPTO-50K-test. To quantify the sampling uncertainty of the primary model-level comparisons, we report 95\% confidence intervals (CIs) for this table. We do not provide CIs for all supplementary tables, since those tables are used for auxiliary analyses and ablation studies.

\paragraph{Estimation procedure.}
For USPTO-50K-test ($n=4{,}972$), CIs are estimated using the normal approximation:
\begin{equation}
\mathrm{CI}_{95\%} = \bar{x} \pm 1.96 \cdot \frac{s}{\sqrt{n}},
\end{equation}
where $\bar{x}$ and $s$ denote the sample mean and standard deviation of the per-target metric across all targets. The large sample size makes the normal approximation appropriate for the reported aggregate metrics.

For URSA-expert-2026 ($n=100$), we estimate CIs using nonparametric bootstrap with $B=2{,}000$ replicates. At each replicate, target molecules are resampled with replacement, the aggregate metric is recomputed, and the $2.5$th and $97.5$th percentiles of the bootstrap distribution are reported as the lower and upper CI bounds. This procedure avoids distributional assumptions and is suitable for bounded ChemCensor scores in the $[0,5]$ range.

All CIs are computed independently for each model and each metric. We do not apply multiplicity corrections; therefore, the intervals should be interpreted as marginal uncertainty estimates rather than simultaneous confidence bands.

\begin{sidewaystable*}
    \centering
    \footnotesize
    \begin{tabular}{@{}l|c|cccc|c|cccc@{}}
\toprule
\multirow{2}{*}{\textbf{Model}} &
\multicolumn{5}{c|}{\textbf{URSA-expert-2026}} &
\multicolumn{5}{c}{\textbf{USPTO-50K-test}} \\
\cline{2-6}\cline{7-11}
& \multirow{2}{*}{\textbf{Unique}} & \multirow{2}{*}{\textbf{Max}} & \multicolumn{3}{c|}{\textbf{Av. PT-Top-K CC}}
& \multirow{2}{*}{\textbf{Unique}} & \multirow{2}{*}{\textbf{Max}} & \multicolumn{3}{c}{\textbf{Av. PT-Top-K CC}} \\
\cline{4-6}\cline{9-11}
&  &  & \textbf{@3} & \textbf{@5} & \textbf{@10}
&  &  & \textbf{@3} & \textbf{@5} & \textbf{@10} \\
\midrule
\multicolumn{11}{c}{\textit{Proprietary Foundation Models}} \\
\midrule
Grok-4.1                   & 43\% & 1.75 $\pm$ 0.18 & 1.29 $\pm$ 0.13 & 0.93 $\pm$ 0.10 & 0.49 $\pm$ 0.06 & -- & -- & -- & -- & -- \\
Gemini 2.5 Flash           & 50\% & 0.66 $\pm$ 0.16 & 0.30 $\pm$ 0.18 & 0.08 $\pm$ 0.18 & 0.02 $\pm$ 0.18 & -- & -- & -- & -- & -- \\
Gemini 3 Flash preview     & 40\% & 1.80 $\pm$ 0.18 & 1.21 $\pm$ 0.14 & 0.83 $\pm$ 0.11 & 0.42 $\pm$ 0.06 & -- & -- & -- & -- & -- \\
GPT 5.1                    & 62\% & 0.73 $\pm$ 0.16 & 0.35 $\pm$ 0.08 & 0.22 $\pm$ 0.08 & 0.11 $\pm$ 0.03 & 
                            \textbf{59\%} & 1.54 $\pm$ 0.05 & 0.72 $\pm$ 0.03 & 0.45 $\pm$ 0.02 & 0.22 $\pm$ 0.01 \\
GPT 5.2                    & 51\% & 0.85 $\pm$ 0.16 & 0.47 $\pm$ 0.10 & 0.29 $\pm$ 0.06 & 0.15 $\pm$ 0.03 & 
                             52\% & 2.04 $\pm$ 0.05 & 1.01 $\pm$ 0.03 & 0.63 $\pm$ 0.02 & 0.32 $\pm$ 0.01 \\
Claude 4.5 Sonnet          & 70\% & 1.45 $\pm$ 0.19 & 0.88 $\pm$ 0.15 & 0.58 $\pm$ 0.11 & 0.29 $\pm$ 0.06 & 
                            \underline{56\%} & 3.37 $\pm$ 0.05 & 1.81 $\pm$ 0.04 & 1.15 $\pm$ 0.03 & 0.58 $\pm$ 0.01 \\
Claude 4.5 Opus            & 54\% & 1.34 $\pm$ 0.20 & 0.73 $\pm$ 0.13 & 0.46 $\pm$ 0.09 & 0.23 $\pm$ 0.05 & 
                             44\% & 3.31 $\pm$ 0.05 & 1.62 $\pm$ 0.04 & 1.00 $\pm$ 0.03 & 0.50 $\pm$ 0.01 \\
\midrule
\multicolumn{11}{c}{\textit{Open-weight Foundation Models}} \\
\midrule
DeepSeek 3.2               & 52\% & 0.39 $\pm$ 0.13 & 0.16 $\pm$ 0.06 & 0.10 $\pm$ 0.04 & 0.05 $\pm$ 0.02 & 
                             55\% & 1.14 $\pm$ 0.05 & 0.46 $\pm$ 0.02 & 0.27 $\pm$ 0.01 & 0.14 $\pm$ 0.01 \\
Qwen3 8B \                 & 52\% & 0.00 $\pm$ 0.00 & 0.00 $\pm$ 0.00 & 0.00 $\pm$ 0.00 & 0.00 $\pm$ 0.00 & 
                             55\% & 0.04 $\pm$ 0.01 & 0.03 $\pm$ 0.01 & 0.02 $\pm$ 0.01 & 0.01 $\pm$ 0.00 \\
Qwen3 14B \                & 52\% & 0.01 $\pm$ 0.04 & 0.00 $\pm$ 0.01 & 0.00 $\pm$ 0.00 & 0.00 $\pm$ 0.00 & 
                             54\% & 0.10 $\pm$ 0.01 & 0.03 $\pm$ 0.01 & 0.02 $\pm$ 0.00 & 0.01 $\pm$ 0.00 \\
Kimi K2                    & 43\% & 1.12 $\pm$ 0.14 & 0.62 $\pm$ 0.10 & 0.38 $\pm$ 0.07 & 0.19 $\pm$ 0.04 & -- & -- & -- & -- & -- \\
LFM2 2.6B                  & 0\%  & 0.00 $\pm$ 0.00 & 0.00 $\pm$ 0.00 & 0.00 $\pm$ 0.00 & 0.00 $\pm$ 0.00 & 
                             0\%  & 0.00 $\pm$ 0.00 & 0.00 $\pm$ 0.00 & 0.00 $\pm$ 0.00 & 0.00 $\pm$ 0.00 \\
\midrule
\multicolumn{11}{c}{\textit{Open-weight Chemical Specialist Models}} \\
\midrule
ether0                     & 20\% & 1.01 $\pm$ 0.14 & 0.55 $\pm$ 0.09 & 0.35 $\pm$ 0.06 & 0.17 $\pm$ 0.03 & 
                             19\% & 2.08 $\pm$ 0.05 & 1.07 $\pm$ 0.03 & 0.67 $\pm$ 0.02 & 0.34 $\pm$ 0.01 \\
NatureLM                   & 27\% & 1.57 $\pm$ 0.16 & 0.97 $\pm$ 0.12 & 0.61 $\pm$ 0.09 & 0.30 $\pm$ 0.04 & 
                             20\% & 3.99 $\pm$ 0.04 & 1.85 $\pm$ 0.03 & 1.14 $\pm$ 0.02 & 0.57 $\pm$ 0.01 \\
RetroDFM-R                 & 12\% & 1.52 $\pm$ 0.19 & 0.70 $\pm$ 0.10 & 0.43 $\pm$ 0.06 & 0.21 $\pm$ 0.03 & 
                             8\% & \textbf{4.35} $\pm$ 0.04 & 1.56 $\pm$ 0.02 & 0.94 $\pm$ 0.01 & 0.47 $\pm$ 0.00 \\
\midrule
\multicolumn{11}{c}{\textit{\ourmodel{}}, Supervised Fine-Tuning} \\
\midrule
\ourmodel{}-LFM2-USPTO         & 40\% & 1.63 $\pm$ 0.20 & 0.96 $\pm$ 0.13 & 0.61 $\pm$ 0.09 & 0.31 $\pm$ 0.04 & 23\% & 4.02 $\pm$ 0.04 & 1.83 $\pm$ 0.04 & 1.13 $\pm$ 0.05 & 0.57 $\pm$ 0.05 \\ 
\ourmodel{}-LFM2-CREED         & 58\% & 1.52 $\pm$ 0.19 & 1.05 $\pm$ 0.13 & 0.75 $\pm$ 0.10 & 0.38 $\pm$ 0.05 & 51\% & 2.03 $\pm$ 0.05 & 1.25 $\pm$ 0.04 & 0.87 $\pm$ 0.03 & 0.45 $\pm$ 0.02\\
\ourmodel{}-LFM2-CREED+USPTO         & 42\% & 1.69 $\pm$ 0.18 & 1.14 $\pm$ 0.14 & 0.78 $\pm$ 0.11 & 0.40 $\pm$ 0.06 & 24\% & 4.21 $\pm$ 0.04 & 2.06 $\pm$ 0.04 & 1.29 $\pm$ 0.04 & 0.65 $\pm$ 0.04 \\ 
\midrule
\multicolumn{11}{c}{\textit{\ourmodel{}}, Reinforcement Learning Fine-Tuning} \\
\midrule
\ourmodel{}-LFM2-RFT-GT         & 43\% & 1.72 $\pm$ 0.19 & 1.14 $\pm$ 0.12 & 0.78 $\pm$ 0.10 & 0.40 $\pm$ 0.05 & 23\% & 4.06 $\pm$ 0.04 & 2.02 $\pm$ 0.04 & 1.28 $\pm$ 0.05 & 0.64 $\pm$ 0.05 \\
\ourmodel{}-LFM2-RFT-MT         & 43\% & 1.79 $\pm$ 0.18 & 1.22 $\pm$ 0.12 & 0.85 $\pm$ 0.10 & 0.43 $\pm$ 0.05 & 23\% & 4.16 $\pm$ 0.04 & 2.13 $\pm$ 0.04 & 1.35 $\pm$ 0.04 & 0.68 $\pm$ 0.04\\
\end{tabular}
    \caption{Main ChemCensor benchmark results with 95\% confidence intervals. CIs are estimated using the normal approximation for USPTO-50K-test ($n=4{,}972$) and nonparametric bootstrap with $B=2{,}000$ replicates for URSA-expert-2026 ($n=100$). Intervals are marginal per model--metric pair and are not corrected for multiple comparisons.}
    \label{tab:ci}
\end{sidewaystable*}
\newpage
\section{ChemCensor Validation}
\label{app:chemcensor_validation}

Human expertise can be applied to the outputs of C3LM and other models to provide an independent, more detailed evaluation. We scored random samples \texttt{C3LM-Qwen3-CREED-CCV+USPTO-AR} (N=25), \texttt{C3LM-Qwen3-RFT-CC} (N=25), Gemini 3 Flash (N=25) and Grok 4.1 (N=25) outputs by experts who use SciFinder to prove their scoring. The comparison of “expert” scores and ChemCensor values is outlined in \autoref{tab:chemcensor_validation}. Obviously, this practice can not be scaled to all outputs from all models. But that was the very purpose of ChemCensor to allow for rapid and scaled evaluation that at some extent mimicking the expertise of chemists. The confusion matrix of this study is provided in \autoref{tab:conf_matrix}.

\begin{table*}[th!]
\centering

\begin{tabular}{|c|c|c|}
\hline
 & Marked as ``positive'' by the expert & Marked as ``negative'' by the expert \\
\hline
Scored as ``positive'' in CC & \cellcolor{green!20}38 & \cellcolor{red!15}3 \\
\hline
Scored as ``negative'' in CC & \cellcolor{red!15}10 & \cellcolor{green!20}49 \\
\hline
\multicolumn{2}{|c|}{Accuracy} & 0.87 \\
\hline
\end{tabular}

\caption{Agreement between expert annotations and ChemCensor-based reaction plausibility classification.}
\label{tab:conf_matrix}
\end{table*}

\includepdf[
  pages=1,
  offset=0 -2mm,
  scale=0.92,
  pagecommand={%
    \thispagestyle{plain}%
    \captionsetup{type=table}%
    \vspace*{0em}%
    \caption{An independent review of models’ output by an expert organic chemist and ChemCensor binary verification.}%
    \label{tab:chemcensor_validation}%
    \vspace*{0em}%
  }
]{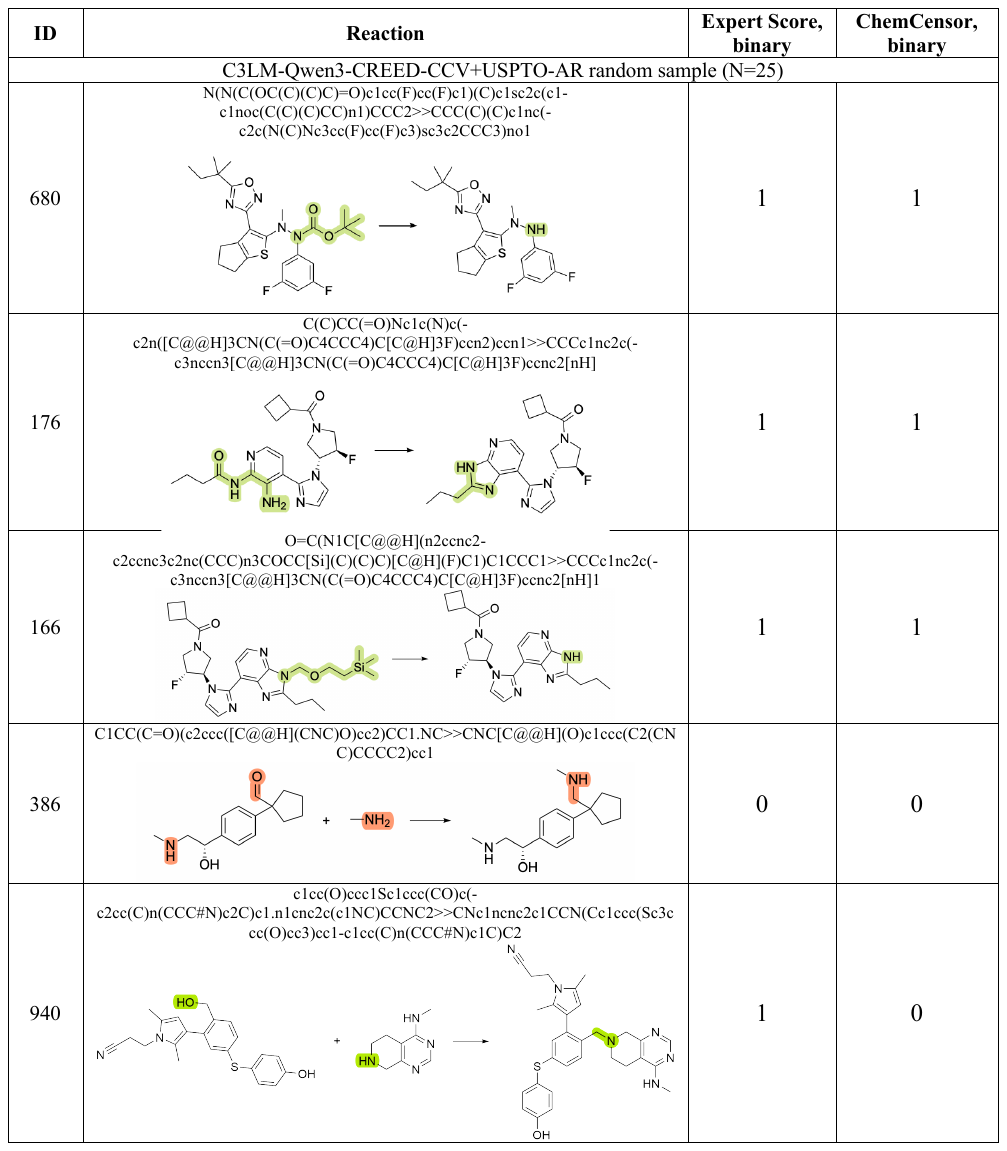}
\includepdf[
  pages=2-,
  scale=0.92,
  pagecommand={%
    \thispagestyle{plain}%
    \captionsetup{type=table}%
    \caption*{Continuation of ~\autoref{tab:chemcensor_validation}.}%
  }
]{figs/appendix/Independent_review_100_reactions.pdf}


\end{document}